\documentclass[10pt,twocolumn,letterpaper]{article}
\usepackage[pagenumbers]{cvpr}

\makeatletter
\@namedef{ver@everyshi.sty}{}
\makeatother

\usepackage[dvipsnames]{xcolor}
\usepackage{algorithm}
\usepackage{algorithm}
\usepackage{algpseudocode}
\usepackage{amsmath}
\usepackage{amssymb}
\usepackage{amsthm}
\usepackage{bbm}
\usepackage{bm}
\usepackage{booktabs}
\usepackage{colortbl}
\usepackage{enumitem}
\usepackage{etoolbox}
\usepackage{graphicx}
\usepackage{lipsum}
\usepackage{listings}
\usepackage{multirow, makecell}
\usepackage{pgfplots}
\usepackage{soul}
\usepackage{subcaption}
\usepackage{subcaption}
\usepackage{textcomp,gensymb} 
\usepackage{xcolor, color, colortbl}

\usepackage{tikz}
\usetikzlibrary{spy,backgrounds}

\definecolor{cvprblue}{rgb}{0.21,0.49,0.74}
\usepackage[pagebackref,breaklinks,colorlinks,citecolor=cvprblue]{hyperref}

\definecolor{tablered}{rgb}{1, 0.7, 0.7}

\newcommand{\best}{\cellcolor{tablered}}
\newcommand{\sbest}{\cellcolor{orange!40}}

\newcommand{\supp}{\textit{Supplementary Material}\xspace}

\newcommand{\REMOVAL}[1]{}

\definecolor{Highlight}{HTML}{39b54a}  

\usepackage{amssymb}
\usepackage{pifont}
\makeatletter
\AfterEndEnvironment{algorithm}{\let\@algcomment\relax}
\AtEndEnvironment{algorithm}{\kern2pt\hrule\relax\vskip3pt\@algcomment}
\let\@algcomment\relax
\newcommand\algcomment[1]{\def\@algcomment{\footnotesize#1}}
\renewcommand\fs@ruled{\def\@fs@cfont{\bfseries}\let\@fs@capt\floatc@ruled
  \def\@fs@pre{\hrule height.8pt depth0pt \kern2pt}%
  \def\@fs@post{}%
  \def\@fs@mid{\kern2pt\hrule\kern2pt}%
  \let\@fs@iftopcapt\iftrue}
\makeatother

\newcommand{\comma}{\,,}
\newcommand{\point}{\,.}

\usepackage[normalem]{ulem}

\newlength\mytmplen
\DeclareRobustCommand{\zoomin}[9]{ %
\begin{tikzpicture}[spy using outlines={rectangle,#9,magnification=#8,size=#6}]   
	\node[anchor=south west,inner sep=0]  {\includegraphics[width=#7]{#1}};
	\spy on (#2, #3) in node at (#4,#5);
\end{tikzpicture}
}

\ExplSyntaxOn
\clist_const:Nn \l_soul_protect_clist { cite , citep , ref , eqref }
\NewDocumentCommand{\robusthl}{m}{
  \tl_set:Nn \l_tmpa_tl { #1 }
  \clist_map_inline:Nn \l_soul_protect_clist {
    \regex_replace_all:nnN { (\c{##1}[^{}]*\{[^{}]*\}) } %
                           { \c{mbox}\{\1\} } \l_tmpa_tl
  }
  \hl\l_tmpa_tl
}
\ExplSyntaxOff

\global\long\def\multiplicationSymbol{\ } 

\global\long\def\convexIndex{J} 

\global\long\def\pointsPerConvex{K}

\global\long\def\setOfPoints{\mathbb{S}}

\newcommand{\methodname}{3DCS\xspace}

\newcommand{\cvxnet}{CvxNet}

\def\paperID{2612}
\def\confName{CVPR}
\def\confYear{2025}

\newcommand{\mysection}[1]{\vspace{2pt}\noindent\textbf{#1}}
\newcommand{\myTitle}[1]{\textbf{#1}}

\title{3D Convex Splatting: Radiance Field Rendering with 3D Smooth Convexes} 

\author{Jan Held\thanks{Equal contributions} $^{1,2}$
\quad
Renaud Vandeghen\footnotemark[1] $^{1}$  
\quad
Abdullah Hamdi\footnotemark[1] $^{3}$  
\quad
Adrien Deliege$^{1}$  
\quad
Anthony Cioppa$^{1}$  
\\
Silvio Giancola$^{2}$  
\quad
Andrea Vedaldi$^{3}$  
\quad
Bernard Ghanem$^{2}$  
\quad
Marc Van Droogenbroeck$^{1}$  
\\ $^1$ {\small University of Liège}
\quad $^2$ {\small KAUST}
\quad $^3$ {\small University of Oxford}
}

\begin{document}
\maketitle

\begin{abstract}

Recent advances in radiance field reconstruction, such as 3D Gaussian Splatting (3DGS), have achieved high-quality novel view synthesis and fast rendering by representing scenes with compositions of Gaussian primitives. 
However, 3D Gaussians present several limitations for scene reconstruction. 
Accurately capturing hard edges is challenging without significantly increasing the number of Gaussians, creating a large memory footprint.
Moreover, they struggle to represent flat surfaces, as they are diffused in space. 
Without hand-crafted regularizers, they tend to disperse irregularly around the actual surface.
To circumvent these issues, we introduce a novel method, named 3D Convex Splatting (\methodname), which leverages 3D smooth convexes as primitives for modeling geometrically-meaningful radiance fields from multi-view images. 
Smooth convex shapes offer greater flexibility than Gaussians, allowing for a better representation of 3D scenes with hard edges and dense volumes using fewer primitives. 
Powered by our efficient CUDA-based rasterizer, \methodname achieves superior performance over 3DGS on benchmarks such as Mip-NeRF360, Tanks and Temples, and Deep Blending. 
Specifically, our method attains an improvement of up to $0.81$ in PSNR and $0.026$ in LPIPS compared to 3DGS while maintaining high rendering speeds and reducing the number of required primitives. 
Our results highlight the potential of 3D Convex Splatting to become the new standard for high-quality scene reconstruction and novel view synthesis.
Project page: \url{convexsplatting.github.io}.

\end{abstract}

\begin{figure}[t]
\centering
\includegraphics[width=0.99\linewidth]{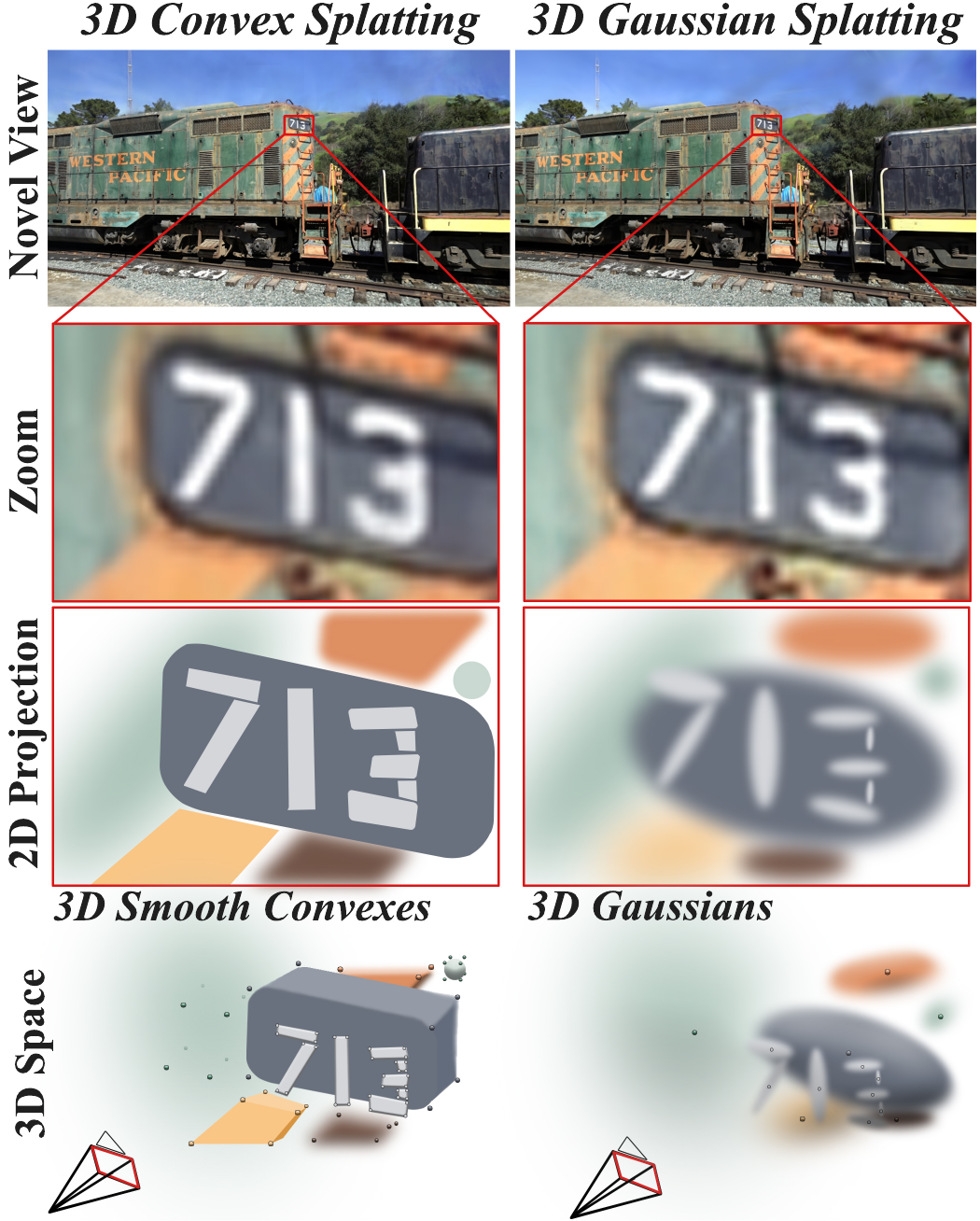} 
\caption{
\myTitle{3D Convex Splatting for Novel View Synthesis.}
We introduce a novel primitive-based pipeline for novel view synthesis with 3D smooth convexes. 
Our 3D smooth convexes share the rendering speed of 3D Gaussians~\cite{Kerbl20233DGaussian} and the flexible representation of smooth convexes~\cite{Deng2020CvxNet}.
As a result, 3D Convex Splatting better reconstructs scenes with fewer primitives.
}%
\label{fig:pullingfigure}
\end{figure}
\section{Introduction}%
\label{sec:intro}

Reconstructing complex scenes and synthesizing novel views have been fundamental challenges in computer vision and graphics~\cite{Faugeras1992What, Agarwal2011Building}, with applications ranging from virtual reality to autonomous navigation~\cite{Newcombe2011KinectFusion, Hamdi2023SPARF, Mai2024TrackNeRF}.
Neural Radiance Fields (NeRF)~\cite{Mildenhall2020NeRF-eccv} revolutionized this area by modeling scenes as continuous volumetric radiance fields, which are optimized to render novel views at high-quality. However, NeRF suffers from slow training and rendering times, limiting its practicality.
To address these issues, 3D Gaussian Splatting (3DGS)~\cite{Kerbl20233DGaussian} emerged as an efficient alternative, by representing scenes with millions of 3D Gaussian.
3DGS significantly accelerated training and enabled real-time rendering while maintaining high-quality outputs.

Despite this progress, Gaussian primitives have two main limitations. 
(1) They lack defined physical boundaries, making them unsuitable for accurately representing flat surfaces or enabling physically meaningful scene decompositions. 
(2) In addition to their specific smoothness and rounded nature, Gaussians are inadequate for capturing hard edges and geometric structures.
Each Gaussian behaves similarly to an ellipsoid, with a symmetrical distribution, struggling to conform to angular boundaries or flat surfaces.
This inherent limitation is reflected in the sphere packing problem~\cite{Hales1992TheSphere, Conway1999Sphere}, where densely packed spherical or ellipsoidal shapes leave gaps and result in inefficient coverage, especially along flat or sharp corners.
As with spheres or ellipsoids, an impractically large number of Gaussian would be needed to fill space without gaps, leading to increased memory consumption and computational overhead.

\begin{figure}[t]
\centering
\includegraphics[width=0.99\linewidth]{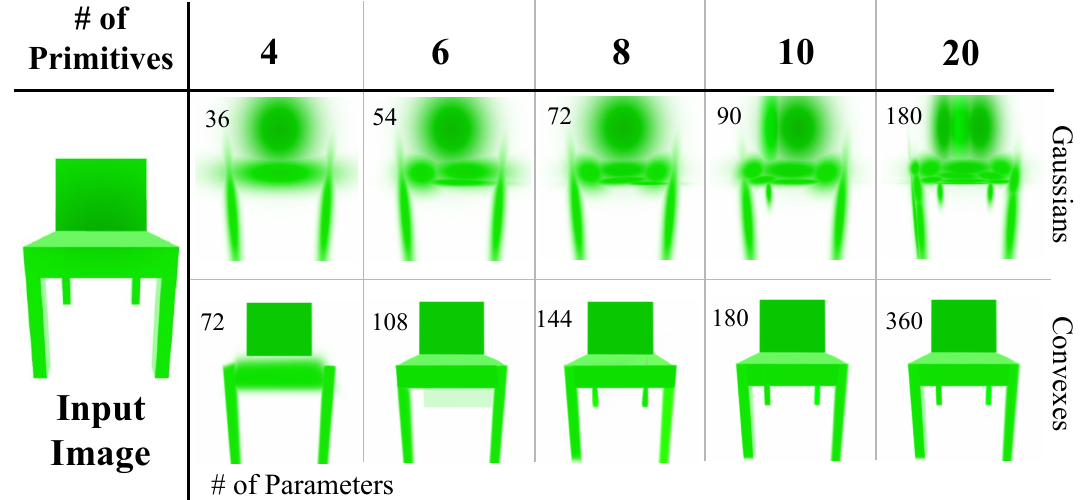}
\caption{\myTitle{Toy Experiment of Modeling a Chair.}
For the chair input image, we use  Gaussians and smooth 6-point convexes to fit the chair with an increasing number of primitives.
Note how the convexes efficiently represent the chair with fewer parameters.
}%
\label{fig:gauss_vs_convex}
\end{figure}

To overcome these limitations, we propose a novel method called \emph{3D Convex Splatting (\methodname)}, which leverages \emph{3D smooth convexes} as primitives for modeling and reconstructing geometrically accurate radiance fields from multi-view images.
3D smooth convexes offer greater flexibility than Gaussians, as they can form dense volumes that accurately capture hard edges and detailed surfaces using fewer primitives.
\Cref{fig:pullingfigure} illustrates this point, showing that \methodname enables the rendering of 3D smooth convexes to generate high-quality novel views of complex scenes.
Moreover, by incorporating \emph{smoothness} and \emph{sharpness} parameters, we can  control the curvature and the diffusion of the smooth convexes, respectively. This enables the creation of shapes that are hard or soft, dense or diffuse.
\Cref{fig:gauss_vs_convex} shows a toy example of how smooth convexes can represent a chair with hard edges with far fewer elements than Gaussians, while utilizing the same optimization. 
For novel view synthesis, we merge the benefits of the fast rendering process from Gaussians~\cite{Kerbl20233DGaussian}, with the flexibility of 3D smooth convexes \cite{Deng2020CvxNet}.
We achieve this by rendering 3D smooth convexes using our efficient CUDA-based rasterizer, which enables real-time rendering and accelerates the optimization process.
To the best of our knowledge, 3D Convex Splatting is the first method to leverage differentiable smooth convex shapes for novel view synthesis on realistic scenes, outperforming previous methods that use other primitives.

\mysection{Contributions.}
We summarize our contributions as follows:
\textbf{(i)} We introduce 3D Convex Splatting (\methodname), utilizing 3D smooth convexes as novel primitives for radiance field representation, addressing the limitations of Gaussian primitives in capturing dense volumetric features.
\textbf{(ii)} We develop an optimization framework and a fast, differentiable GPU-based rendering pipeline for our 3D smooth convexes, enabling high-quality 3D scene representations from multi-view images and high rendering speeds.
\textbf{(iii)} \methodname surpasses existing rendering primitives on Mip-NeRF360, Tanks and Temples, and Deep Blending datasets, achieving better performance than 3D Gaussian Splatting while using a reduced number of primitives per scene.

\begin{figure*}[th]
\centering
\includegraphics[width=\linewidth]{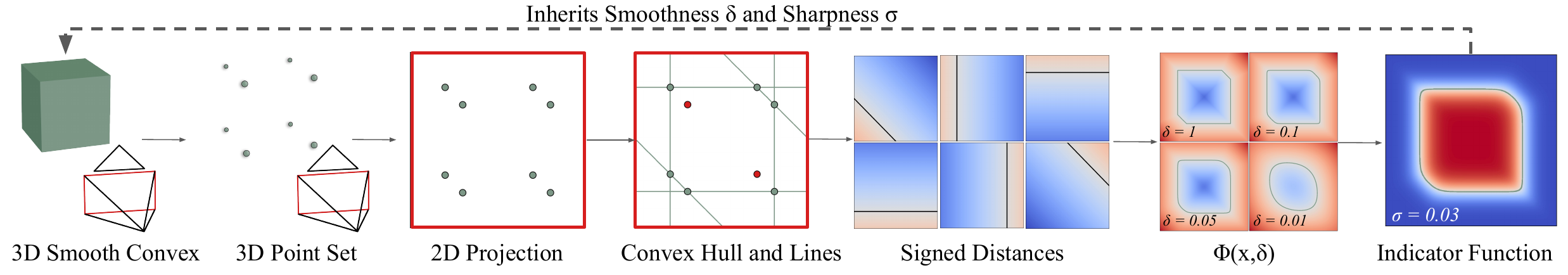}
\caption{
\myTitle{Convex Splatting Pipeline.} 
The 3D smooth convex is represented with a point set that is projected in the 2D camera plane.
We extract the line-delimited convex hull of the projected points and define the signed distance function for each line.
The lines are combined to define an indicator function for each pixel based on smoothness $\delta$ and sharpness $\sigma$ of the 3D convex.
The pipeline is differentiable end-to-end, which allows the parameters of the smooth convex primitives to be optimized based on the rendered images.
}%
\label{fig:pipeline}
\end{figure*}

\label{sec:related_work}

\mysection{Neural radiance fields (NeRF).}
Recovering the 3D structure of a scene from images captured from multiple viewpoints is a fundamental problem in computer vision~\cite{Faugeras1992What, Agarwal2011Building}. 
The introduction of Neural Radiance Fields (NeRF)~\cite{Mildenhall2020NeRF-eccv} revolutionized this field by representing scenes as volumetric radiance fields, enabling high-quality novel view synthesis~\cite{Verbin2022RefNeRF, Barron2021MipNeRF, Barron2022MipNeRF360}. 
NeRF employs multi-layer perceptrons to encode scene geometry and view-dependent appearance, optimized via photometric loss through volume rendering~\cite{Drebin1988Volume,Kajiya1984RayTracing,Levoy1990Efficient,Max1995Optical}. 
Enhancements to NeRF include grid-based representations for faster training~\cite{Muller2022Instant,Chen2022TensoRF,FridovichKeil2022Plenoxels,Sun2022Direct,Kulhanek2023TetraNeRF-arxiv},
baking techniques for accelerated rendering~\cite{Reiser2021KiloNeRF,Hedman2021Baking,Yariv2023BakedSDF-arxiv,Reiser2023MERF},
as well as addressing challenges such as antialiasing~\cite{Barron2022MipNeRF360,Barron2023ZipNeRF}, modeling unbounded scenes~\cite{Barron2022MipNeRF360,Zhang2020NeRF++-arxiv}, and adapting to few-shot~\cite{Jain2021Putting, Kim2022InfoNeRF, Du2023Learning} and one-shot settings~\cite{Yu2021pixelNeRF, Chan2022Efficient}.
In this work, we do not rely on neural networks to model radiance fields like other NeRFs, but instead optimize 3D smooth convexes to fit 3D scenes efficiently.
Yet, 3D Convex Splatting allows for strong modeling capacity that rivals MipNeRF-360~\cite{Barron2022MipNeRF360} in visual fidelity but with real-time rendering speed.

\mysection{Primitive-based differentiable rendering.}
Differentiable rendering techniques enable gradient computation through the rendering pipeline, facilitating the optimization of scene parameters from image observations~\cite{Loper2014OpenDR,Kato2018Neural,Liu2019SoftRasterizer,Liu2018Paparazzi,Petersen2019Pix2Vex-arxiv,Gross2007Point}.
Neural point-based rendering~\cite{Kato2018Neural} represents scenes with points that store learned features for geometry and texture. 
3D Gaussian Splatting (3DGS)~\cite{Kerbl20233DGaussian} introduces Gaussian primitives parameterized by positions, covariances, and appearance attributes. 
By optimizing millions of Gaussians, 3DGS 
and achieves high-quality rendering with significantly faster training times and real-time rendering capabilities. 
Enhancements to this approach include antialiasing techniques~\cite{Yu2024MipSplatting}, exact volumetric rendering of ellipsoids~\cite{Mai2024EVER-arxiv}, 
and extensions to dynamic scene modeling~\cite{Zhou2024HUGS-arxiv, Luiten2023Dynamic-arxiv}.
However, Gaussian primitives have inherent limitations due to their very specific smoothness, making it challenging to capture hard edges and dense volumetric structures without significantly increasing the number of primitives. 
This leads to increased memory consumption and computational overhead, hindering scalability and efficiency.
Alternative primitives have been explored to improve geometric representation. 
GES~\cite{Hamdi2024GES} utilizes generalized exponential functions to better capture signals with harder edges.
2D Gaussian splatting~\cite{Huang20242DGaussian} collapses the 3D Gaussians into oriented planar Gaussians to better represent surfaces.
In our work, we introduce 3D smooth convex shapes as a novel primitive for real-time rendering of novel views. 
These shapes address the limitations of the Gaussian primitives by efficiently capturing dense volumetric shapes.

\mysection{Convex shapes.}
Convex shapes have been extensively studied in 
computer graphics and computer vision due to their geometric simplicity and flexibility in representing complex objects~\cite{Graham1972AnEfficient, Preparata1977Convex}.
In 3D reconstruction, convex shape representations have proven highly effective for decomposing complex structures into simpler components like planes, spheres, cubes, cylinders, and superquadrics~\cite{Paschalidou2019Superquadrics, Tulsiani2016Learning-arxiv, Wei2022Approximate-arxiv}.
\cvxnet~\cite{Deng2020CvxNet} and BSP-Net~\cite{Chen2020BSPNet} introduce neural networks that learn hyperplanes to construct flexible convex shapes, enabling more accurate differentiable modeling of geometries with primitives.
A concurrent work utilizes rigid convex polyhedra and differentiable mesh rendering to fit simple 3D shapes with few primitives using multi-view supervision \cite{Ren2024Differentiable}.  However, these methods are limited to simple shapes with few optimized primitives and do not scale to large scenes or allow for accurate novel view synthesis.

In our work, we introduce splatting-based rasterization with an array of smooth, high-capacity primitives. This allows us to achieve rendering fidelity levels comparable to volume rendering techniques in modeling complex scenes.

\section{Methodology}

3D Convex Splatting (\methodname) combines smooth convexes (\cref{sec:preliminary}) from \cvxnet~\cite{Deng2020CvxNet} with the primitive-based representation from 3DGS~\cite{Kerbl20233DGaussian} for efficient, real-time novel view synthesis. 
\methodname uses a point-based convex shape representation and convex hull operations to enable straightforward differentiable projection of 3D convex shapes onto the 2D image plane (\cref{sec:rep3D}).
Further operations for smoothing, splatting, and adaptive densification are used to optimize the representation from posed images (\cref{sec:opti}).
\Cref{fig:pipeline} shows an overview of our convex splatting pipeline. 

\subsection{Preliminaries on 3D Smooth Convexes}%
\label{sec:preliminary}
 
Following \cvxnet~\cite{Deng2020CvxNet}, we define a convex polyhedron with $\convexIndex$ planes ($\convexIndex > 3$).
We define the signed distance $L_j(\mathbf{p})$ between a point $\mathbf{p} \in \mathbb{R}^3$ and a plane $\mathcal{H}_j$ as follow: 
\begin{equation}\label{eq:signed_distance}
    L_j(\mathbf{p}) = \mathbf{n}_j \cdot \mathbf{p} + d_j
\end{equation}
with $\mathbf{n}_j$ being the plane normal pointing towards the outside of the shape and $d_j$ its offset.
The signed distance from $\mathbf{p}$ to the convex shape is calculated as the maximum of all the $\convexIndex$ distances defined by $\tilde \phi(\mathbf{p}) = \max_{j=1,\cdots,\convexIndex}~ L_j(\mathbf{p})$.
However, to create a smooth representation of the convex shape, we use a smooth approximate signed distance function $\phi(\mathbf{p})$ $\approx \tilde \phi(\mathbf{p})$ using the LogSumExp function from \cvxnet~\cite{Deng2020CvxNet}:
\begin{equation} \label{eq:smoothness}
\phi(\mathbf{p}) = \log \left( \sum_{j=1}^{\convexIndex} \exp\left( \delta \multiplicationSymbol L_j(\mathbf{p}) \right) \right)\comma
\end{equation}
where the \emph{smoothness} parameter $\delta > 0$ controls the curvature of the convex approximation. Larger values of $\delta$ approximate $\phi(\mathbf{p})$ to $\tilde \phi(\mathbf{p})$ more closely, resulting in harder edges, while smaller values soften the vertices.

\begin{figure}[t]
\centering
\begin{tikzpicture}
\draw[->, thick] (-2.5,-1.5) -- (2.5,-1.5) node[midway, below] {\textit{(soft)} ~~~~~~~~~~~~~~~~~~~~~~ $\delta$ ~~~~~~~~~~~~~~~~~~~~~~ \textit{(hard)}};

\draw[->, thick] (-3.9,-1) -- (-3.9,1) node[midway, left, rotate=90, anchor=south] {\textit{(diffuse)} ~~~ $\sigma$ ~~~ \textit{(dense)}};

\matrix (m) [nodes={inner sep=0pt}, row sep=2pt, column sep=2pt] at (0,0) {
\node{\includegraphics[trim={120 140 100 140},clip,width=0.30\linewidth]{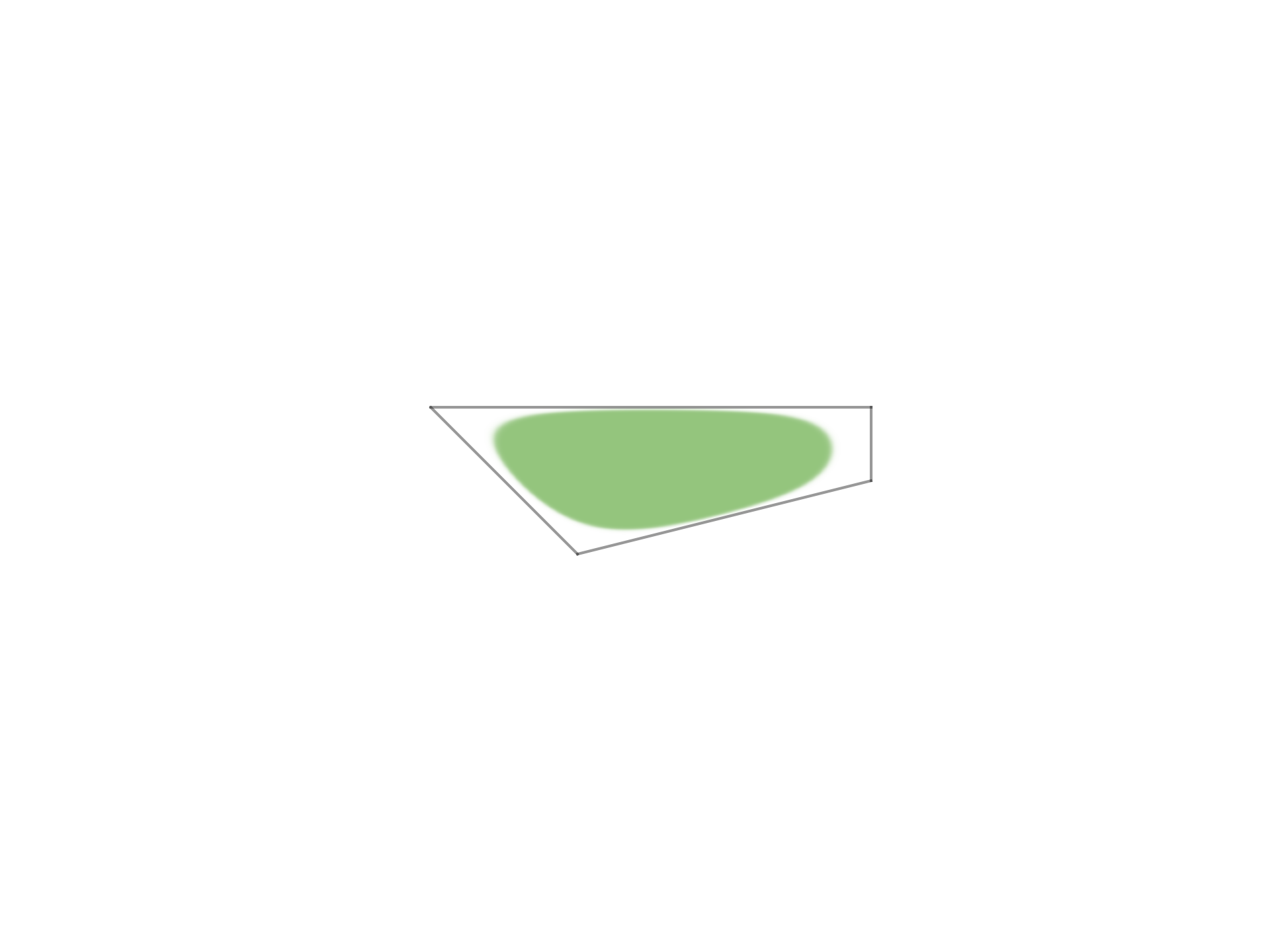}};&
\node{\includegraphics[trim={120 140 100 140},clip,width=0.30\linewidth]{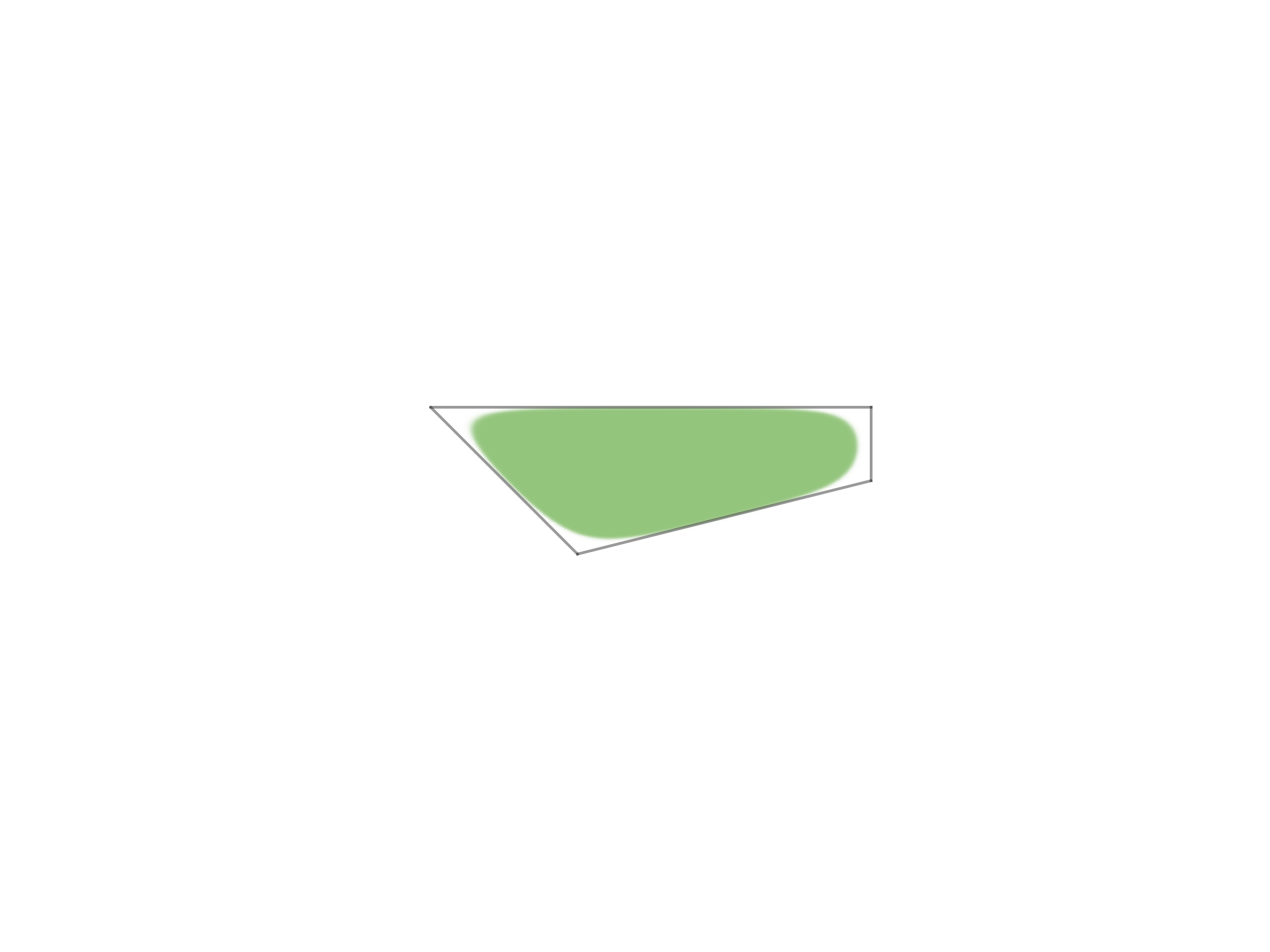}};& 
\node{\includegraphics[trim={120 140 100 140},clip,width=0.30\linewidth]{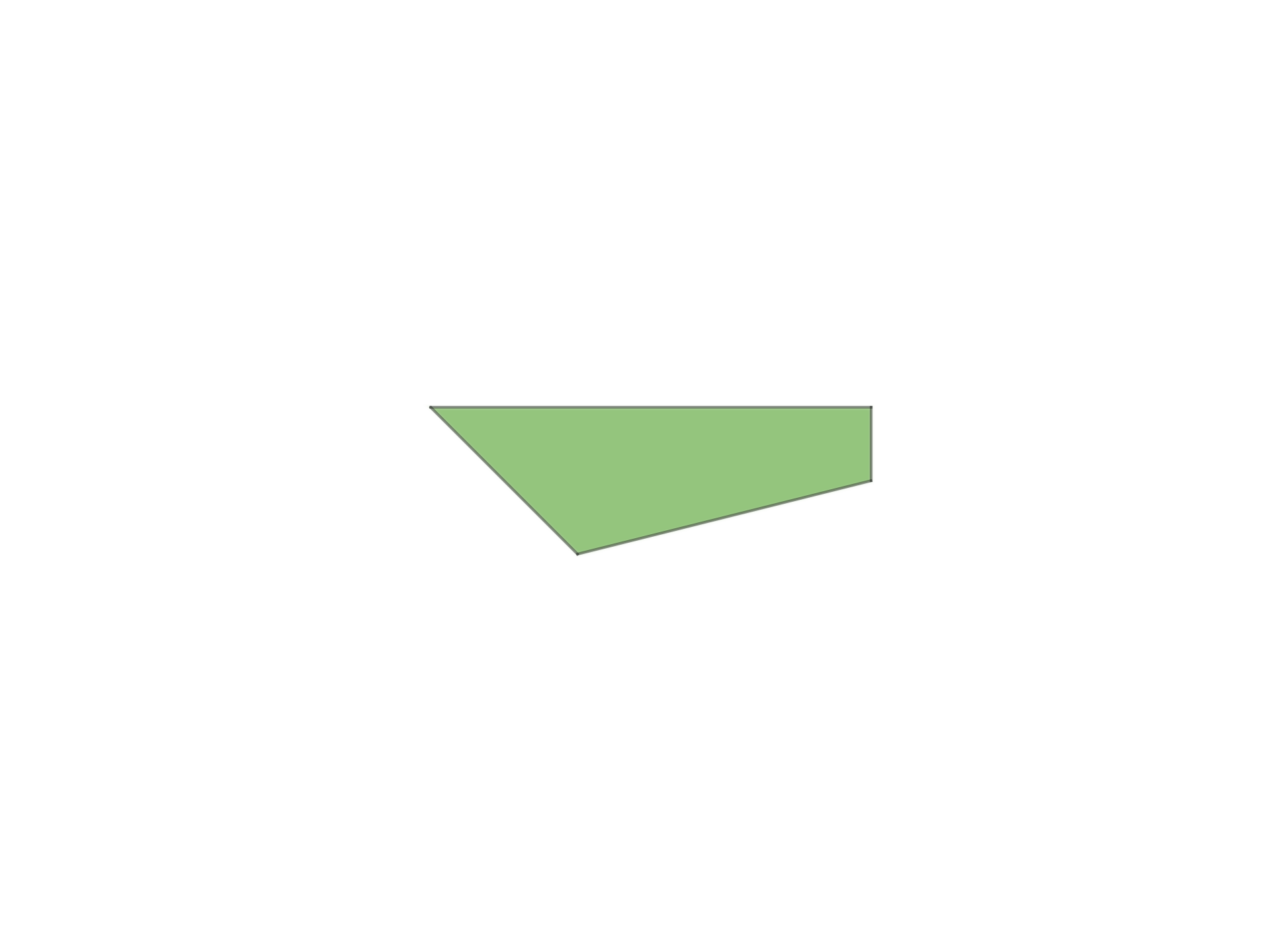}};\\
\node{\includegraphics[trim={120 140 100 120},clip,width=0.30\linewidth]{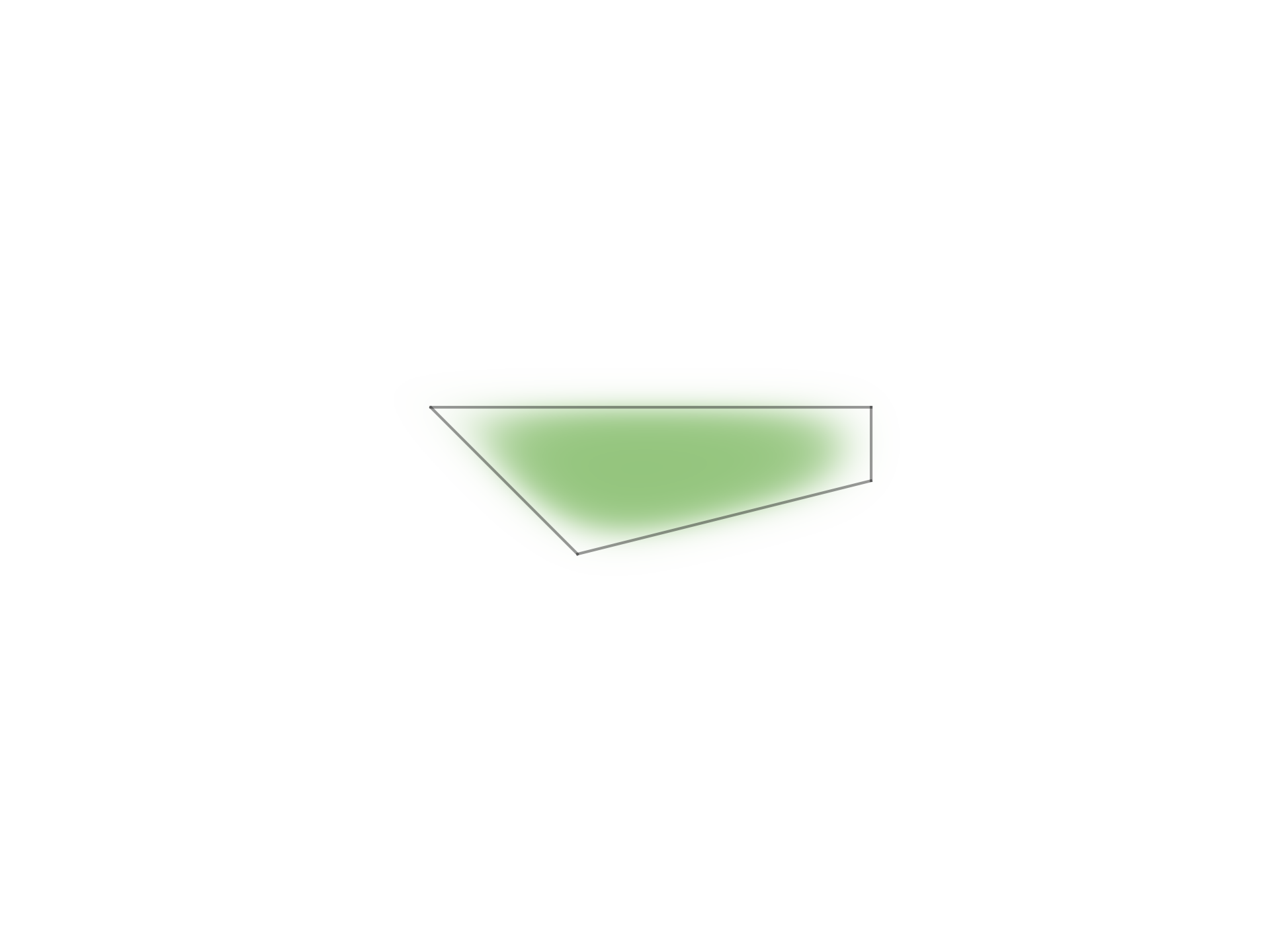}};& 
\node{\includegraphics[trim={120 140 100 120},clip,width=0.30\linewidth]{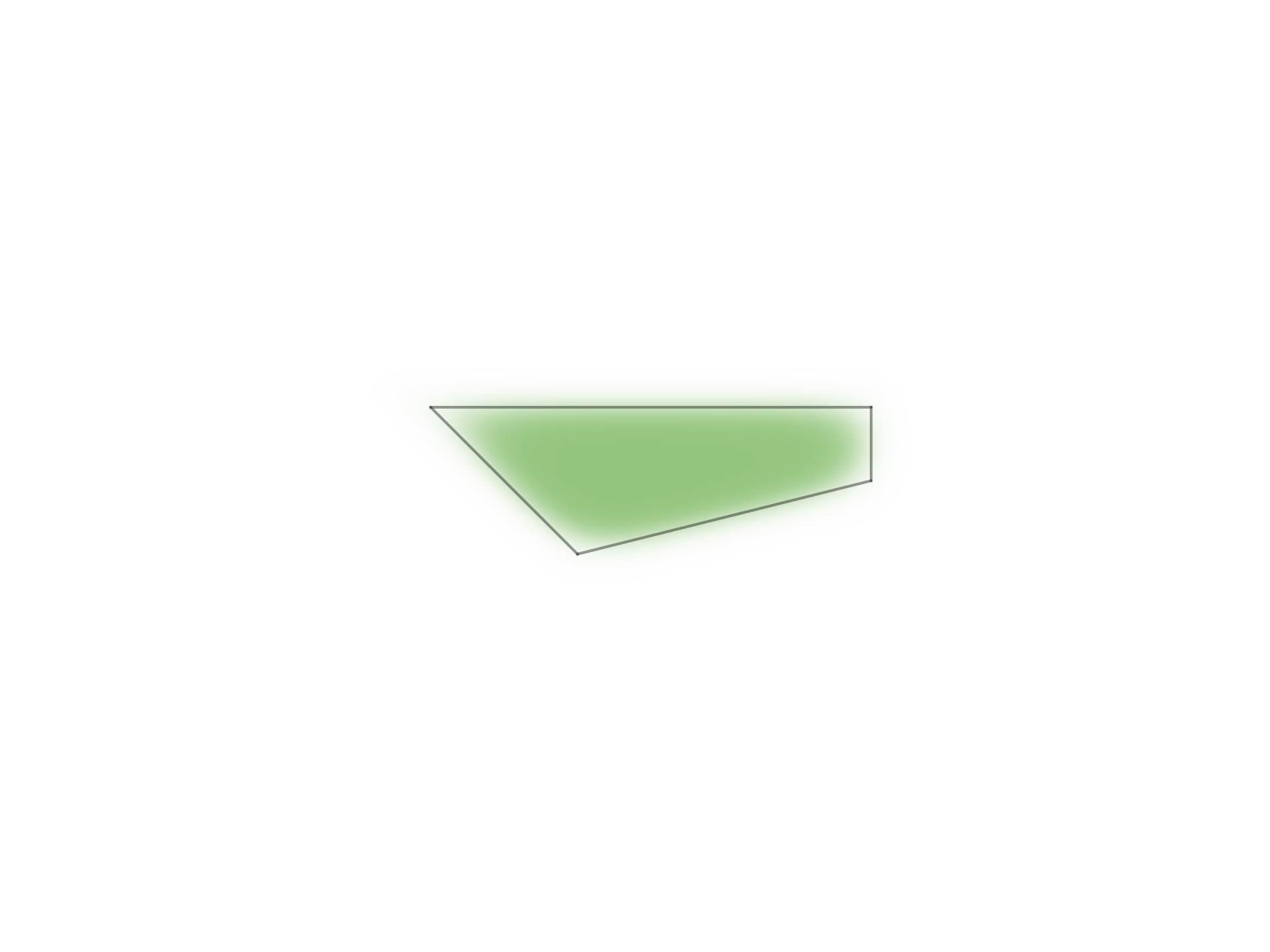}};& 
\node{\includegraphics[trim={120 140 100 120},clip,width=0.30\linewidth]{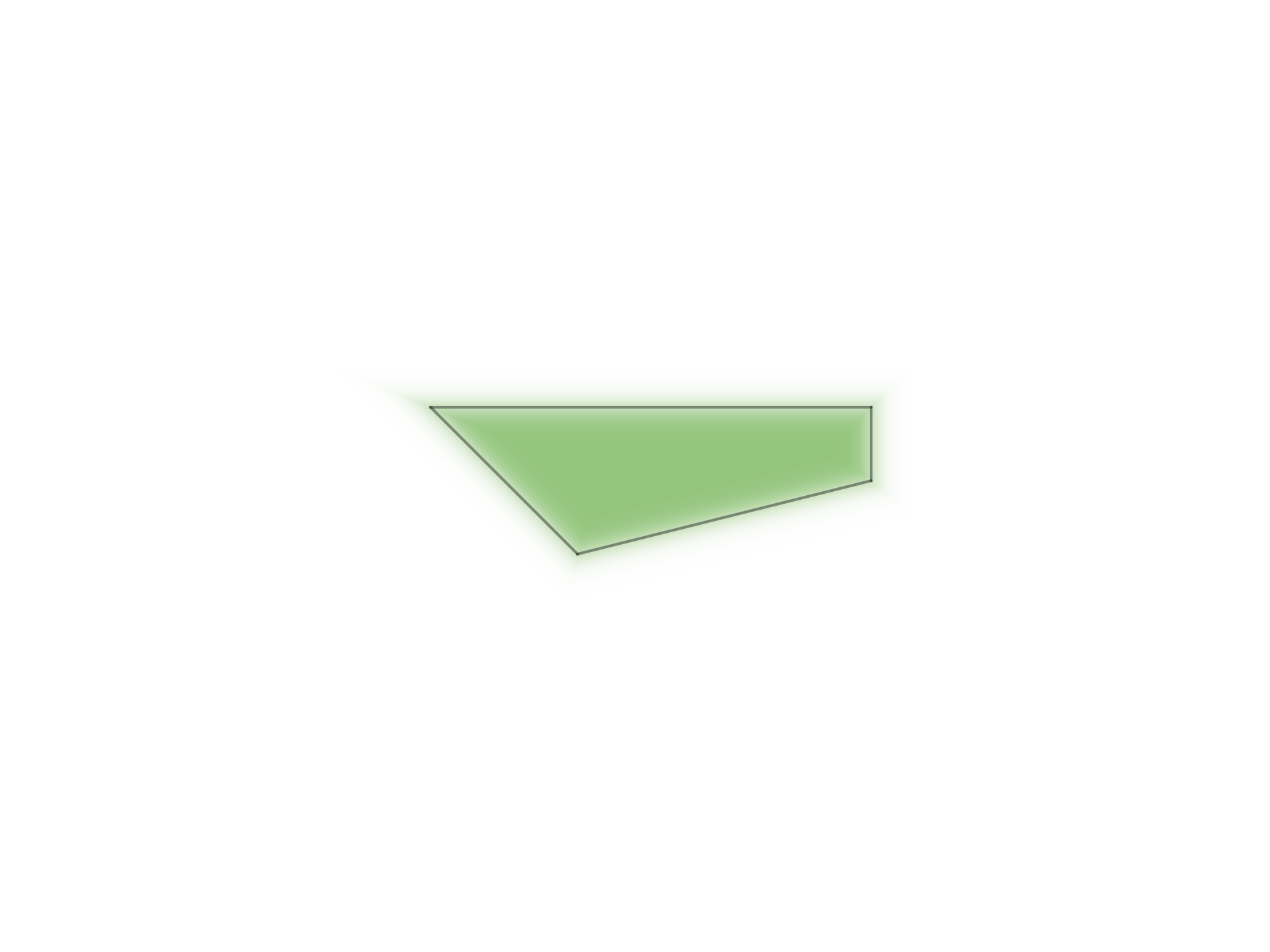}};\\
\node{\includegraphics[trim={120 130 100 110},clip,width=0.30\linewidth]{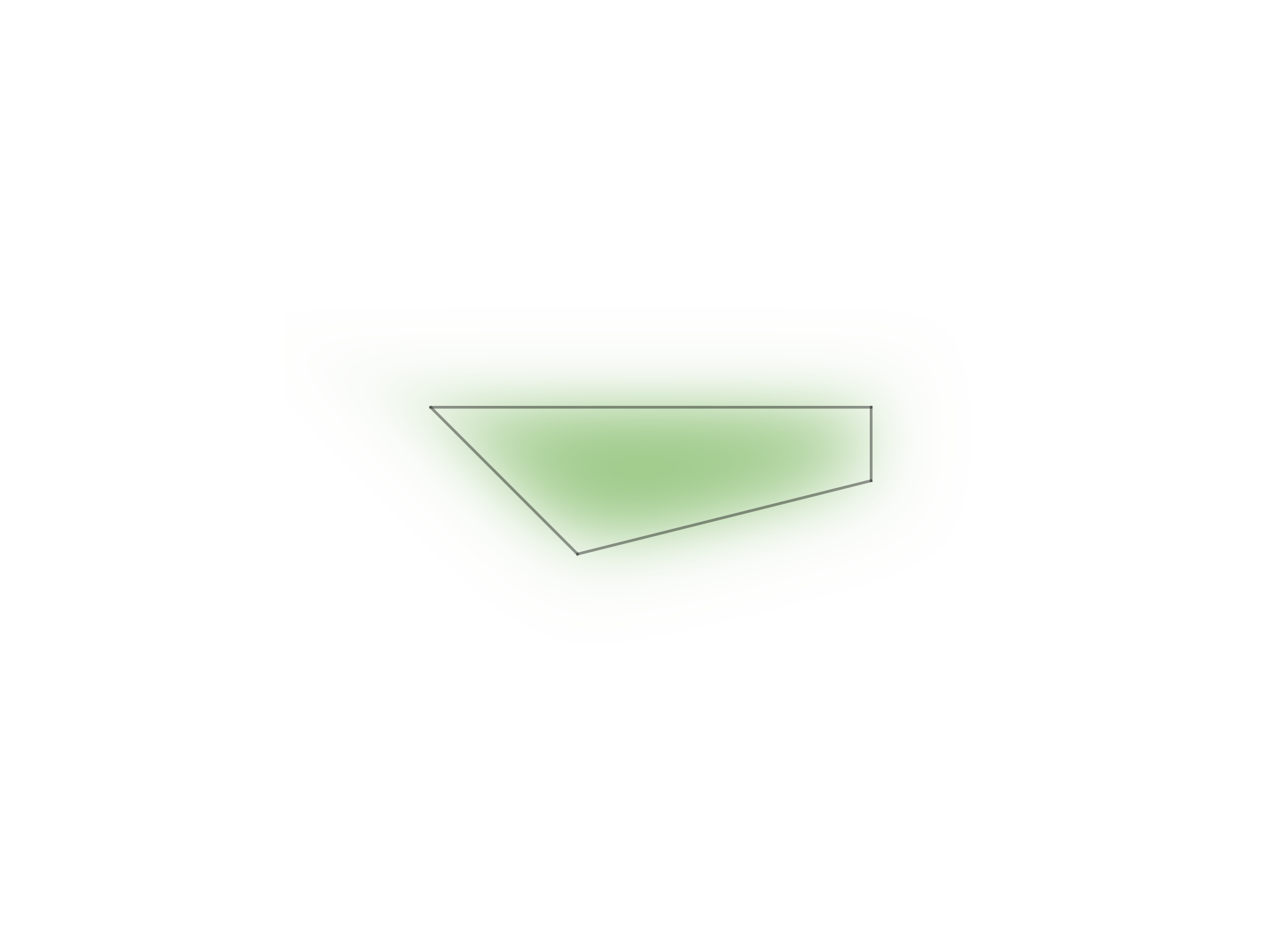}};& 
\node{\includegraphics[trim={120 130 100 110},clip,width=0.30\linewidth]{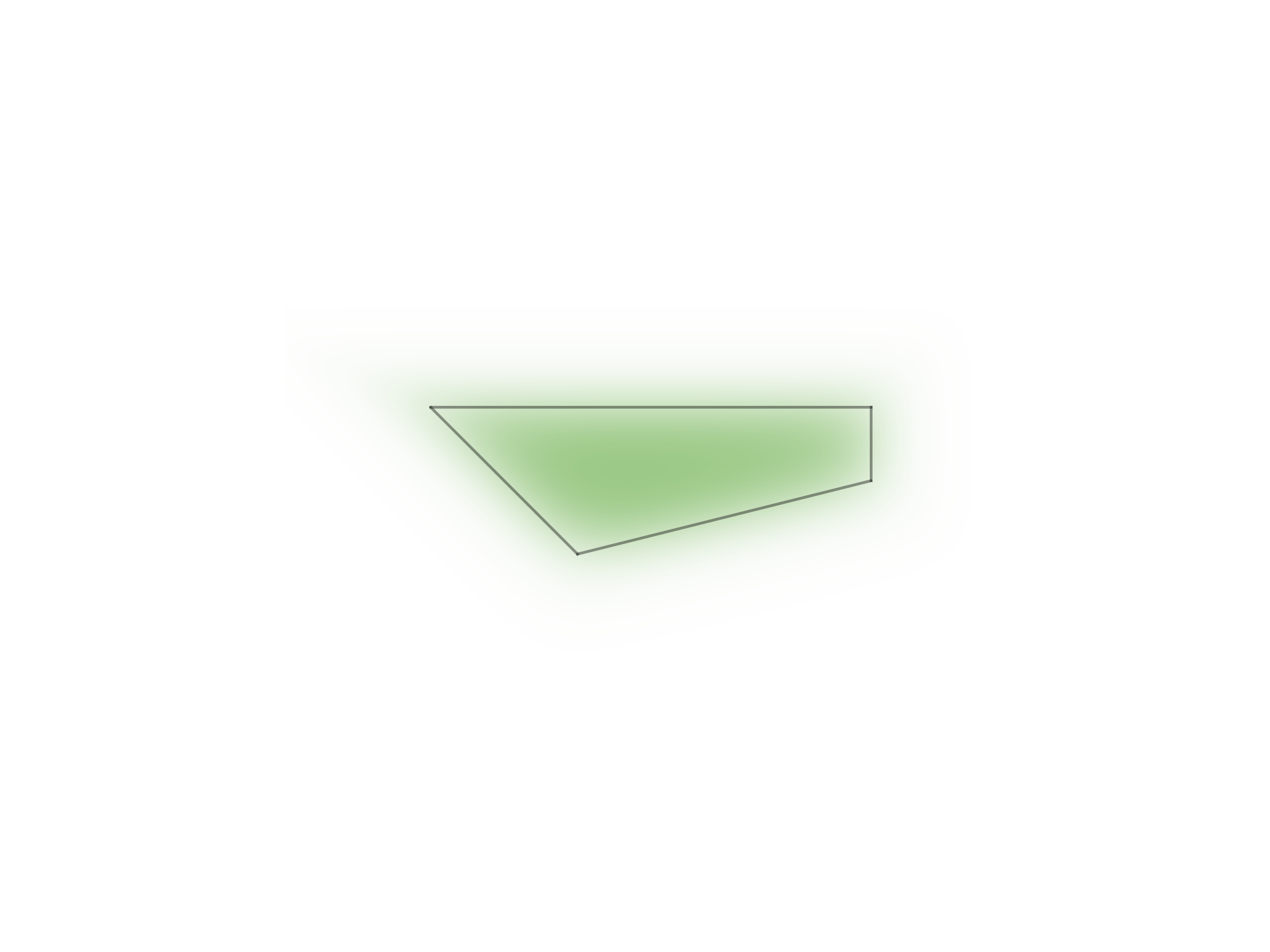}};& 
\node{\includegraphics[trim={120 130 100 110},clip,width=0.30\linewidth]{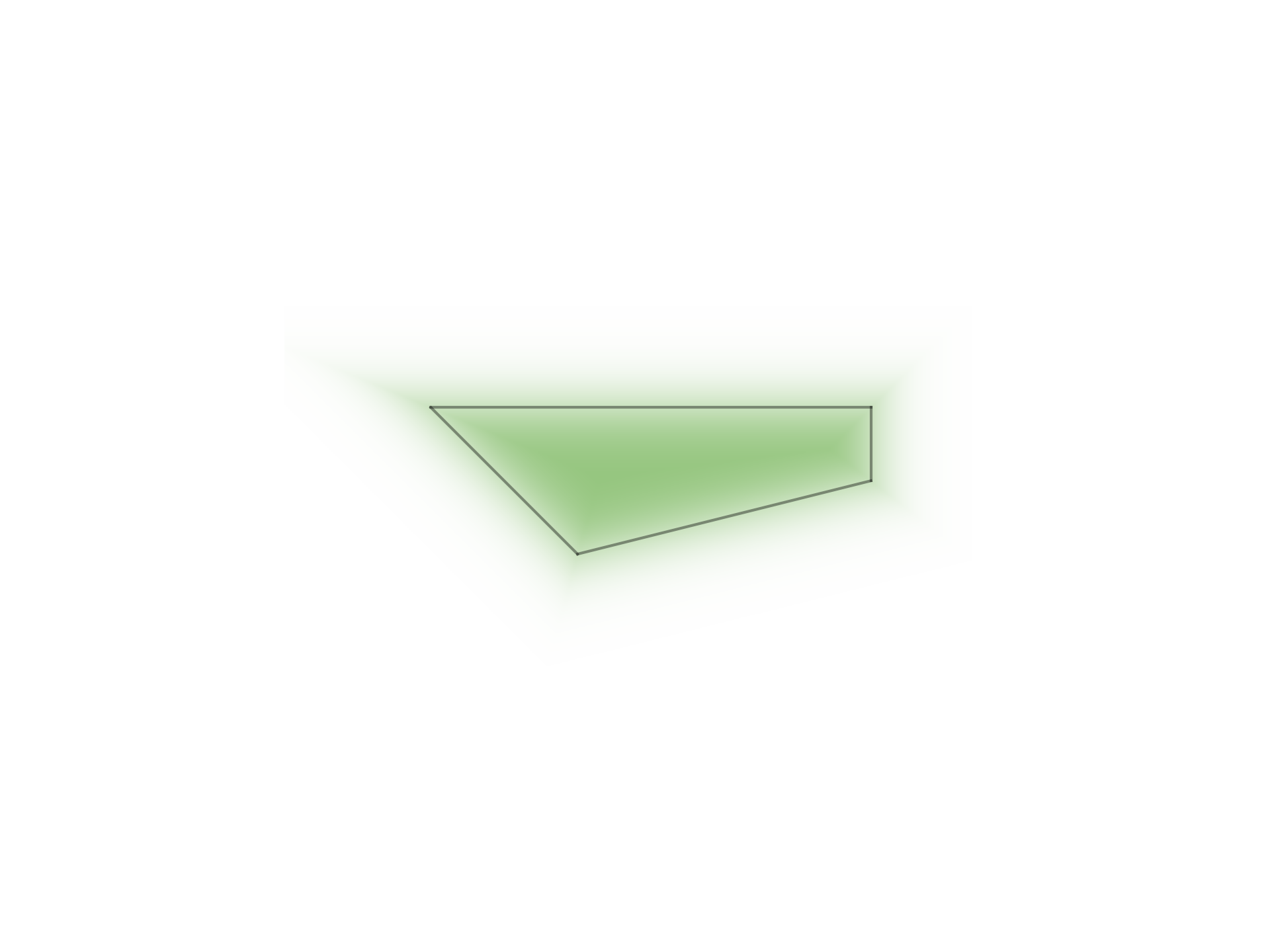}};\\ 
};
\end{tikzpicture}
\caption{
\myTitle{Effects of $\delta$ and  $\sigma$ on Splatting.} 
The smoothness $\delta$ characterizes vertices and edges, from soft to hard, while the sharpness $\sigma$ characterizes radiance field transitions, from diffuse to dense.
}%
\label{fig:influence_delta_sigma}
\end{figure}
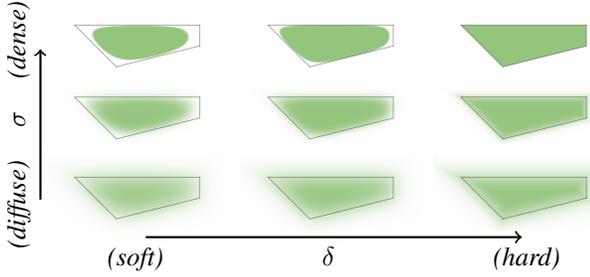

The indicator function $I(\mathbf{p})$ of the smooth convex is then defined by applying a sigmoid function to the approximate signed distance function~\cite{Deng2020CvxNet}:
\begin{equation} \label{eq:sharpness}
I(\mathbf{p}) = \operatorname{Sigmoid}\left( -\sigma \multiplicationSymbol \phi(\mathbf{p}) \right)\comma
\end{equation}
where the \emph{sharpness} parameter $\sigma > 0$ controls how rapidly the indicator function transitions at the boundary of the underlying convex shape. Higher values of $\sigma$ result in a steeper transition, making the shape's boundary more defined, whereas lower values result in a more diffuse shape.
At the boundary of the convex shape, where $\phi(\mathbf{p}) = 0$, the indicator function of the smooth convex satisfies $I(\mathbf{p}) = 0.5$.
More details about smooth convexes can be found in~\cite{Deng2020CvxNet}. \cref{fig:influence_delta_sigma} illustrates the effect of sharpness $\sigma$ and smoothness $\delta$ on the indicator function $I(.)$.

\subsection{\methodname: Splatting 3D Smooth Convexes} \label{sec:rep3D}

\mysection{Point-based 3D convex shape representation.}
Plane-based representations of 3D convexes are impractical for camera plane projections.
Unlike \cvxnet~\cite{Deng2020CvxNet}, we define a 3D convex as the convex hull of a 3D point set $\setOfPoints = \{\mathbf{p}_1, \mathbf{p}_2, \ldots, \mathbf{p}_\pointsPerConvex\}$. 
During optimization, the 3D points can move freely, allowing for flexible positioning and morphing of the convex shape.
Note that this set of $\pointsPerConvex$ points does not necessarily correspond to the \textit{explicit} vertices of a convex polyhedron, but rather the hull of the 3D convex shape.

\mysection{Differentiable projection onto the 2D image plane.} 
To be efficient, we do not explicitly build the 3D convex hull and project it into 2D, but instead, we project the 3D points into 2D and then construct its 2D convex hull.
Specifically, we project each 3D point $\mathbf{p}_k \in \setOfPoints$ onto the 2D image plane using the pinhole perspective camera projection model. The projection involves the intrinsic camera matrix $\mathbf{K}$ and extrinsic parameters (rotation $\mathbf{R}$ and translation $\mathbf{t}$):
\begin{equation}
\label{eq:projection}
\mathbf{q}_k = \mathbf{K} \left( \mathbf{R} \mathbf{p}_k + \mathbf{t} \right), ~~ \forall k = 1, 2, \dots, \pointsPerConvex\point
\end{equation}
This projection is differentiable, allowing gradients to flow back to the 3D points during optimization.

\mysection{2D convex hull computation.}
To construct the convex shape in 2D, we apply the Graham Scan algorithm~\cite{Graham1972AnEfficient}, which efficiently computes the convex hull by retaining only the points that define the outer boundary of the projected shape.
This approach ensures that the 2D projection accurately represents the convex outline needed for rendering. 
The Graham Scan starts by sorting the points based on their polar angle relative to a reference point.
After sorting, the Graham Scan algorithm is applied to construct the convex hull by iteratively adding points to the hull while maintaining convexity.
The convexity is ensured by checking the cross product of the last two points $\mathbf{q}_i$ and $\mathbf{q}_j$ on the convex hull and the current point $\mathbf{q}_k$, removing points that form a right turn (negative cross product).

\mysection{Differentiable 2D convex indicator function.}
We define the 2D convex indicator function of our convex hull by extending the smooth convex representation from 3D to 2D, reusing the equations introduced in \cref{sec:preliminary}.
We define $\phi(\mathbf{q})$ and $I(\mathbf{q})$ as in \cref{eq:smoothness,eq:sharpness}, but replace the 3D point $\mathbf{p}$ with the 2D point $\mathbf{q}$ and the planes delimiting the 3D convex hull by the lines delimiting the  resulting 2D convex hull.
The parameters $\sigma$ and $\delta$ are inherited from the 3D smooth convex (from \cref{eq:smoothness,eq:sharpness}) and still control the sharpness and smoothness of the projected 2D shape boundaries.
To account for perspective effects in the 2D projection, we scale $\delta$ and $\sigma$ by the distance $d$, ensuring that the appearance of the convex shape remains consistent with respect to its distance to the camera.

\mysection{Efficient differentiable rasterizer.}
To enable real-time rendering, we build our rasterizer following the 3DGS tile-based rasterizer~\cite{Kerbl20233DGaussian}, which allows for efficient backpropagation across an arbitrary number of primitives. All computations, including 3D-to-2D point projection, convex hull calculation, line segment definition, and indicator function implementation—are fully differentiable and executed within our custom CUDA kernels to maximize efficiency and rendering speed.
During rendering, we rasterize the 3D shape into the target view using $\alpha$-blending. 
For a given camera pose $\theta$ and $N$ smooth convexes to render each pixel $\mathbf{q}$, we order the $N$ convexes by sorting them according to increasing distance defined from the camera to their centers, and compute the color value for each pixel $\mathbf{q}$:
\begin{equation}
    C(\mathbf{q}) = \sum_{n=1}^N \mathbf{c}_n o_n I(\mathbf{q}) \left( \prod_{i=1}^{n-1} \left(1 - o_i I(\mathbf{q})\right) \right)\comma
\end{equation}
where $\mathbf{c}_n$ is the color of the $n$-th smooth convex, stored as spherical harmonics and converted to color based on the pose $\theta$, $o_n$ the opacity of the $n$-th shape, and
$I(\mathbf{q})$ is the indicator function adapted to our case from \cref{eq:sharpness}.

\subsection{Optimization}%
\label{sec:opti}

\mysection{Initialization and losses.}
We optimize the position of each point set in 3D, $\delta$ and $\sigma$ parameters, the opacity $o$, and the spherical harmonic color coefficients $\mathbf{c}$.
To constrain opacity within the range $[0, 1]$, we apply a sigmoid activation function.
For $\delta$ and $\sigma$, we use an exponential activation to ensure their values remain positive.
We initialize each convex shape with a set of points uniformly distributed around a sphere centered in the points of the point cloud, using the Fibonacci sphere algorithm. 
We define the size of each convex shape based on its average distance to the three nearest smooth convexes. This results in smaller smooth convexes in densely populated regions and larger shapes in sparser areas, allowing for adaptive scaling based on local geometry.
Following the approach used in 3DGS, we apply a standard exponential decay scheduling technique for the learning rate similar to Plenoxels~\cite{FridovichKeil2022Plenoxels}, but only for optimizing the position of the 3D points.
We experimented with applying this technique to $\delta$ and $\sigma$ as well, but we did not observe any performance improvements.
During training, we use the same regularization loss $\mathcal{L}_m$ as in~\cite{Lee2024Compact} to reduce the number of convexes.
Our final loss function combines $\mathcal{L}_1$ with a D-SSIM term and the loss for the mask, following~\cite{Lee2024Compact}:
\begin{equation}
    \mathcal{L} = (1 - \lambda) \mathcal{L}_1 + \lambda \mathcal{L}_{\text{D-SSIM}} + \beta \mathcal{L}_m\, ,
\end{equation}
where $\lambda$ controls the balance between $\mathcal{L}_1$ and $\mathcal{L}_{\text{D-SSIM}}$.
For all tests, we use $\lambda$ = 0.2 as in 3DGS~\cite{Kerbl20233DGaussian} and $\beta$ = 0.0005.

\begin{figure}[t]
    \centering
    \includegraphics[width=0.8\linewidth]{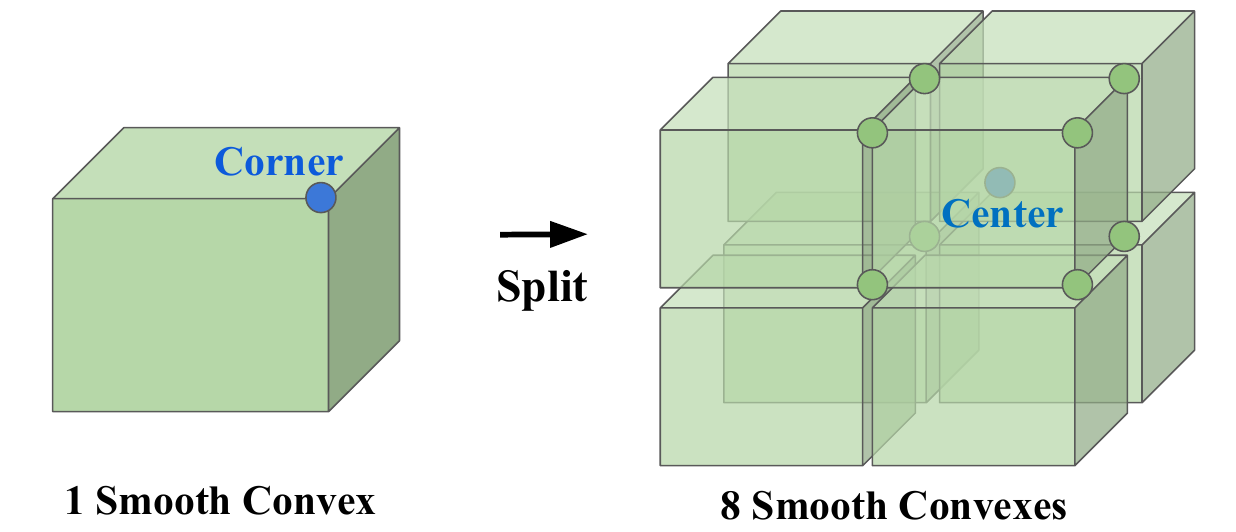}
    \caption{\myTitle{Adaptive Convex Densification Scheme.}
    We divide each convex, here exemplified with $\pointsPerConvex=8$ points, into as many scaled-down occurrences of the convex, centered at the initial points, each with reduced opacity. 
    }%
    \label{fig:splitting}
\end{figure}

\mysection{Adaptive convex shape refinement.}%
\label{adaptive_density_control}
The initial set of smooth convexes is generated from a sparse point set obtained through Structure-from-Motion. 
Since this initial number of smooth convexes is insufficient to accurately represent complex scenes, we employ an adaptive control mechanism to add smooth convexes dynamically.
In 3DGS, additional Gaussians are introduced by splitting or cloning those with large view-space positional gradients. 
However, in \methodname, positional gradients do not consistently correspond to regions with missing geometric features (``under-reconstruction'') or areas where convexes over-represent large portions of the scene (``over-reconstruction'').
Instead, we observe that \methodname exhibits a large sharpness $\sigma$ loss in both under-reconstructed and over-reconstructed regions. 
Rather than cloning and differentiating between small and large shapes, we consistently split our smooth convexes. Instead of splitting a smooth convex into just two new convexes, we split it directly into $K$ new convexes.
Each new convex shape is scaled down, and the centers of these new convexes correspond to the $\pointsPerConvex$ points defining the initial convex shape.
By placing the centers of the new convexes at the 3D points of the initial convex shape, we ensure that the new shapes collectively cover the volume of the original convex to maintain the overall completeness of the 3D representation (see \cref{fig:splitting} for illustration).
To encourage the formation of denser volumetric shapes during optimization, we increase the sharpness $\sigma$ throughout splitting, while keeping the smoothness $\delta$ the same.
We prune transparent convexes, \ie convexes that have an opacity lower than a predefined threshold, as well as, convexes that are too large. Details are provided in the experimental setup~\cref{subsec:real_world_exp}.

\begin{figure}[t]
\centering
\renewcommand{\arraystretch}{0} 
\setlength{\tabcolsep}{0pt} 
\begin{tabular}{cc||c|c|c|c}

& Ground & \multicolumn{2}{c|}{3DGS~\cite{Kerbl20233DGaussian}} & \multicolumn{2}{c}{\methodname ($N=1$)} \\
&&&&{\color{white}.} \\ 
& Truth  &  ($N=1$) & ($N=8$)      & ($K=3$)       & ($K=6$) \\
    \midrule
& \includegraphics[width=.090\textwidth]{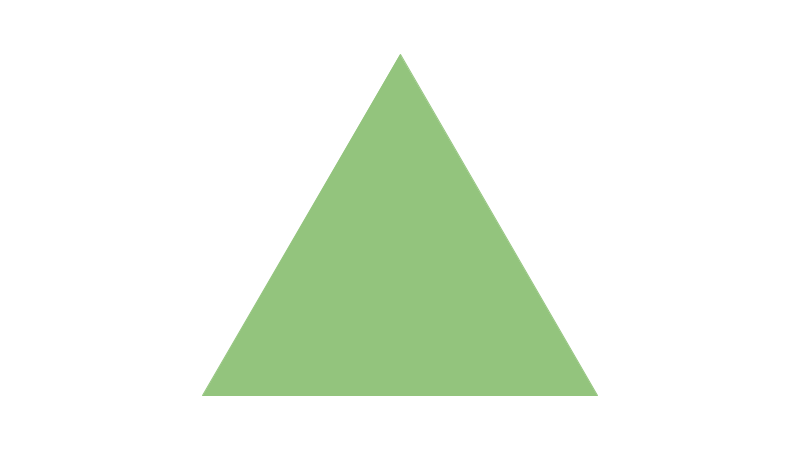}
& \includegraphics[width=.090\textwidth]{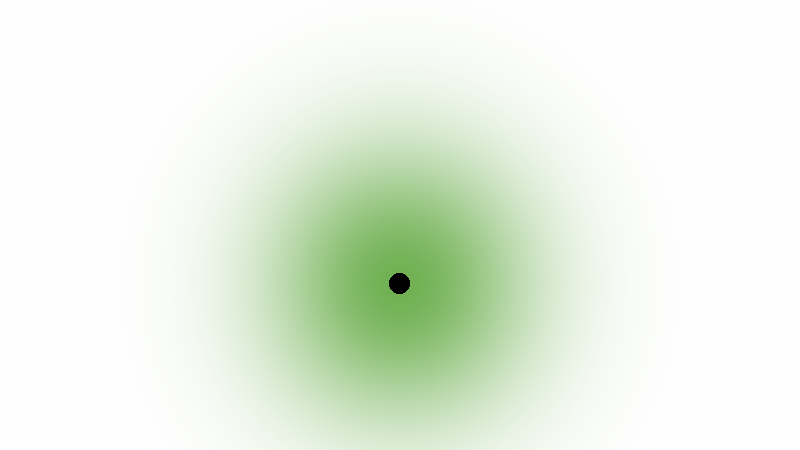}
& \includegraphics[width=.090\textwidth]{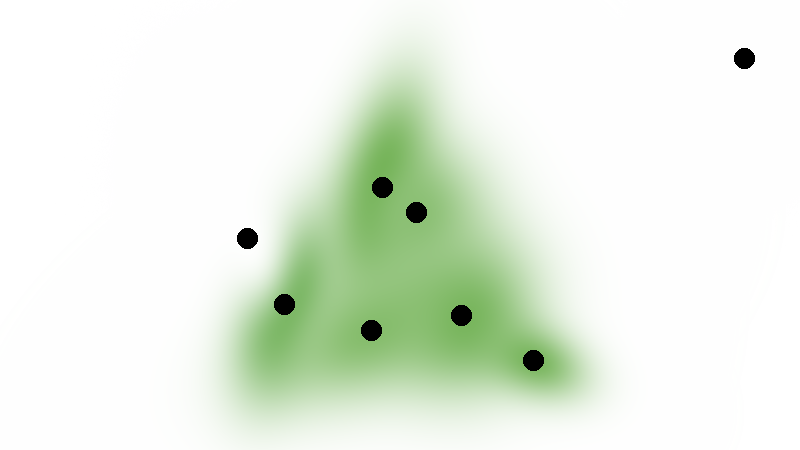}
& \includegraphics[width=.090\textwidth]{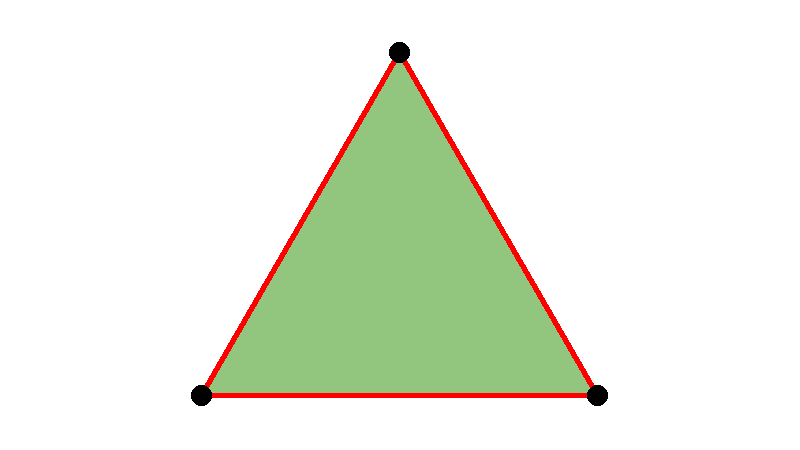}
& \includegraphics[width=.090\textwidth]{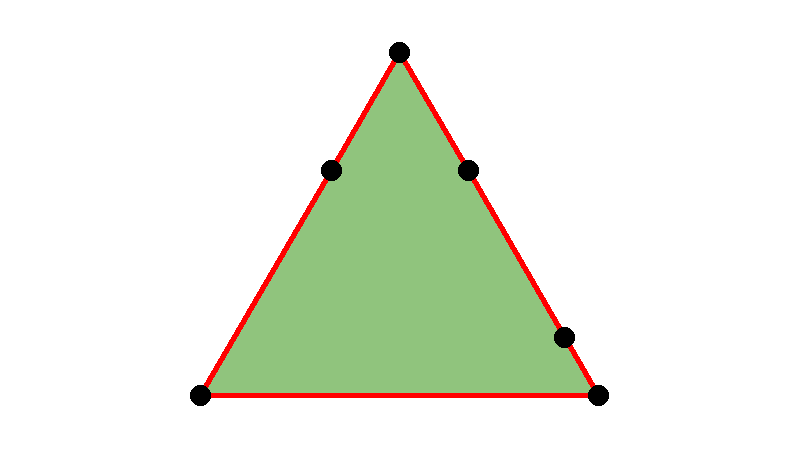}
    \\
    \hline
& \includegraphics[width=.090\textwidth]{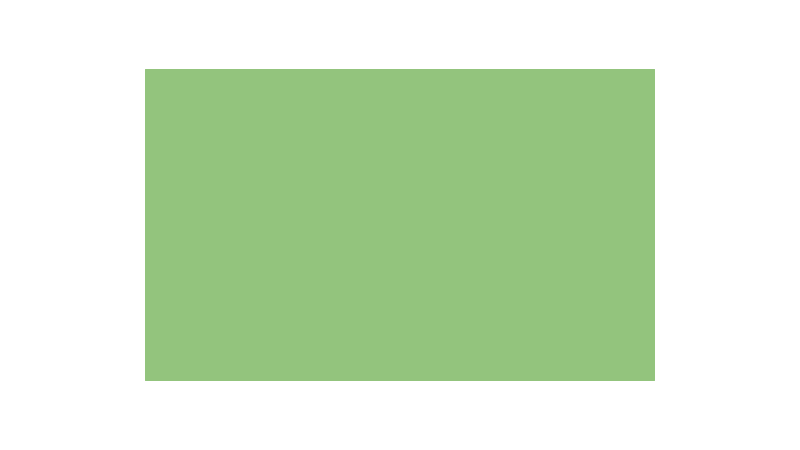}
& \includegraphics[width=.090\textwidth]{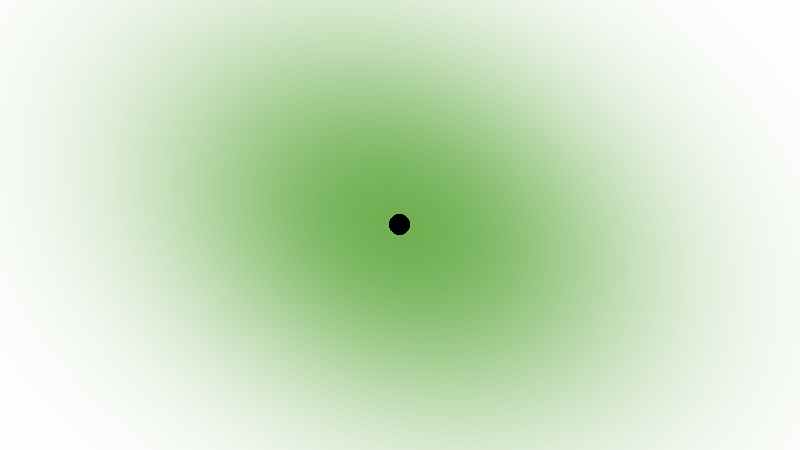}
& \includegraphics[width=.090\textwidth]{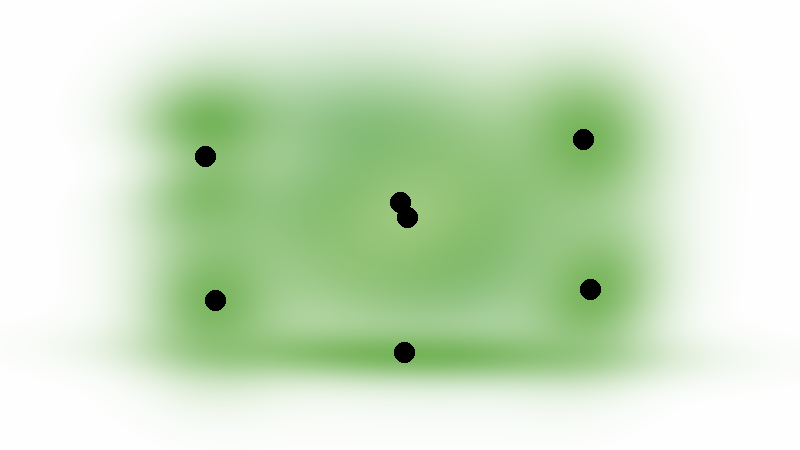}
& \includegraphics[width=.090\textwidth]{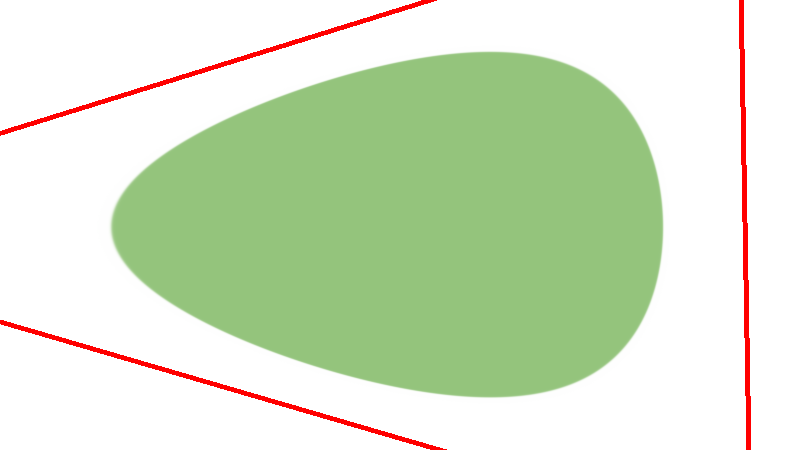}
& \includegraphics[width=.090\textwidth]{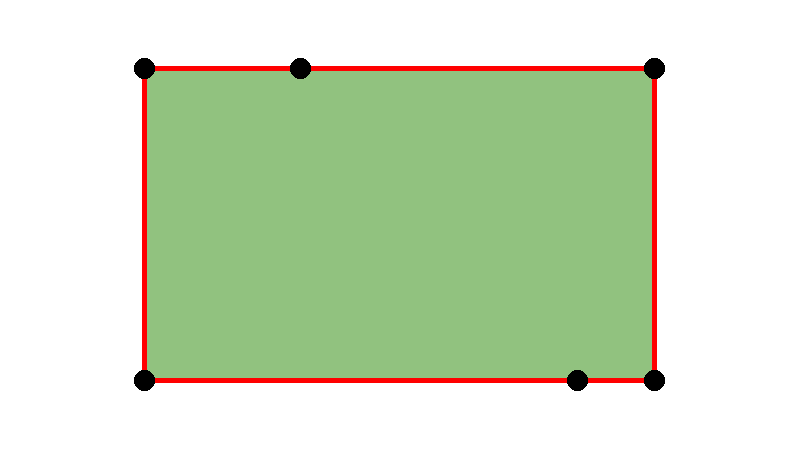}
    \\
    \hline
& \includegraphics[width=.090\textwidth]{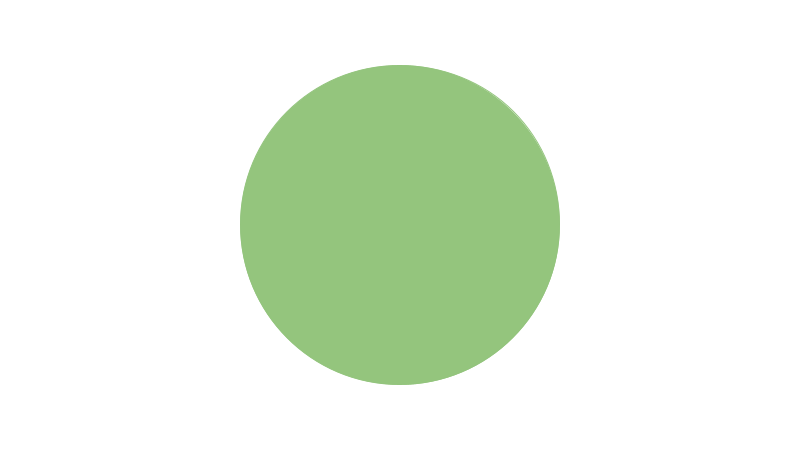}
& \includegraphics[width=.090\textwidth]{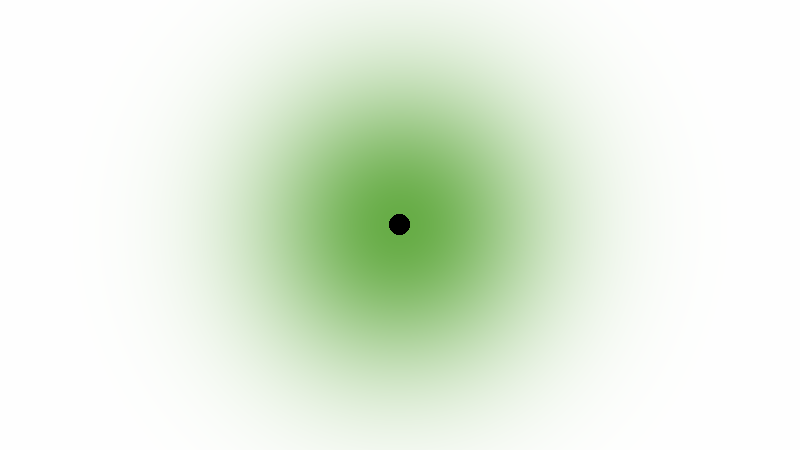}
& \includegraphics[width=.090\textwidth]{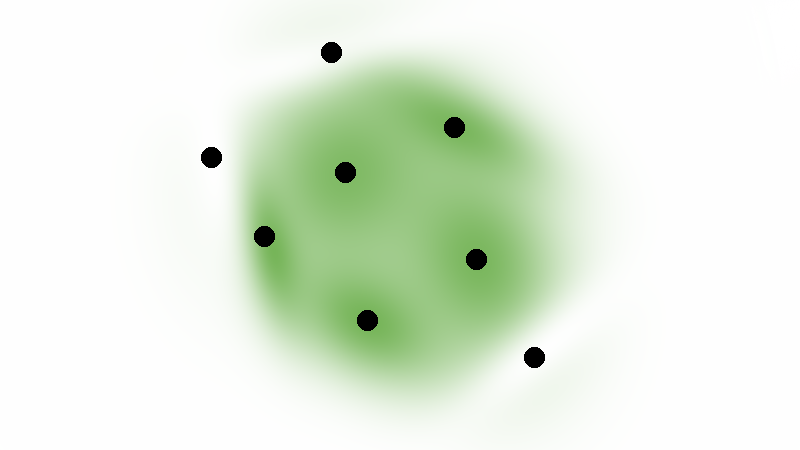}
& \includegraphics[width=.090\textwidth]{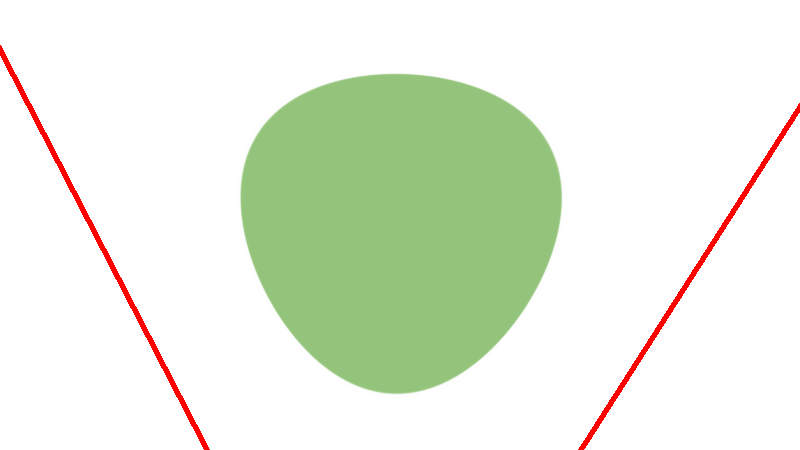}
& \includegraphics[width=.090\textwidth]{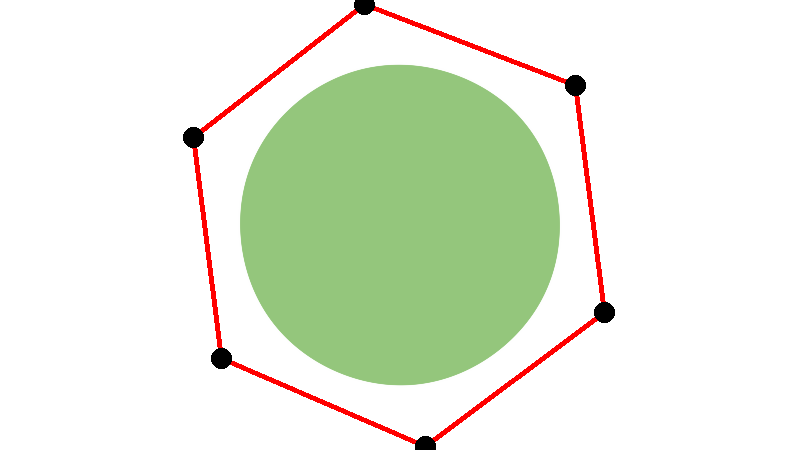}
    \\
    \hline
& \includegraphics[width=.090\textwidth]{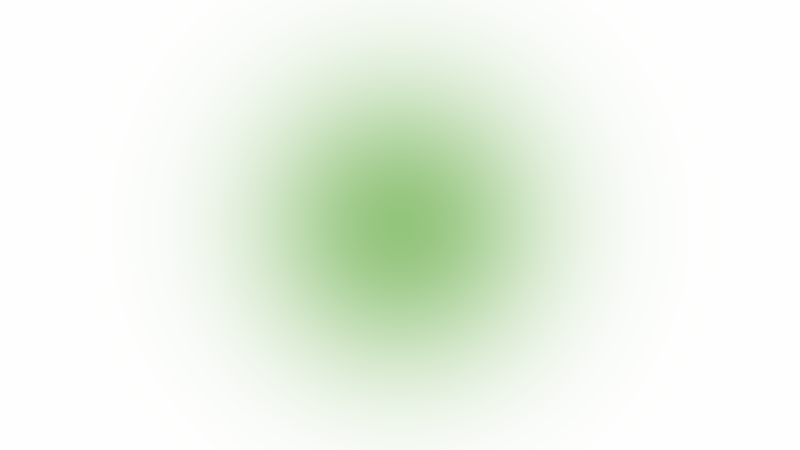}
& \includegraphics[width=.090\textwidth]{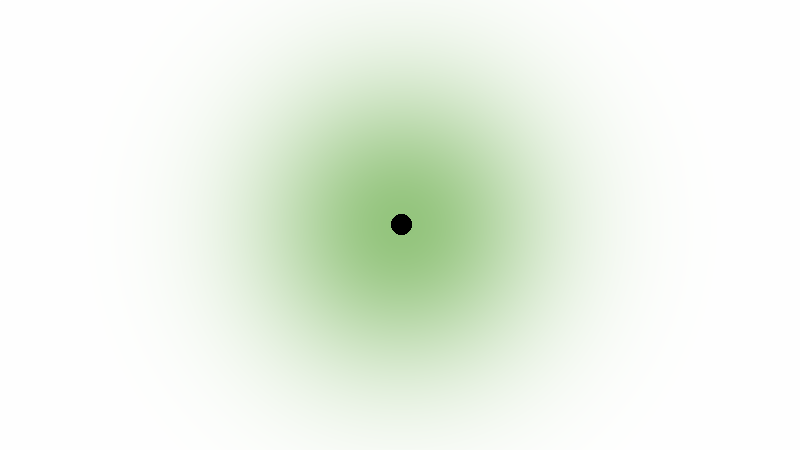}
& \includegraphics[width=.090\textwidth]{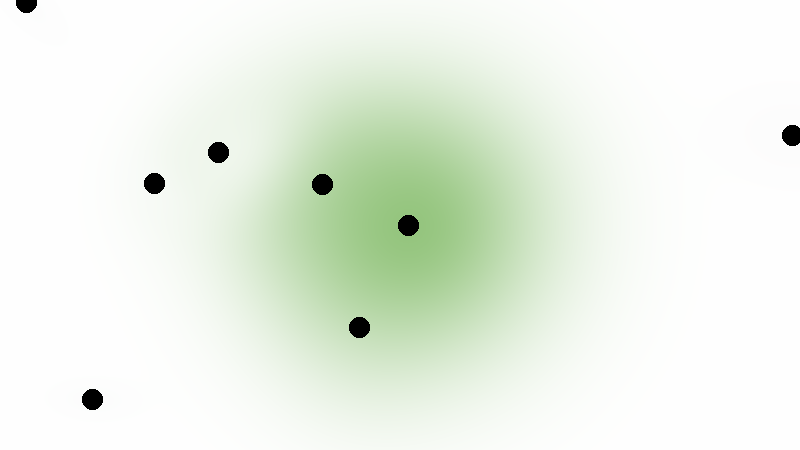}
& \includegraphics[width=.090\textwidth]{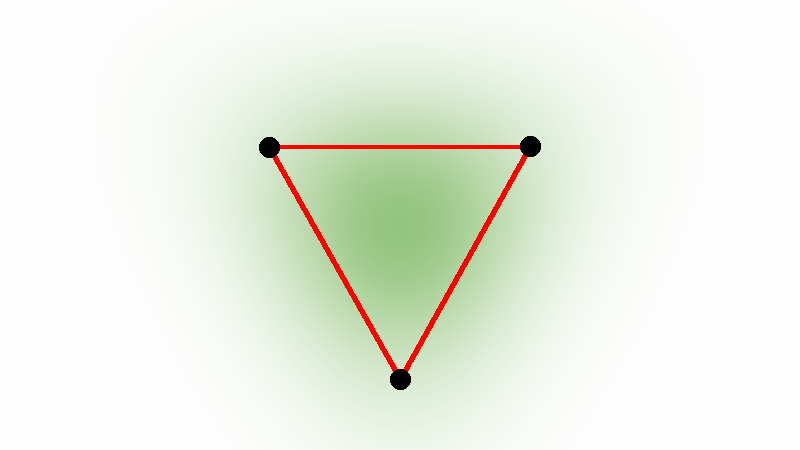}
& \includegraphics[width=.090\textwidth]{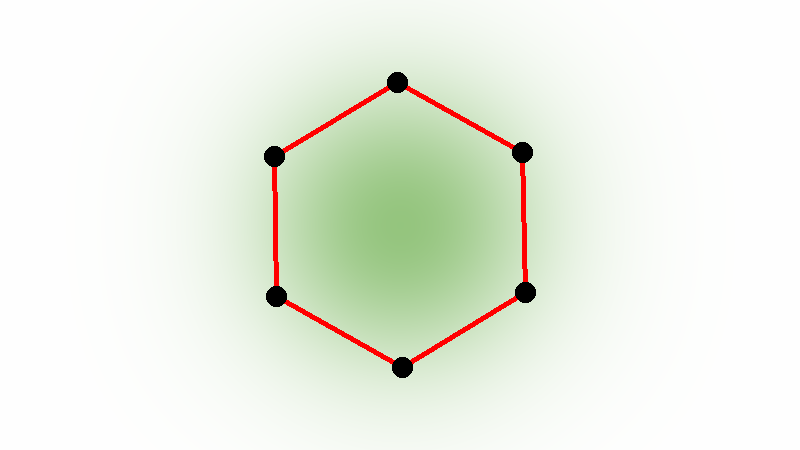}
    \\
    \hline
& \includegraphics[width=.090\textwidth]{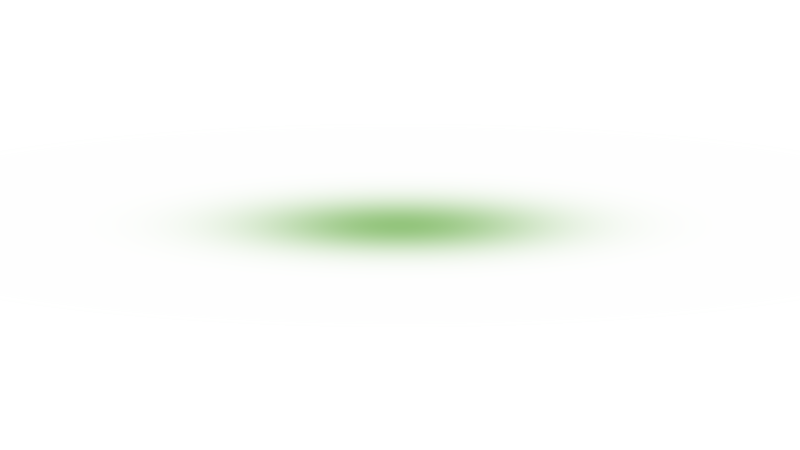}
& \includegraphics[width=.090\textwidth]{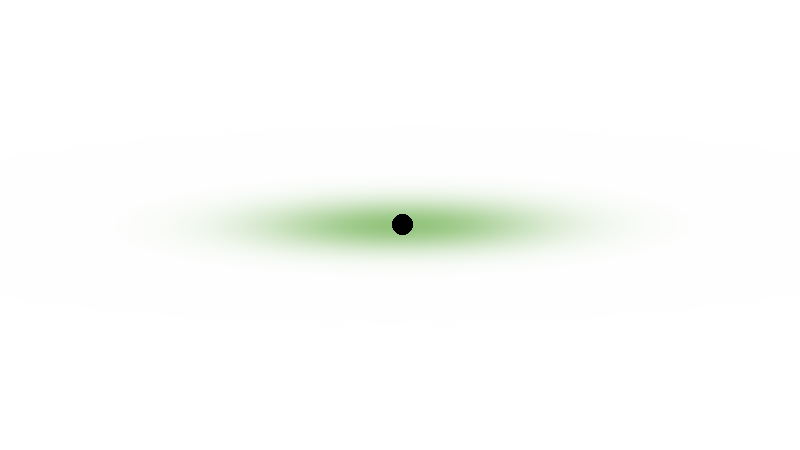}
& \includegraphics[width=.090\textwidth]{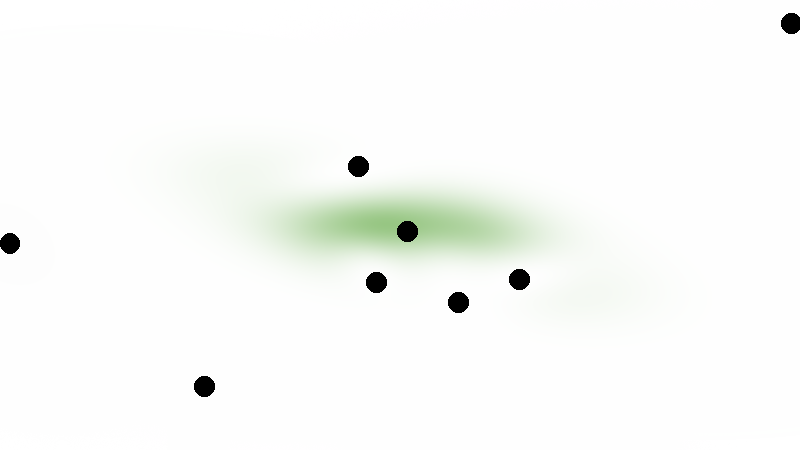}
& \includegraphics[width=.090\textwidth]{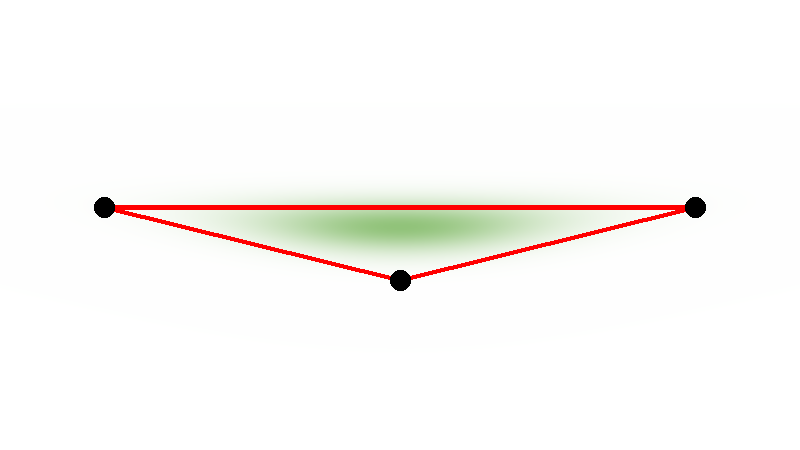}
& \includegraphics[width=.090\textwidth]{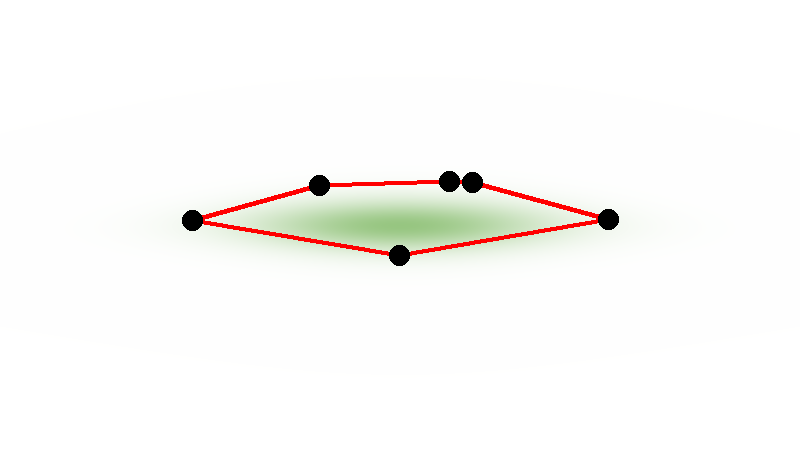}

\end{tabular}
\caption{
\textbf{Reconstruction of Simple Shapes with Primitives.}
Smooth convex primitives  reconstruct simple shapes better than Gaussians, as they can create sharper geometric boundaries. For 3DCS, the red lines describe the convex hull, whereas the black dots represent the point set. 
For 3DGS, the black dots represent the Gaussian centers. 
}
\label{fig:SyntheticExample1}
\end{figure}

\begin{table*}[ht!]
	\small
\resizebox{1.0\linewidth}{!}{
  \tabcolsep=0.07cm
\begin{tabular}{l|cccccc|cccccc|cccccc}
			
	Dataset & \multicolumn{6}{c|}{Tanks\&Temples}  & \multicolumn{6}{c|}{Deep Blending} & \multicolumn{6}{c}{Mip-NeRF360 Dataset}\\
	Method|Metric
	& $LPIPS^\downarrow$  & $PSNR^\uparrow$    & $SSIM^\uparrow$   & Train$^\downarrow$  & FPS$^\uparrow$ & Mem$^\downarrow$  
	& $LPIPS^\downarrow$  & $PSNR^\uparrow$    & $SSIM^\uparrow$   & Train$^\downarrow$ & FPS$^\uparrow$  & Mem$^\downarrow$  
	& $LPIPS^\downarrow$  & $PSNR^\uparrow$    & $SSIM^\uparrow$   & Train$^\downarrow$  & FPS$^\uparrow$ & Mem$^\downarrow$  \\
	\midrule 
	Mip-NeRF360\cite{Barron2022MipNeRF360} & 0.257  &   22.22 &  0.759 & 48h & 0.14 &  8.6MB &   0.245 &   29.40 &   0.901 & 48h & 0.09 &  8.6MB & 0.237 & \best 27.69 & 0.792 & 48h & 0.06 &  8.6MB\\
 3DGS\cite{Kerbl20233DGaussian} & 0.183 & 23.14 & 0.841 & 26m &  154 & 411MB & \sbest 0.243 & 29.41   & \best 0.903 & 36m &  137 & 676MB & \sbest 0.214 & 27.21 & \best 0.815 & 42m &  134 & 734MB\\ 
	GES \cite{Hamdi2024GES} &   0.198   & 23.35     &  0.836   &  21m  &  210  &  222MB  &  0.252  &  \sbest29.68  & 0.901   &  30m &  160 &   399MB &  0.250  &   26.91  &    0.794 &  32m  &  186  &  377MB  \\
 2DGS\cite{Huang20242DGaussian}$\dagger$ & 0.212 & 23.13 & 0.831 & 14m & 122 & 200MB & 0.257 & 29.50 & 0.902 & 28m & 76 & 353MB & 0.252 & 27.18 & \sbest 0.808 & 29m & 64 & 484MB\\
\midrule

  \textbf{\methodname (light)} & \sbest 0.170   & \sbest 23.71 & \sbest 0.842 &   46m & 40  &  83 MB & 
   0.245  & 29.61 & 0.901  & 84m  & 30  & 110 MB & 0.266 & 26.66  & 0.769 & 53m & 47  & 77MB \\

 \textbf{\methodname} & \best 0.157  & \best 23.95 & \best 0.851 &  60m   & 33 &  282MB
 & \best 0.237   & \best 29.81 & \sbest 0.902  & 71m  & 30 & 332 MB & \best 0.207 & \sbest 27.29   &  0.802 & 87m   & 25   & 666MB \\
\end{tabular}
	}
\caption{\myTitle{Comparative Analysis of Novel View Synthesis Methods.} We conduct a quantitative comparison of our \methodname method against other primitive rendering approaches across three datasets. \methodname achieves higher-quality results in novel view synthesis with reduced memory consumption, all while achieving fast rendering performance. No codebooks or post-processing compression \cite{Lee2024Compact, Niedermayr2024Compressed} are used to reduce the size of any of the methods. The best performances are shown in {\best red} and the second best in {\best orange}. $\dagger$ indicates reproduced results.}
 \label{tab:comparisons}
\end{table*}

\section{Experiments}%
\label{sec:results}

We first evaluate 3D Convex Splatting (\methodname) with synthetic experiments to showcase its superior shape representation over Gaussian primitives. 
Then, we present the real-world setup, followed by results and an ablation study.

\subsection{Experiments on Synthetic Data}
\Cref{fig:SyntheticExample1} compares the representation capabilities of using $1$ or $8$  Gaussian primitives  against using a single convex shape defined by $3$ or $6$ points.
The results demonstrate that smooth convexes effectively approximate a wide range of shapes, including both polyhedra and Gaussians, while requiring fewer primitives for accurate representation.

\subsection{Experimental Setup}%
\label{subsec:real_world_exp}

\mysection{Datasets.}
To evaluate \methodname on real-world novel view synthesis, we use the same datasets as 3DGS~\cite{Kerbl20233DGaussian}. This includes two scenes from Deep Blending (DB)~\cite{Hedman2018Deep}, two scenes from Tanks and Temples (T\&T)~\cite{Knapitsch2017Tanks}, and all scenes from the Mip-NeRF360 dataset~\cite{Barron2022MipNeRF360}.

\mysection{Baselines.}
We compare our 3DCS method with three other primitives for novel-view synthesis: 3D Gaussians~\cite{Kerbl20233DGaussian}, Generalized Exponential Functions (GES)~\cite{Hamdi2024GES}, and 2D Gaussians~\cite{Huang20242DGaussian}.
While many follow-up studies have built upon 3DGS and introduced various enhancements, we focus on the basic primitives for comparison.
This choice ensures that the evaluation is based on the core principles of each approach.
Furthermore, we evaluate our method against the Mip-NeRF360 method in \cite{Barron2022MipNeRF360}.

\mysection{Metrics.}
We use common metrics in the novel view synthesis literature, such as SSIM, PSNR, and LPIPS, to assess the visual quality of the synthesized images. Furthermore, we report the average training time, rendering speed, and memory usage.
This allows for a thorough comparison between \methodname and the other methods.

\mysection{Implementation details of \methodname.}
For each experiment, we initialize the number of points per convex shape to $\pointsPerConvex=6$ and  a spherical harmonic degree of 3, resulting in a total of 69 parameters per convex shape. For each 3D Gaussian, 59 parameters are needed. In the following sections, we compare two variants of \methodname: a best-performing model and a lightweight variant.
Our best model employs different hyperparameters for indoor and outdoor scenes, whereas the lightweight model uses a unified set of parameters.
For our light version, we increase the threshold criterion for densifying convexes, effectively reducing the number of shapes.
Furthermore, we store the 3D convex parameters with 32-bit precision for the high-quality model and 16-bit precision for the lightweight version.
A list of hyperparameters can be found in the \supp.
Notably, no compression methods are applied to reduce memory usage.

\begin{figure*}[t]
\centering
\setlength\mytmplen{0.23\linewidth}
\resizebox{\linewidth}{!}{ 

\begin{tabular}{c@{\hskip 0.2in}c@{\hskip 0.2in}c@{\hskip 0.2in}c}
    
    \makebox[\mytmplen]{Ground Truth} &
    \makebox[\mytmplen]{\textbf{\methodname (ours)}} &
    \makebox[\mytmplen]{3DGS} &
    \makebox[\mytmplen]{2DGS} \\

    \rotatebox{90}{\parbox{2.2cm}{\centering Flowers}}
    \zoomin{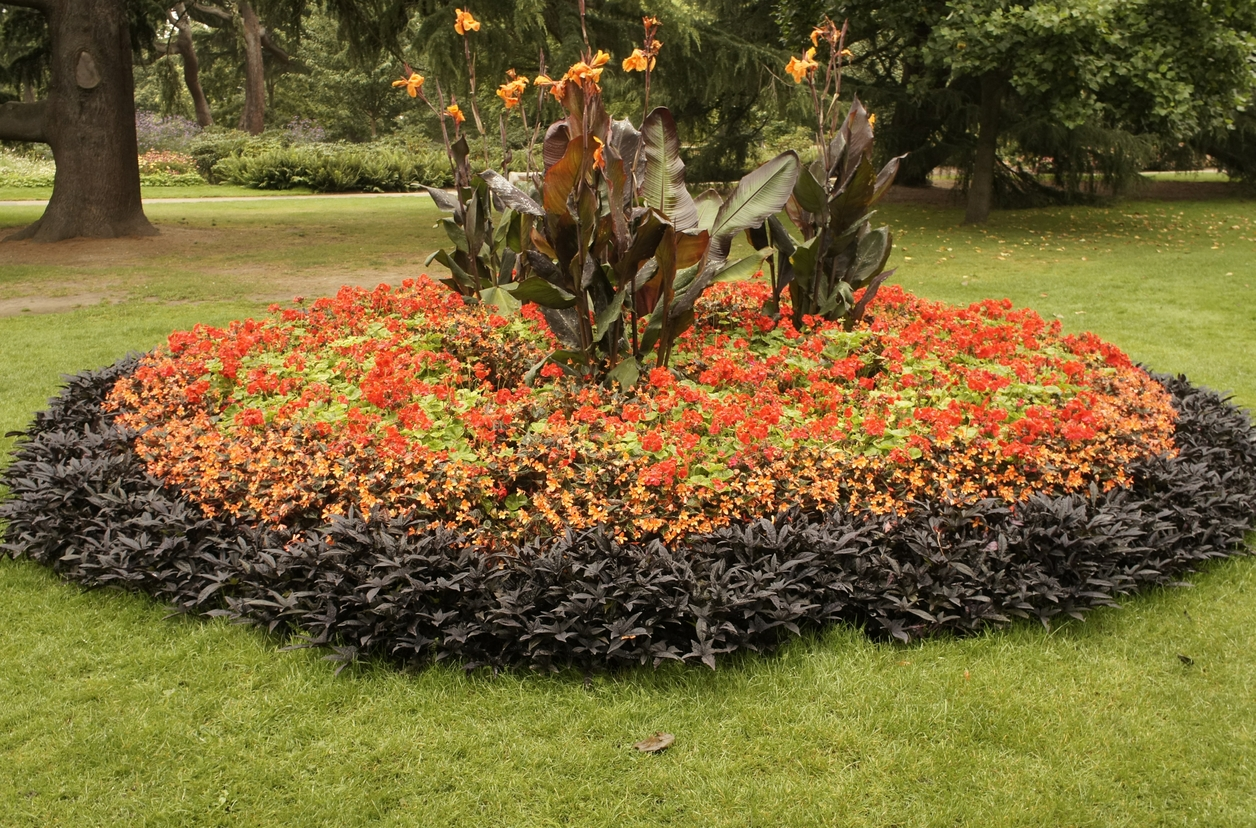}{0.55\mytmplen}{0.09\mytmplen}{0.16\mytmplen}{0.16\mytmplen}{1.2cm}{\mytmplen}{2.5}{red} &
    \zoomin{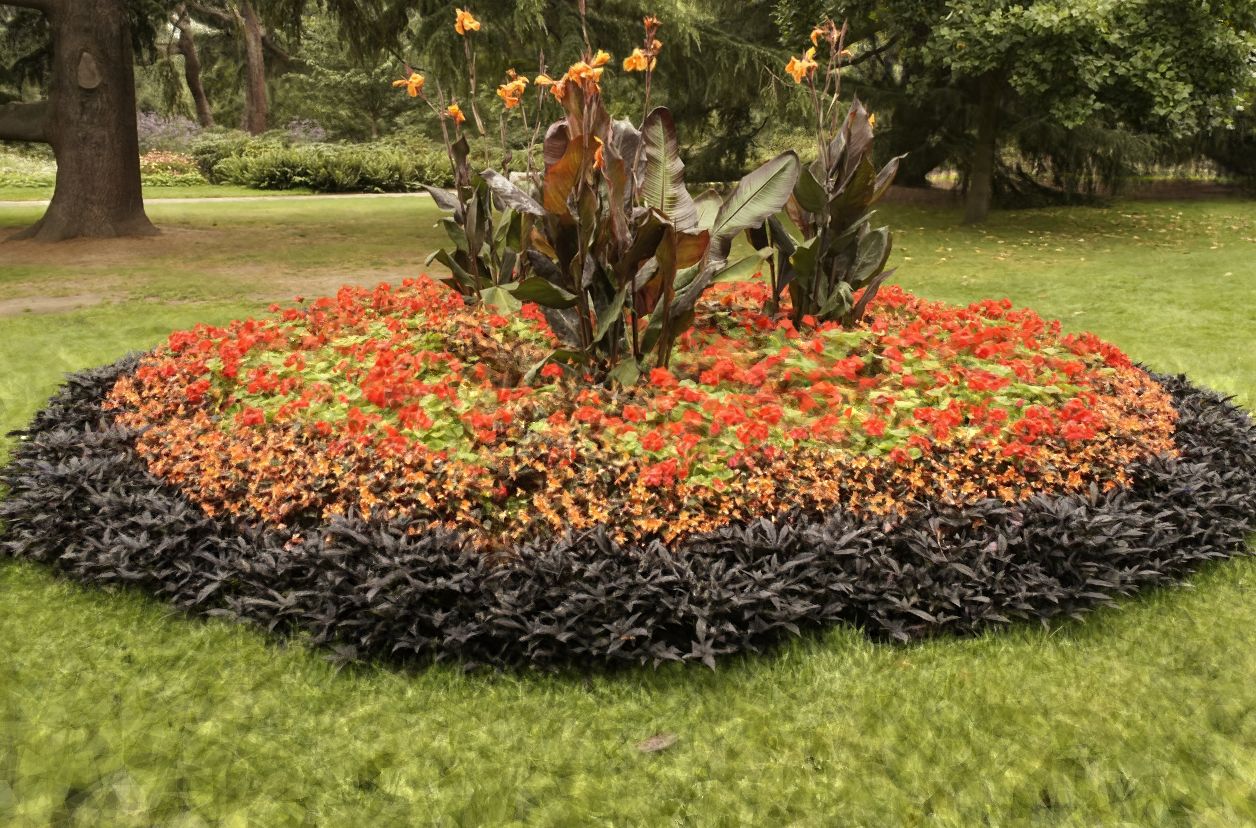}{0.55\mytmplen}{0.09\mytmplen}{0.16\mytmplen}{0.16\mytmplen}{1.2cm}{\mytmplen}{2.5}{red} &
    \zoomin{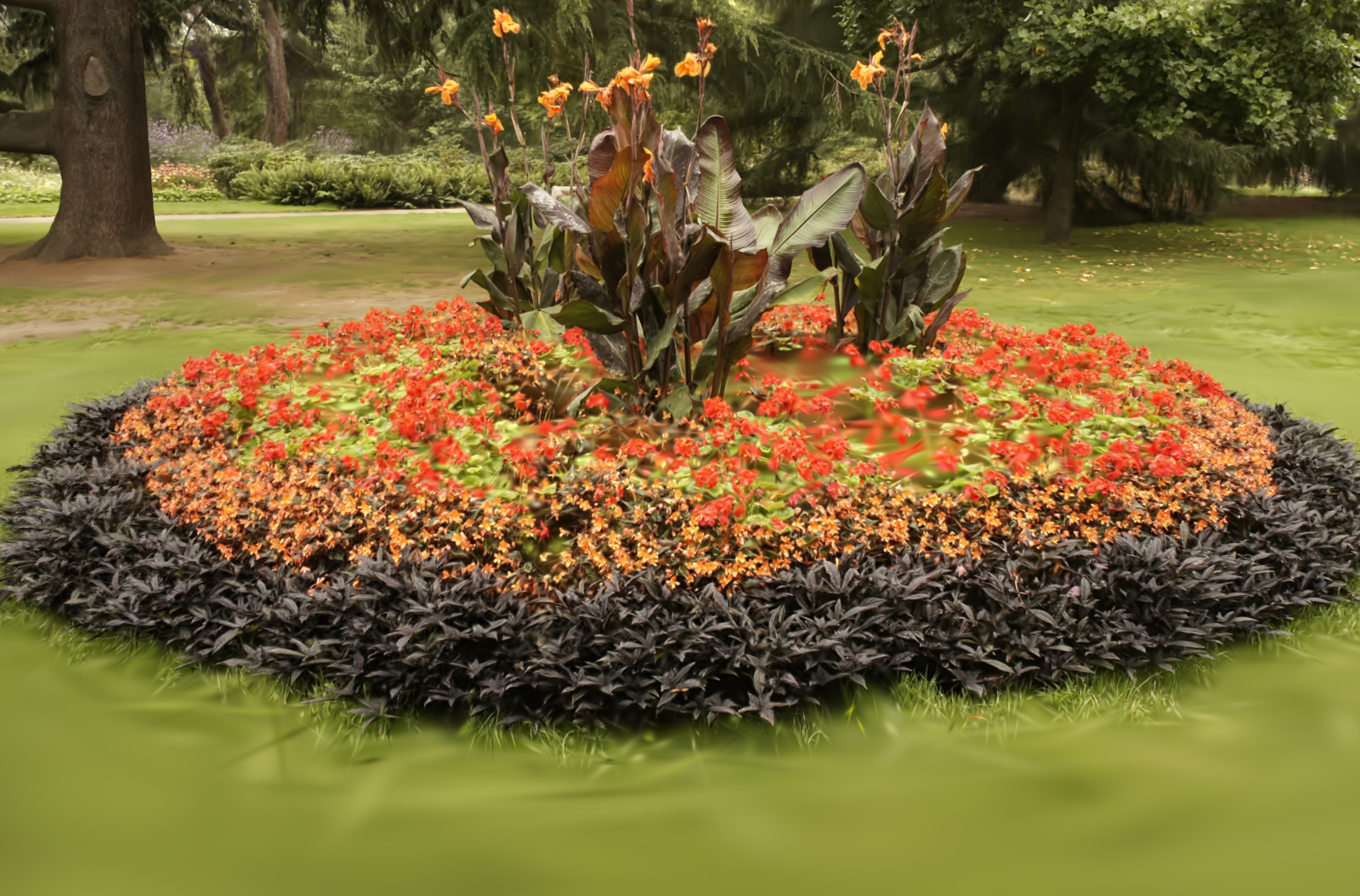}{0.55\mytmplen}{0.09\mytmplen}{0.16\mytmplen}{0.16\mytmplen}{1.2cm}{\mytmplen}{2.5}{red} &
    \zoomin{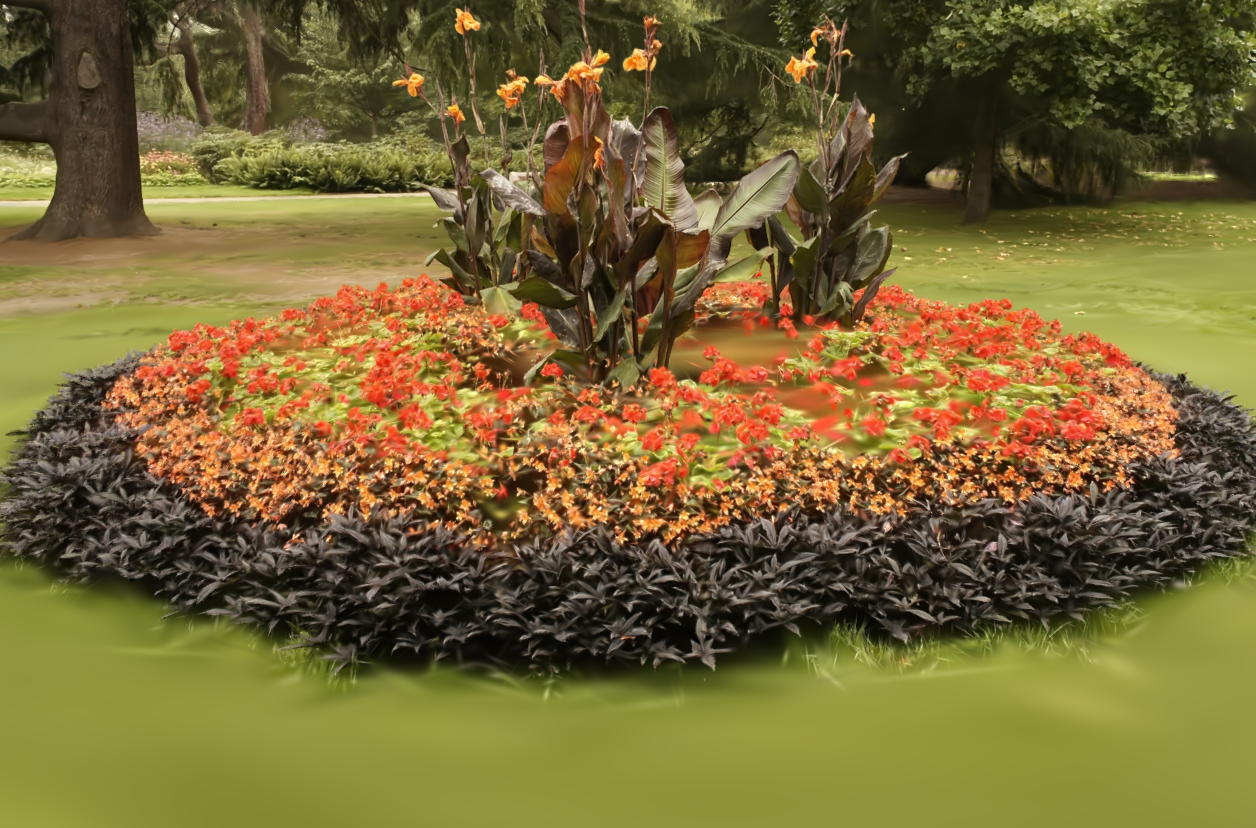}{0.55\mytmplen}{0.09\mytmplen}{0.16\mytmplen}{0.16\mytmplen}{1.2cm}{\mytmplen}{2.5}{red} \\

    \rotatebox{90}{\parbox{2.2cm}{\centering Bicycle}}
    \zoomin{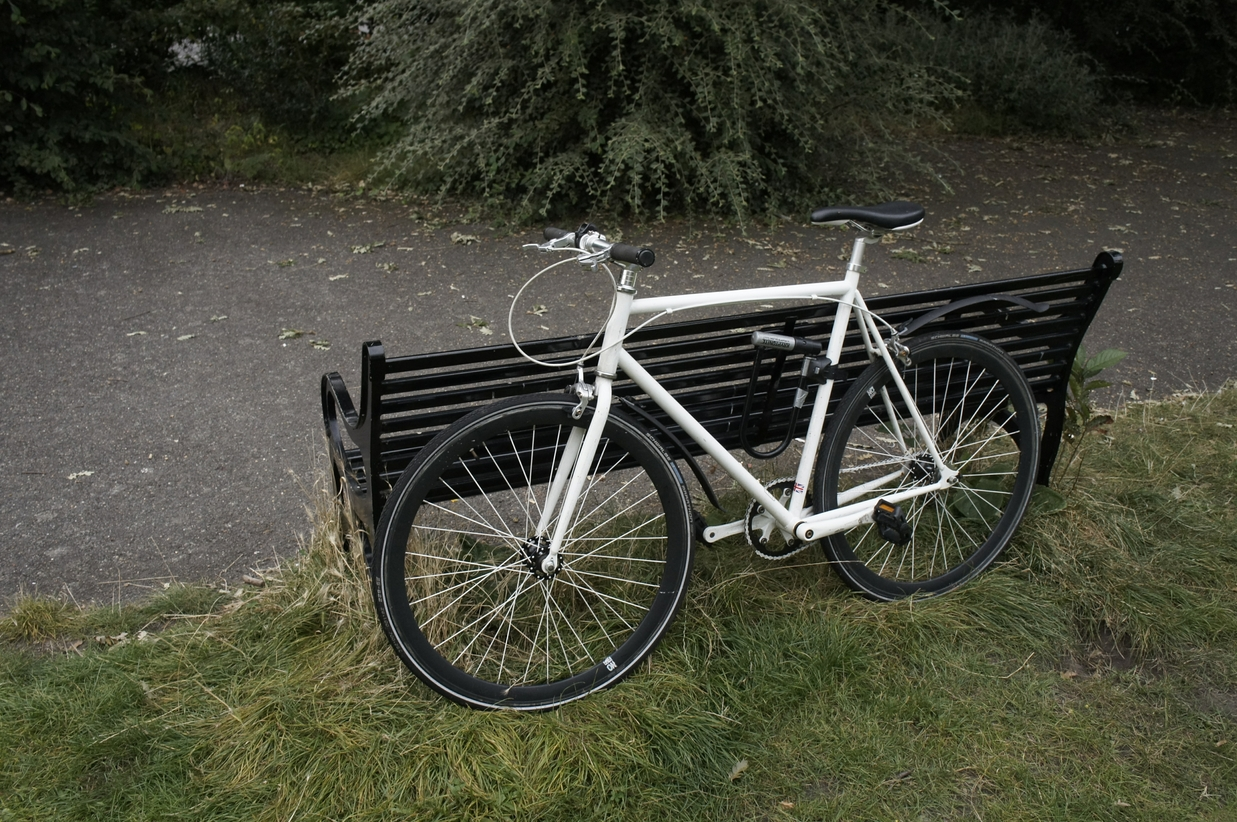}{0.13\mytmplen}{0.47\mytmplen}{0.16\mytmplen}{0.16\mytmplen}{1.2cm}{\mytmplen}{3.5}{red} &
    \zoomin{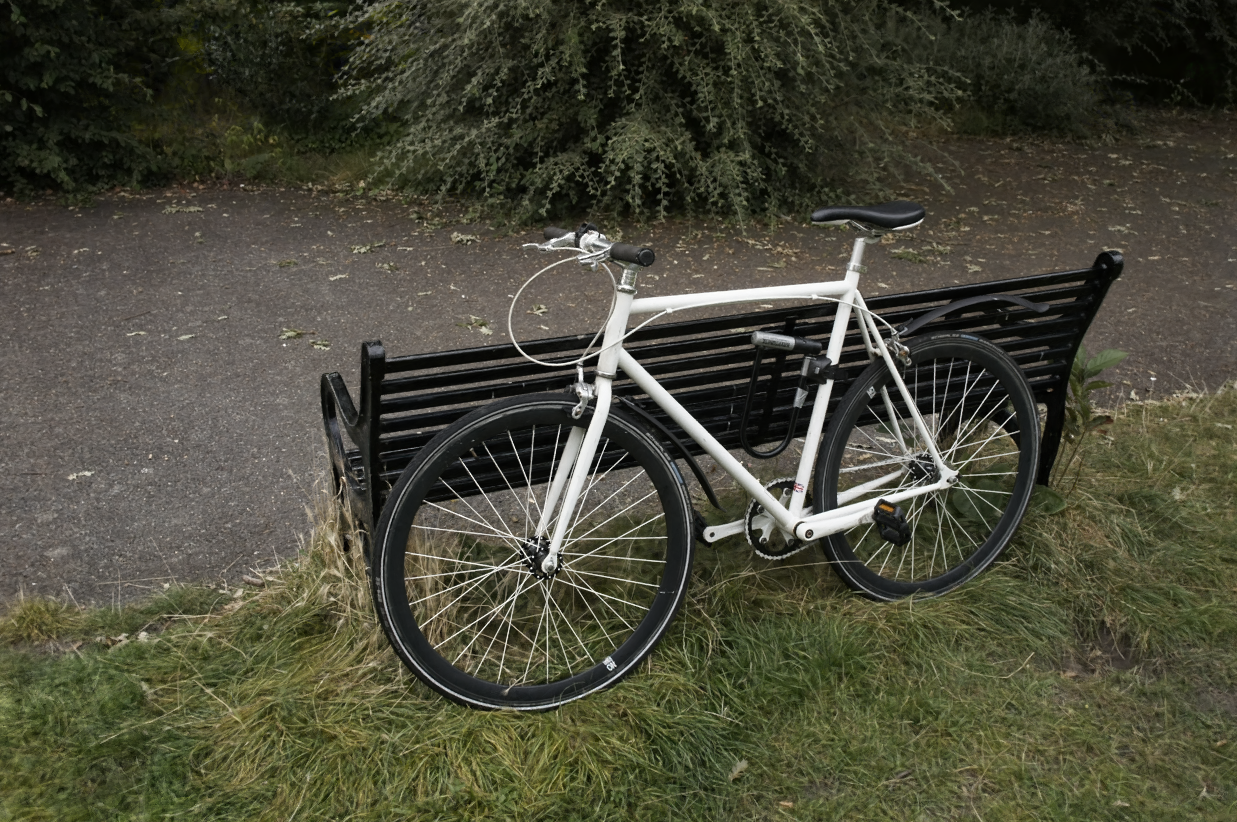}{0.13\mytmplen}{0.47\mytmplen}{0.16\mytmplen}{0.16\mytmplen}{1.2cm}{\mytmplen}{3.5}{red} &
    \zoomin{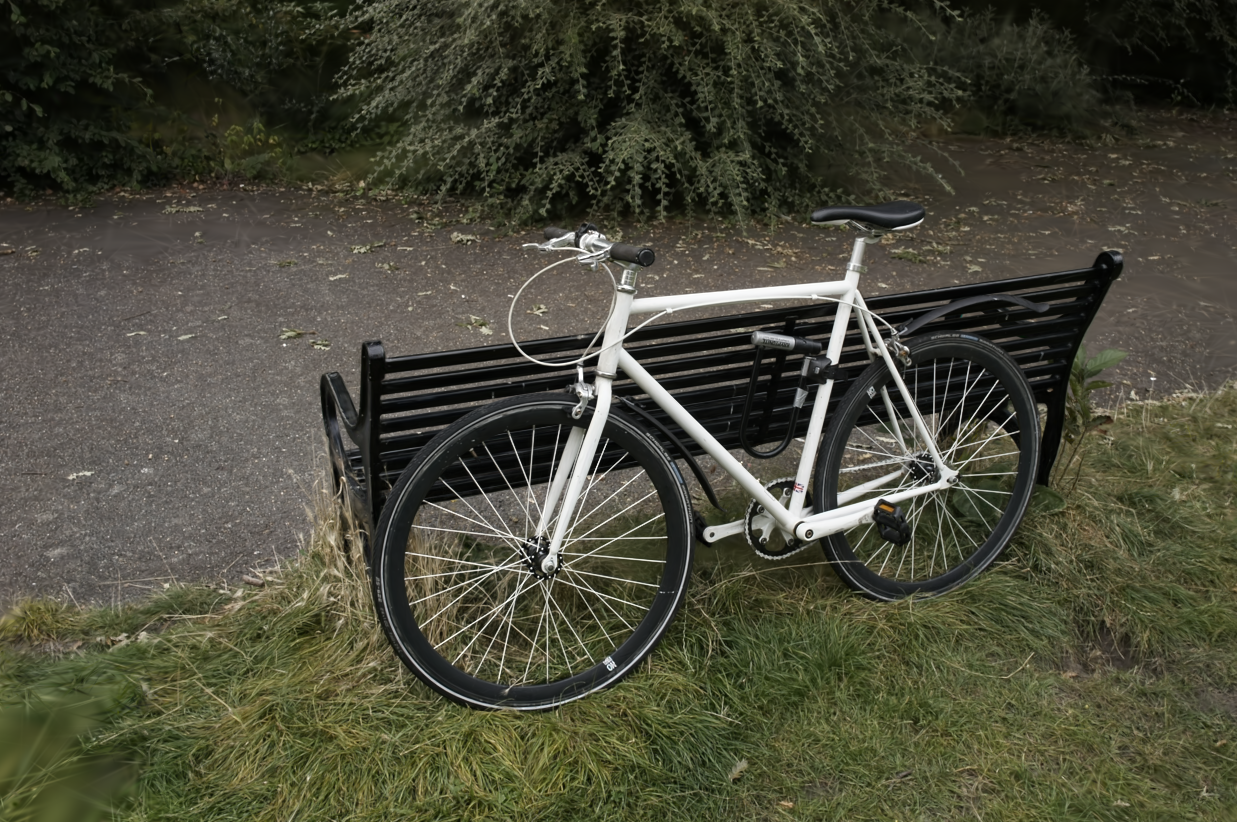}{0.13\mytmplen}{0.47\mytmplen}{0.16\mytmplen}{0.16\mytmplen}{1.2cm}{\mytmplen}{3.5}{red} &
    \zoomin{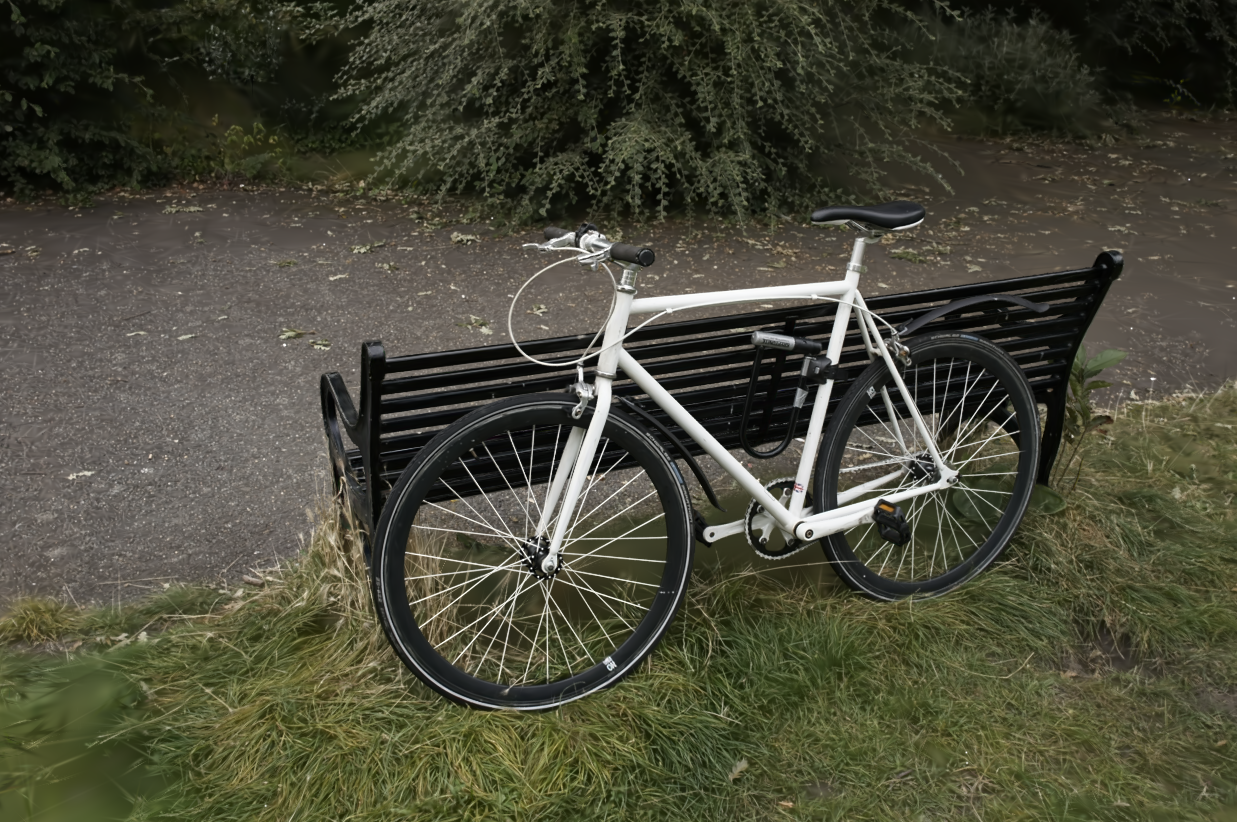}{0.13\mytmplen}{0.47\mytmplen}{0.16\mytmplen}{0.16\mytmplen}{1.2cm}{\mytmplen}{3.5}{red}\\

    \rotatebox{90}{\parbox{2.2cm}{\centering Train}}
    \zoomin{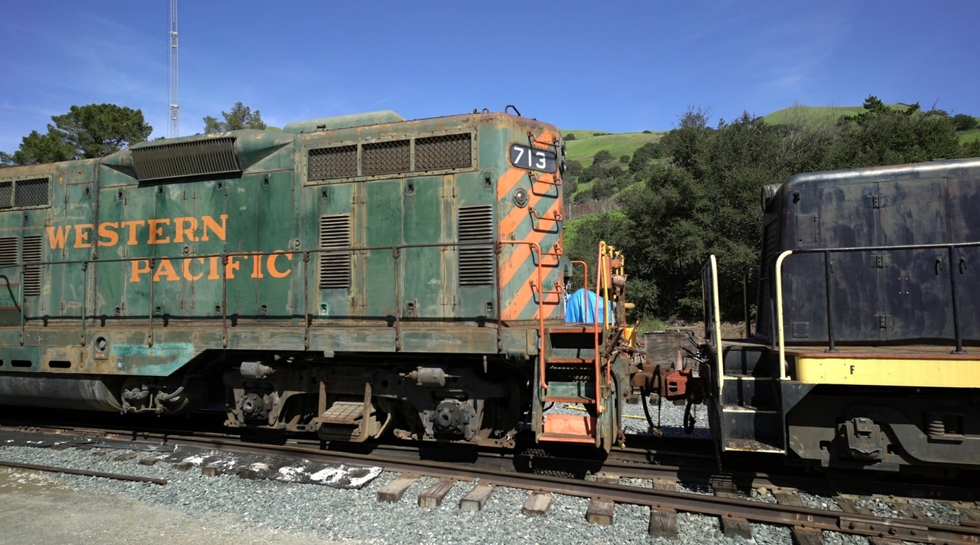}{0.65\mytmplen}{0.4\mytmplen}{0.16\mytmplen}{0.16\mytmplen}{1.2cm}{\mytmplen}{2.5}{red} &
    \zoomin{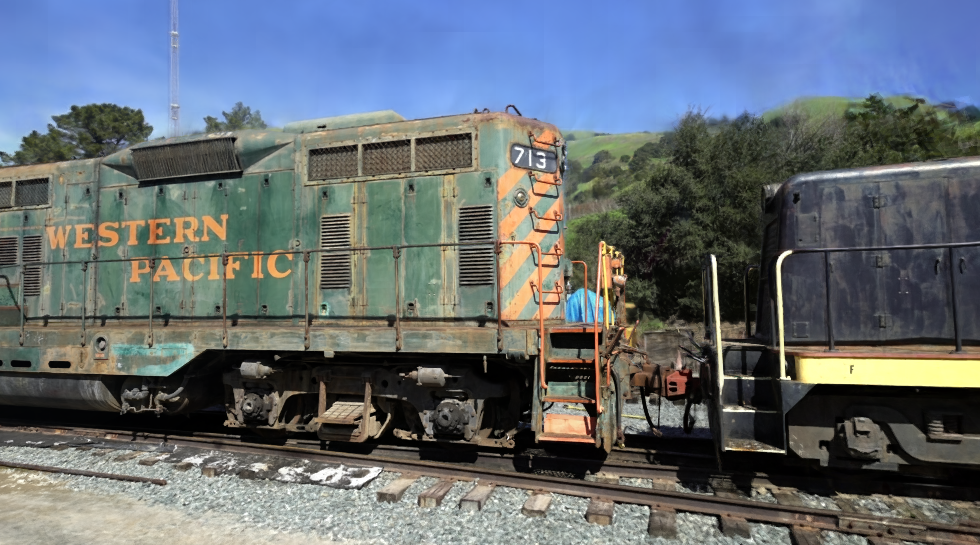}{0.65\mytmplen}{0.4\mytmplen}{0.16\mytmplen}{0.16\mytmplen}{1.2cm}{\mytmplen}{2.5}{red} &
    \zoomin{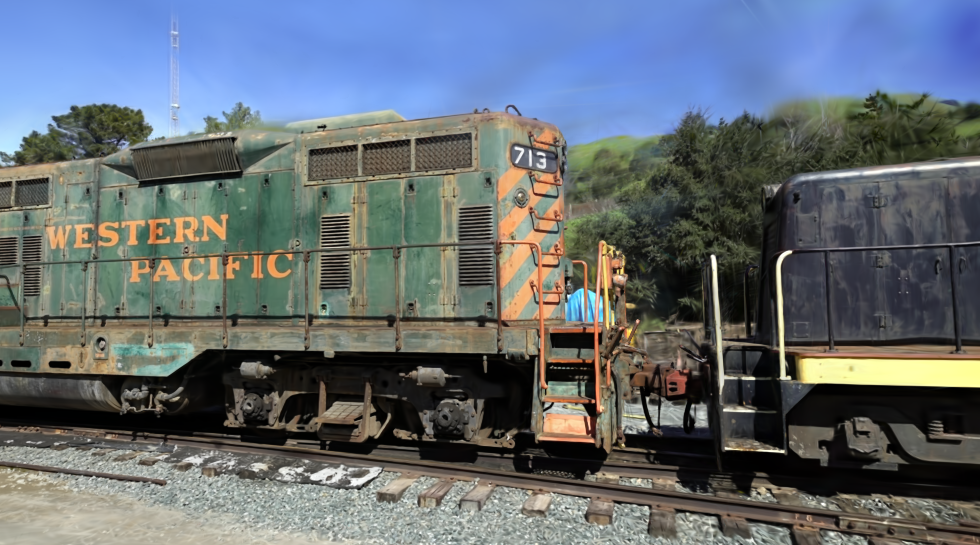}{0.65\mytmplen}{0.4\mytmplen}{0.16\mytmplen}{0.16\mytmplen}{1.2cm}{\mytmplen}{2.5}{red} &
    \zoomin{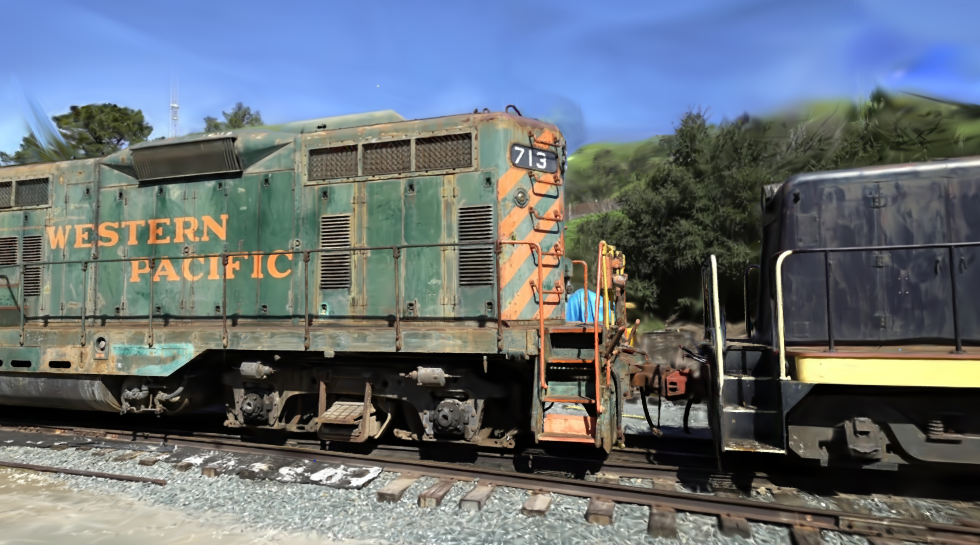}{0.65\mytmplen}{0.4\mytmplen}{0.16\mytmplen}{0.16\mytmplen}{1.2cm}{\mytmplen}{2.5}{red} \\

    \rotatebox{90}{\parbox{2.2cm}{\centering Truck}}
    \zoomin{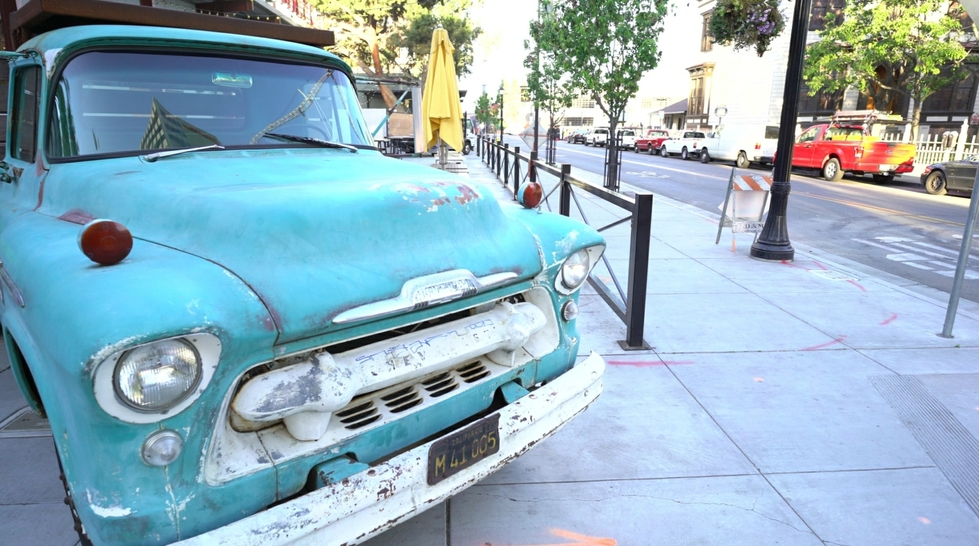}{0.73\mytmplen}{0.32\mytmplen}{0.875\mytmplen}{0.13\mytmplen}{1.0cm}{\mytmplen}{2.5}{red} &
    \zoomin{Images/figure7/truck2cvx.png}{0.73\mytmplen}{0.32\mytmplen}{0.875\mytmplen}{0.13\mytmplen}{1.0cm}{\mytmplen}{2.5}{red} &
    \zoomin{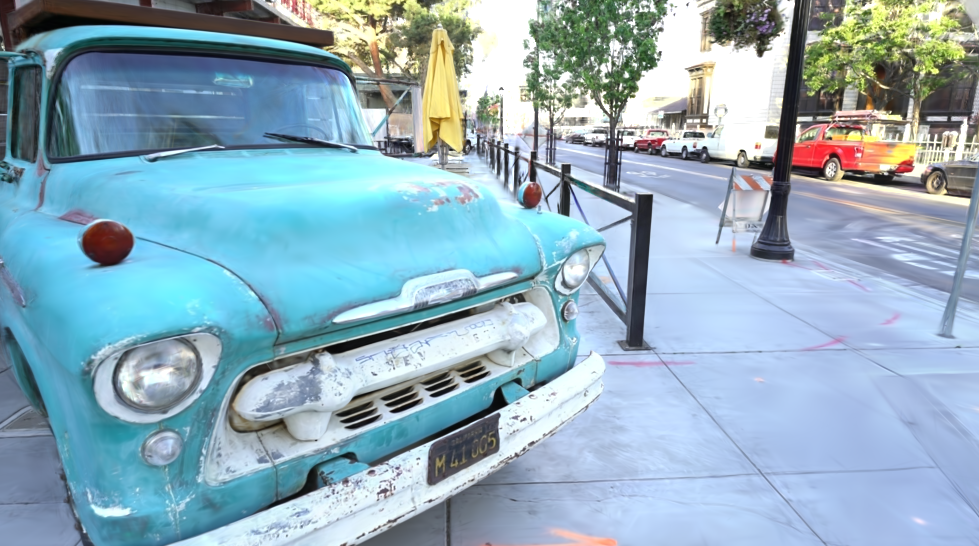}{0.73\mytmplen}{0.32\mytmplen}{0.875\mytmplen}{0.13\mytmplen}{1.0cm}{\mytmplen}{2.5}{red} &
    \zoomin{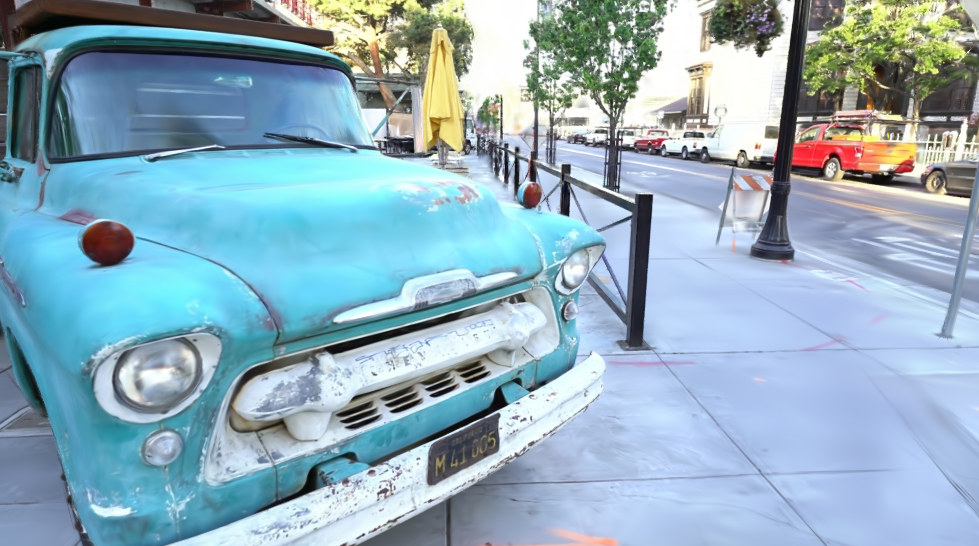}{0.73\mytmplen}{0.32\mytmplen}{0.875\mytmplen}{0.13\mytmplen}{1.0cm}{\mytmplen}{2.5}{red} \\

\end{tabular}
}

\caption{
\myTitle{Qualitative Comparison between \methodname, 3DGS and 2DGS.} 
Our \methodname captures finer details and provides a more accurate approximation of real-world scenes compared to Gaussian splatting methods, which often produce blurrier results.
}
\label{fig:qualityresults}
\end{figure*}

\subsection{Real-world Novel View Synthesis}%
\label{subsec:real-world-experiments}

\mysection{Main results.}
\Cref{tab:comparisons} presents the quantitative results.
As can be seen, our \methodname method consistently matches or surpasses the rendering quality of existing methods across all evaluated datasets.
Specifically, \methodname outperforms 3DGS, GES and 2DGS in most metrics on the T\&T and DB datasets, while also achieving the second highest PSNR and lowest LPIPS on the Mip-NeRF360 dataset.
\methodname effectively balances memory usage and training time, sitting in between the ones of Mip-NeRF360 and 3DGS.
Particularly, while it consumes more memory than Mip-NeRF360, it significantly reduces training time, requiring only $63$ minutes compared to the $48$ hours of Mip-NeRF360.
Moreover, it delivers better visual quality, especially on the T\&T dataset, where \methodname demonstrates a notable performance advantage of over $1.73$ PSNR compared to Mip-NeRF360.
In comparison with 3DGS, \methodname exhibits a slightly longer training time and lower rendering speed.
Yet, \methodname still operates within real-time rendering capabilities.
Thanks to its greater adaptability, \methodname efficiently utilizes only $70\%$ of the memory needed by 3DGS, while achieving higher visual quality.
\Cref{fig:qualityresults} strikes a qualitative comparison between \methodname, 3DGS and 2DGS.
Notably, our method achieves sharp and detailed rendering even in challenging regions, \textit{e.g.} the background in the \textit{Train} scene.
In contrast, Gaussian primitives tend to oversmooth areas, resulting in images with pronounced artifacts, as observed in the \textit{Flower}, \textit{Bicycle}, and \textit{Truck} scenes. The convex-based approach, however, produces results that closely align with the ground truth, showcasing higher fidelity and a superior ability to realistically represent 3D environments.

\begin{figure}[t]
\centering
\setlength\mytmplen{0.48\linewidth}
\zoomin{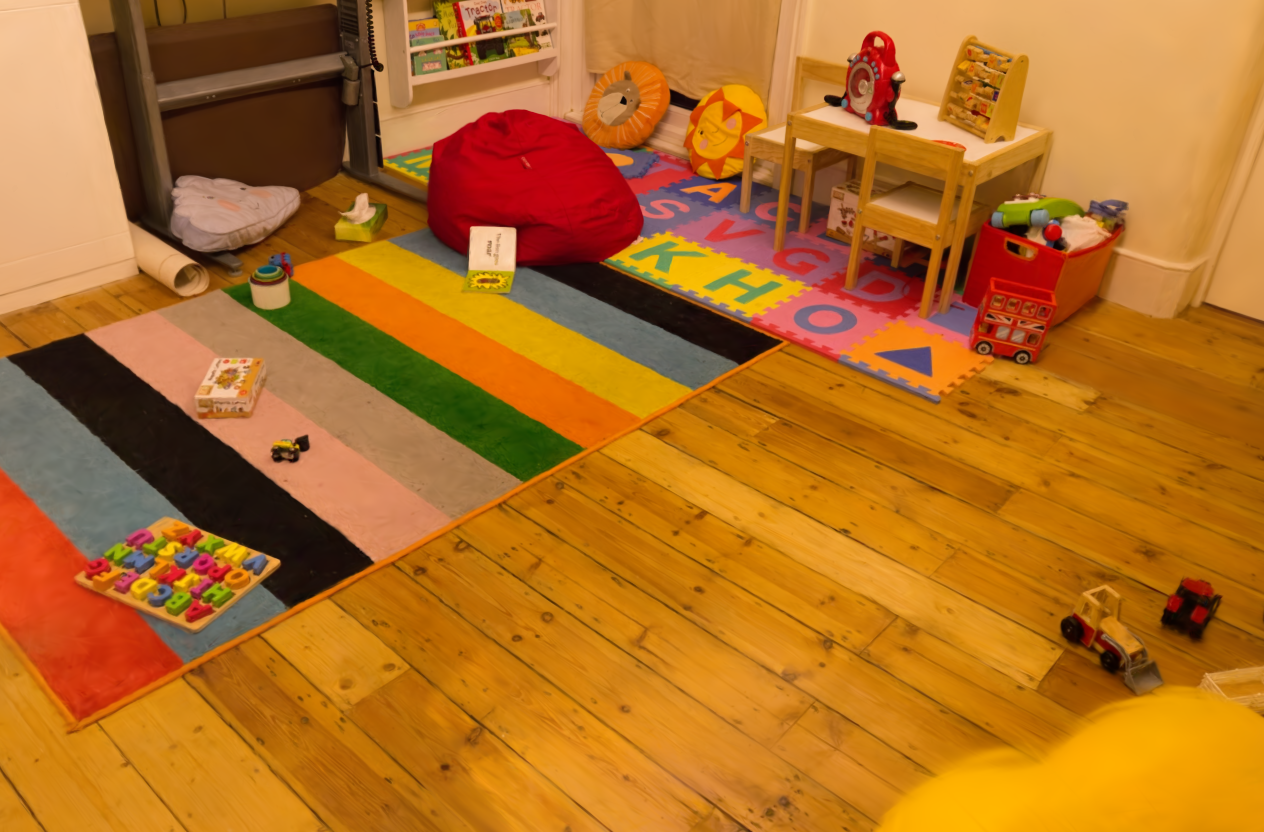}{0.5\mytmplen}{0.57\mytmplen}{0.845\mytmplen}{0.163\mytmplen}{1.2cm}{\mytmplen}{3.0}{red}
\zoomin{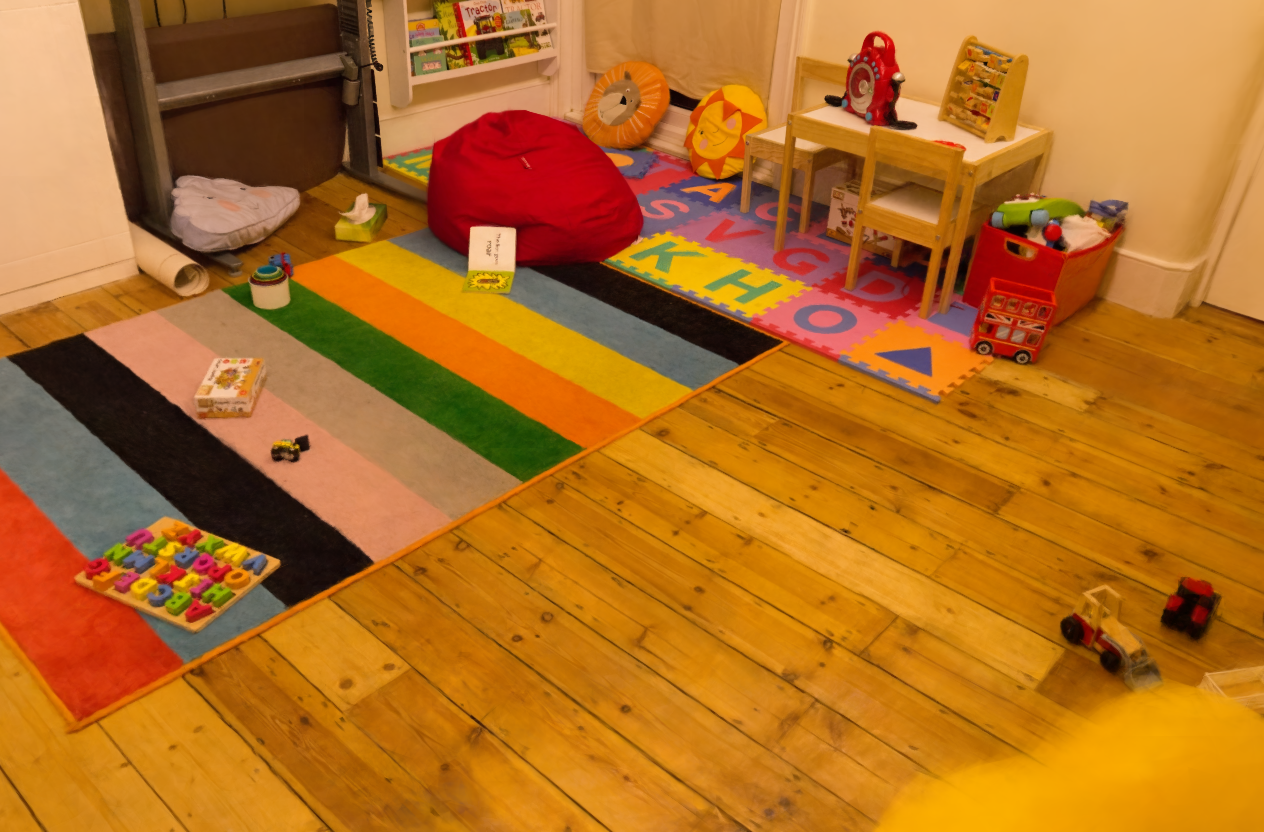}{0.5\mytmplen}{0.57\mytmplen}{0.845\mytmplen}{0.163\mytmplen}{1.2cm}{\mytmplen}{3.0}{red}

\caption{\myTitle{Visual Comparison Between our Light Model and 3DGS.}
The light model (right) shows high visual quality compared to 3DGS (left), using less than 15\% of the memory.
}%
\label{fig:light_vs_best}
\end{figure}

\methodname light outperforms 3DGS and GES on the T\&T and DP dataset, while using less memory.
\Cref{fig:light_vs_best} contains a visual comparison between 3DGS and light \methodname.
\begin{table}[t]
\centering
\resizebox{0.98\columnwidth}{!}{
\begin{tabular}{@{}l|ccc|ccc}
& \multicolumn{3}{c@{}|}{Outdoor Scene} & \multicolumn{3}{c@{}}{Indoor scene} \\ 
& LPIPS~$\downarrow$ & PSNR~$\uparrow$ & SSIM~$\uparrow$ & LPIPS~$\downarrow$ & PSNR~$\uparrow$ & SSIM~$\uparrow$ \\
\midrule
MipNeRF360 & 0.283 & 24.47 & 0.691 & \sbest 0.180 & \best 31.72 & 0.917 \\
3DGS & \best 0.234 & \sbest 24.64 & \sbest 0.731 & 0.189 & 30.41 & 0.920 \\
GES & 0.243 & 24.46 & 0.724 & 0.189 & 30.85 & 0.922 \\
2DGS & 0.246 & 24.34 & 0.717 & 0.195 & 30.40 & 0.916 \\
\hline
\hline
\methodname (ours) & \sbest 0.238 & 24.07 & 0.700 & \best 0.166 & \sbest 31.33 & \best 0.927

\end{tabular}
}
\caption{\textbf{Quantitative Results on Mip-NeRF 360~\cite{Barron2022MipNeRF360} Dataset.}
We evaluate our method on both indoor and outdoor scenes, demonstrating substantial performance improvements over all 3DGS-based methods in indoor scenes and surpassing MipNeRF360 in SSIM and LPIPS metrics.
}%
\label{tab:mipnerf360}
\end{table}

\mysection{Indoor versus outdoor scenes.}
\Cref{tab:mipnerf360} presents a comparative analysis of indoor versus outdoor scenes from the Mip-NeRF360 dataset.
Indoor scenes consist of structured, flat surfaces with hard edges, while outdoor scenes generally have more unstructured surfaces.
This structural difference advantages convex shapes, which are better suited for capturing the geometric characteristics of indoor environments.
In fact, for indoor scenes, it can be seen that \methodname significantly outperforms 3DGS with an improvement of $0.9$ PSNR, $0.007$ SSIM, and $0.023$ LPIPS, surpassing all other Gaussian-based methods.
Moreover, \methodname achieves superior results in terms of SSIM and LPIPS metrics compared to Mip-NeRF360.
Even in outdoor scenes containing a lot of human-made structures—such as the \textit{Truck} and \textit{Train} scenes from T\&T, \methodname substantially outperforms 3DGS, demonstrating its ability to effectively handle structured geometries.
However, in outdoor scenes dominated by nature and unstructured elements like trees and vegetation, the strengths of \methodname become less pronounced.
While 3DGS and \methodname achieve comparable LPIPS results, 3DGS achieves better PSNR and SSIM.
Yet, qualitatively, we can see in Figure \cref{fig:qualityresults} that \methodname appears significantly closer to the ground truth in terms of visual quality.
Specifically, in the highlighted region of the \textit{Flower} scene, our reconstruction better represents the real grass even though the PSNR of this area is $20.17$ for \methodname and $21.65$ for 3DGS. This showcases the popularly observed mismatch between PSNR and perceived visual quality.
This is mainly due to the fact that PSNR is highly sensitive to pixel-level differences and, therefore, tends to favor blurrier images.

\subsection{Ablation Study and Discussion}

We analyze key design choices affecting the performance and efficiency of our convex splatting framework.
We evaluate the impact of densification strategies, the number of points per convex shape, and the influence of reducing the number of shapes on rendering quality.

\begin{figure}[ht]
\setlength\mytmplen{0.3\linewidth}
  \centering
  \zoomin{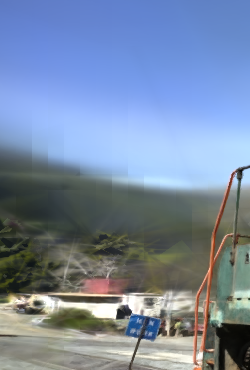}{0.22\mytmplen}{0.53\mytmplen}{0.25\mytmplen}{1.23\mytmplen}{1.2cm}{\mytmplen}{2.5}{red} 
  \zoomin{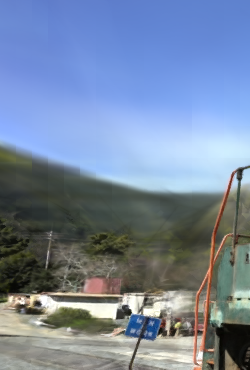}{0.22\mytmplen}{0.53\mytmplen}{0.25\mytmplen}{1.23\mytmplen}{1.2cm}{\mytmplen}{2.5}{red} 
  \zoomin{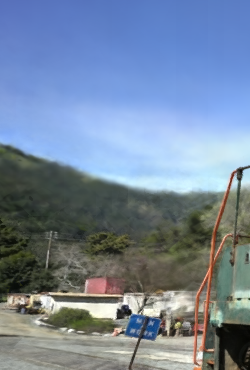}{0.22\mytmplen}{0.53\mytmplen}{0.25\mytmplen}{1.23\mytmplen}{1.2cm}{\mytmplen}{2.5}{red} 
\caption{\myTitle{Ablation of densification strategy.} From left to right, we split each convex into 2, 3, or 6 new convexes.}%
\label{fig:split}
\end{figure}

\begin{table}[h]
    \centering
          \tabcolsep=0.1cm
    \begin{tabular}{lcccc}
    \toprule
     \multirowcell{1}{}  & \multirowcell{1}{$LPIPS^\downarrow$} & \multirowcell{1}{$PSNR^\uparrow$} & \multirowcell{1}{$SSIM^\uparrow$} & \multirowcell{1}{Train$^\downarrow$} \\
    \midrule
    \methodname (K=3) & 0.241 & 22.40 & 0.794 & 44m  \\
    \methodname (K=4)  & 0.159 & 23.73 & 0.848 & 52m \\
    \methodname (K=5)  & 0.160 & 23.70 & 0.848 & 60m \\
    \methodname (K=6)  & \bf 0.157 & \bf 23.90 & 0.850 & 71m \\
    \methodname (K=7)  & \bf 0.157 & \bf 23.90 & \bf 0.851 & 73m \\
    \bottomrule
    
    \end{tabular}
    \footnotesize
    \caption{\textbf{Ablation Study of the Number of Points per Convex.}  We study the impact of the number of points per convex on reconstruction quality and training time on the T\&T dataset. With only 4 points, our \methodname performs better than 3DGS.}
    \label{tab:num_points_per_convex}
\end{table}

\begin{figure}[t]
 \centering
 \includegraphics[width=1\linewidth, trim=0mm 0mm 0mm 0mm, clip]{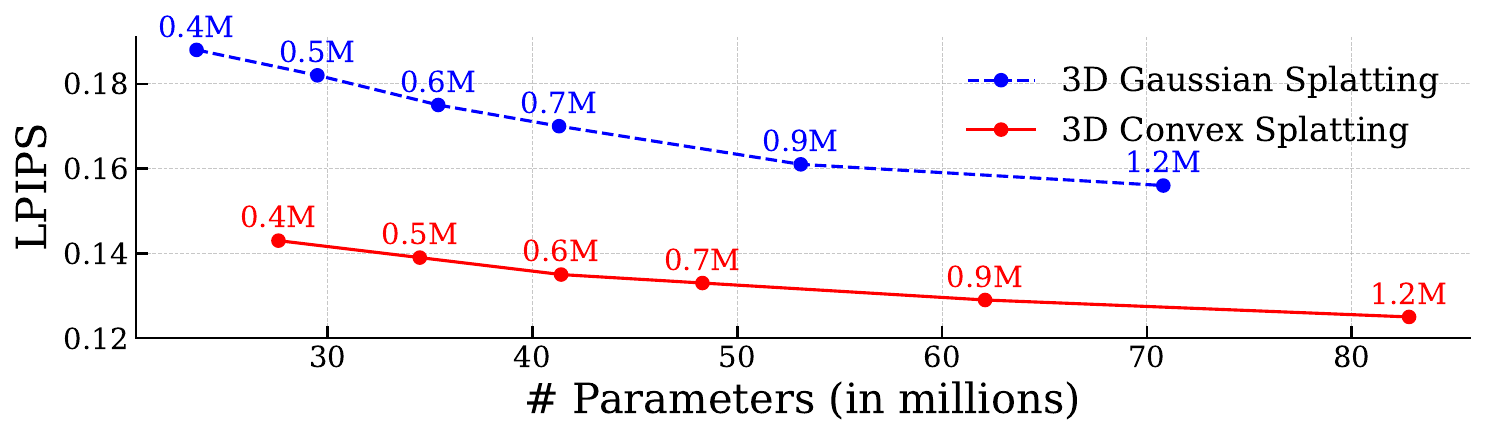}
 \caption{\myTitle{\# Parameters \vs $LPIPS^\downarrow$ (\textit{Truck} scene).}
 The number of primitives is indicated for each point.
 \methodname achieves a better regime than 3DGS for a comparable number of parameters.
 }%
 \label{fig:shapes_vs_PSNR}
 \end{figure}

\mysection{Densification strategy.}
We evaluate the effectiveness of splitting each convex shape into new convex shapes, as described in \cref{adaptive_density_control}.
Specifically, we analyze the impact of dividing a convex shape defined initially by $\pointsPerConvex=6$ points into $2$, $3$, or $6$ (default value) new convex shapes.
For splitting into $2$ or $3$ shapes, the new convex shapes are centered on $2$ or $3$ randomly selected points from the original convex shape.
As can be seen in \cref{fig:split}, splitting a convex shape into more shapes results in higher visual quality, particularly in capturing finer details in the background.

\mysection{Number of points per shape.}
Increasing $\pointsPerConvex$ provides greater flexibility in representing convex shapes but comes at the cost of longer training times. Notably, the case of 3 points represents a special configuration, resulting in a non-volumetric triangle in 3D space, analogous to the 2D Gaussian Splatting approach~\cite{Huang20242DGaussian} in terms of its dimensionality constraints. 
\Cref{tab:num_points_per_convex} shows that using $\pointsPerConvex\geq 4$ points per convex consistently outperforms 3DGS. However, increasing beyond 6 points has no significant performance gain.

\begin{figure}[t]
\setlength\mytmplen{0.48\linewidth}
\centering
\zoomin{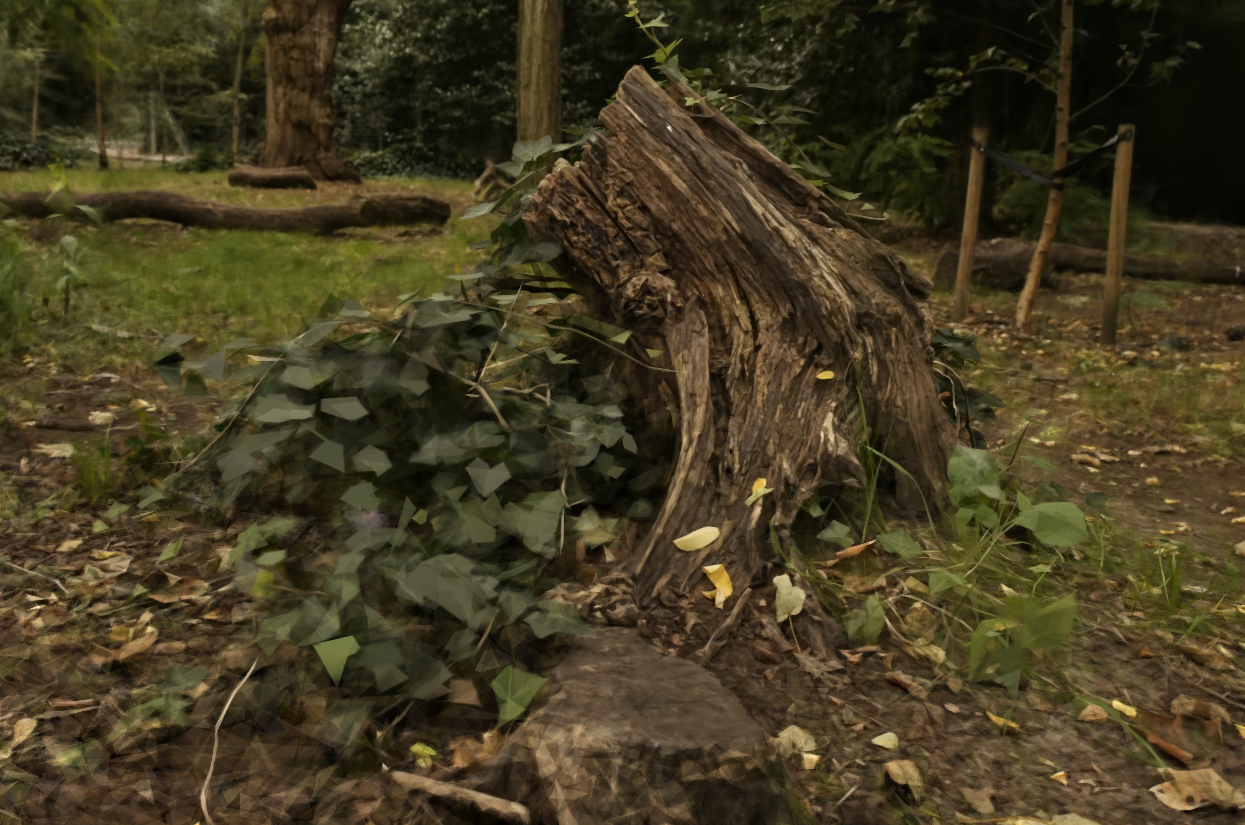}{0.28\mytmplen}{0.28\mytmplen}{0.83\mytmplen}{0.16\mytmplen}{1.2cm}{\mytmplen}{3}{red}
\zoomin{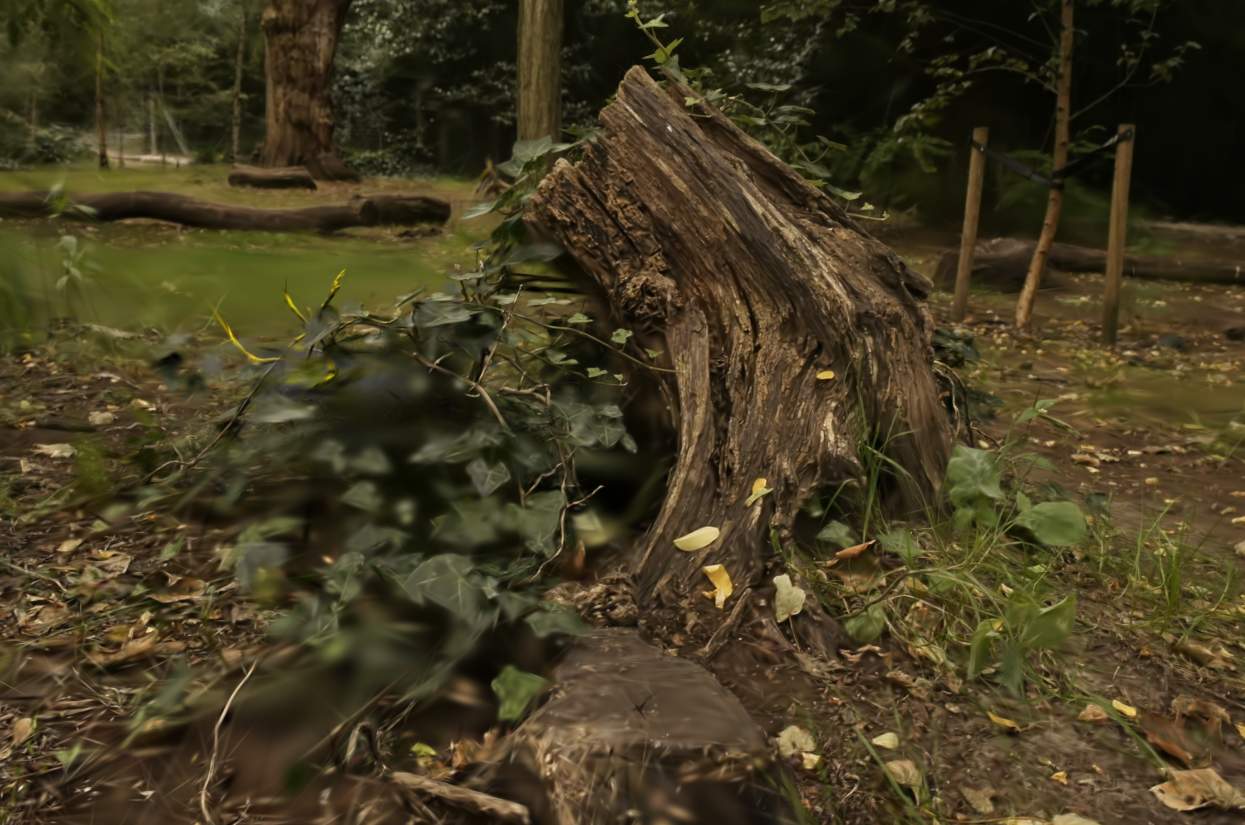}{0.28\mytmplen}{0.28\mytmplen}{0.84\mytmplen}{0.16\mytmplen}{1.2cm}{\mytmplen}{3}{red}
\caption{ \myTitle{\methodname \vs 3DGS with fewer shapes. }
Convex splatting (left) can decompose objects into meaningful convex shapes, enabling a realistic and compact 3D representation of the world.
}%
\label{fig:gaus_vs_conv}
\end{figure}

\mysection{Influence of less primitives on rendering quality.}
\Cref{fig:shapes_vs_PSNR} shows how LPIPS on the T\&T dataset changes with the number of primitives and parameters.
Notably, our method \methodname consistently outperforms 3DGS.

\mysection{Physically meaningful 3D representations.}
\Cref{fig:gaus_vs_conv} visually compares the performance of 3DGS and \methodname when the number of primitives $N$ is reduced.
With fewer shapes, 3DGS produces blurry images due to the limited flexibility of 3D Gaussians, which struggle to form visually meaningful representations of real objects.
In contrast, \methodname preserves image clarity by effectively decomposing objects into convex shapes.
\methodname represents the leaves on the stump as either a single convex shape or a collection of convex shapes, with each shape capturing a physically meaningful part of the real-world object.
Ultimately, \methodname offers a significant advantage by delivering more physically meaningful 3D representations.
By leveraging the adaptability of convex shapes, we bridge the gap between visual accuracy and interpretability, enabling high-quality, geometrically meaningful 3D modeling.

\section{Conclusion}%
\label{sec:conclusion}

We introduce 3D Convex Splatting (\methodname), a novel method for radiance field rendering that leverages 3D smooth convex primitives to achieve high-quality novel view synthesis. 
Particularly, our method overcomes the limitations of 3D Gaussian Splatting, delivering denser representations with fewer primitives and parameters. 
Furthermore, \methodname demonstrates substantial improvements on the novel view synthesis task, particularly on the Tanks\&Temples dataset and indoor scenes from the Mip-NeRF360 dataset. 
By combining the adaptability of convex shapes with the efficiency of primitive-based radiance field rendering, \methodname achieves high-quality, real-time, and flexible radiance field reconstruction.
We envision this new primitive to set the ground for further research in the field.

\mysection{Acknowledgments}
J. Held, A. Deliege and A. Cioppa are funded by the F.R.S.-FNRS. The research reported in this publication was supported by funding from KAUST Center of Excellence on GenAI, under award number 5940. This work was also supported by KAUST Ibn Rushd Postdoc Fellowship program. The present research benefited from computational resources made available on Lucia, the Tier-1 supercomputer of the Walloon Region, infrastructure funded by the Walloon Region under the grant agreement n°1910247.

\maketitlesupplementary

\section{Initialization \& Hyperparameters}

We initialize each convex shape with a set of points uniformly distributed around a sphere centered at points from the point cloud, using the Fibonacci sphere algorithm. The initial sphere radius is set to 1.2 times the mean distance to the three nearest neighbors in the point cloud. This adaptive initialization ensures that dense 3D regions contain many small convex shapes, while sparser regions are represented by larger convexes.
The initial values for the smoothness parameter $\delta$ and sharpness parameter $\sigma$  are set to 0.1 and 0.00095, respectively. These values are chosen to produce initially more diffuse shapes, as this configuration was empirically found to result in better performance during optimization.
The initial opacity is set to 0.1.
For our light model, we apply the same set of hyperparameters across all scenes for consistency.
The learning rates are configured as follows: the learning rates for $\sigma$ and $\delta$ are set to 0.0045 and 0.005, respectively. The learning rate for the convex point positions starts at 5e-4 and then is gradually reduced to a final value of 5e-6. 
The learning rate for the mask is set to 0.01.
During cloning, each convex shape is split into six new convex shapes whenever the loss of $\sigma$ exceeds 0.000004. 
The centers of the new convex shapes are positioned at the six points defining the original convex shape. Each new convex shape is scaled down, made more transparent, and have a higher $\sigma$ value. This adjustment encourages the optimization process to generate denser representations of the shapes.
The densification process starts after 500 iterations and we densify and prune every 200 iterations thereafter.
Convex shapes with an opacity lower than 0.03 are removed, as well as those whose size exceeds 0.3 times the scene size. This scaling ensures that larger scenes can have proportionally larger shapes.
We stop densification after 9,000 iterations, but we continue removing shapes until the end of training.
The final weights of our light model are stored in 16-bit precision, effectively reducing memory requirements while preserving high-quality rendering. 
For our best model, we fine-tune the hyperparameters specifically for indoor and outdoor scenes. In contrast to our light model, we lower the densification threshold to increase the number of convex shapes for a more detailed representation.
For indoor scenes, the split convex shapes are scaled down by a factor of 0.7, while for outdoor scenes, they are scaled down by a factor of 0.6. Additionally, in outdoor scenes, we further reduce $\sigma$ of the split convex shapes, leading to denser representations. This is particularly useful as outdoor environments may require diffuse shapes, for instance, to represent elements like the sky or clouds. 
In indoor scenes, where most objects are human-made, denser shapes are required for an accurate decomposition of the scene.
Finally, all weights of our full model are saved in 32-bit precision.

\section{Methodology Details}

\mysection{2D equations.}
We define the 2D convex indicator function for our convex hull by adapting the smooth convex representation from 3D to 2D, utilizing the equations introduced in \cref{sec:preliminary}.
Specifically, we define $\phi(\mathbf{q})$ and $I(\mathbf{q})$ as in \cref{eq:smoothness,eq:sharpness}, but substitute the 3D point $\mathbf{p}$ with the 2D point $\mathbf{q}$ and replace the planes delimiting the 3D convex hull with the lines that delimit the resulting 2D convex hull.

\begin{equation} \label{eq:smoothness_supp}
\phi(\mathbf{q}) = \log \left( \sum_{t=1}^{T} \exp\left( d \multiplicationSymbol \delta \multiplicationSymbol L_j(\mathbf{q}) \right) \right)\comma
\end{equation}
where $T$ is the total number of lines delimiting the 2D convex shape.

The indicator function $I(\mathbf{p})$ of the smooth convex is then defined by:
\begin{equation} \label{eq:sharpness_supp}
I(\mathbf{q}) = \operatorname{Sigmoid}\left( -d \multiplicationSymbol \sigma \multiplicationSymbol \phi(\mathbf{q}) \right)\comma
\end{equation}

\section{Ablation Study}

\mysection{Perspective-Aware Scaling in 2D Projection.}
To incorporate perspective effects in the 2D projection, we scale $\delta$ and $\sigma$ by the distance d, ensuring that the appearance of the convex shape remains consistent regardless of its distance from the camera. 
Table \ref{tab:distance_d_s} provides an ablation study demonstrating the necessity of scaling $\delta$ and $\sigma$ as well as analyzing the impact of the scaling magnitude.

\begin{table}[h]
    \centering
          \tabcolsep=0.1cm
    \begin{tabular}{l|cc|cc}
     \multirowcell{1}{Magnitude}  & \multirowcell{1}{Truck} & \multirowcell{1}{Train} & \multirowcell{1}{DrJohnson} & \multirowcell{1}{Playroom} \\
    \midrule
    $1$ & 19.47  & 19.14  & 29.17  &  28.82 \\
    $\sqrt{d}$  & 25.49  & 21.41 & 29.49  & 29.98 \\
    $d$  & \best 25.65  & \best 22.23 & \best 29.54 & \best 30.08 \\
    $d^2$  & 7.08 & 8.91 & 8.42 & 8.99 \\
    
    \end{tabular}
    \footnotesize
    \caption{\textbf{Perspective-Aware Scaling in 2D Projection.} We evaluate the PSNR under varying scaling magnitudes. }
    \label{tab:distance_d_s}
\end{table}

\begin{figure*}[t]
\centering
\setlength\mytmplen{0.23\linewidth}
\newlength{\labelheight}
\setlength{\labelheight}{1cm} 
\resizebox{\linewidth}{!}{ 
\begin{tabular}{c@{\hskip 0.2in}c@{\hskip 0.2in}c@{\hskip 0.2in}c}

    \rotatebox{90}{\parbox[c][\labelheight][c]{2.2cm}{\centering Rectangle}}
    \includegraphics[width=0.23\linewidth]{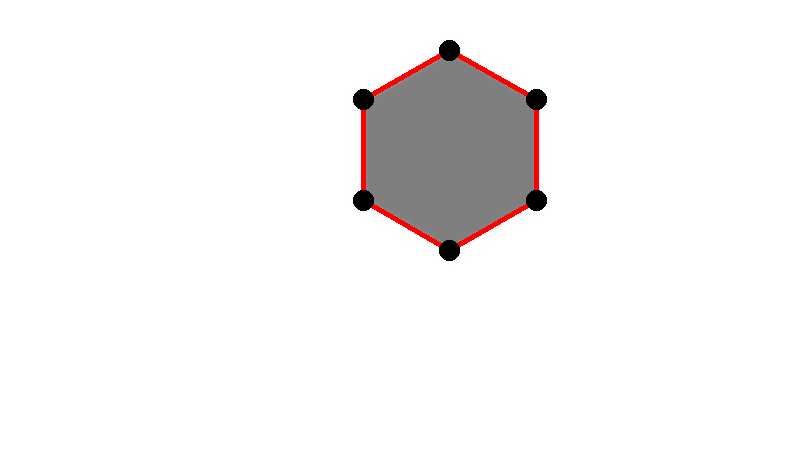} &
    \includegraphics[width=0.23\linewidth]{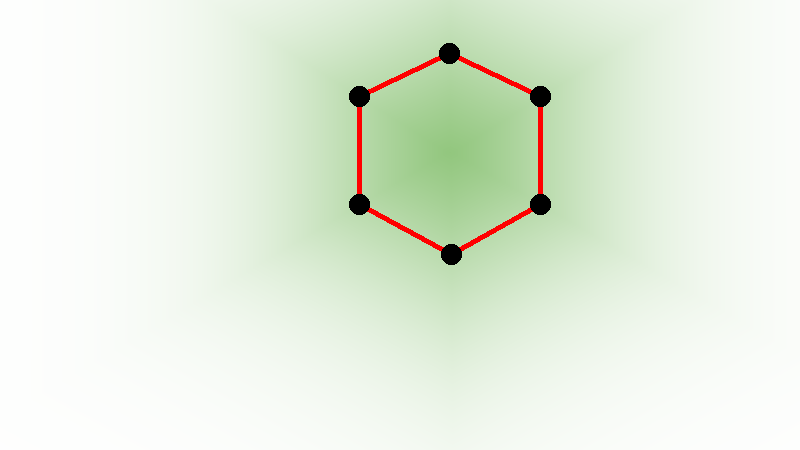} &
    \includegraphics[width=0.23\linewidth]{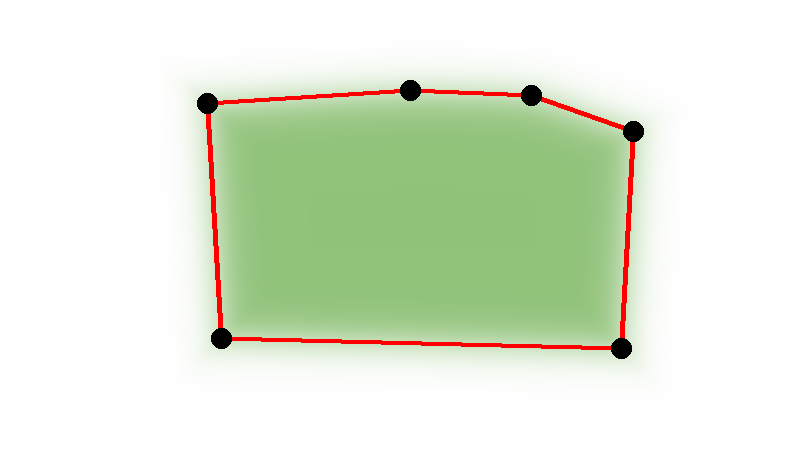} &
    \includegraphics[width=0.23\linewidth]{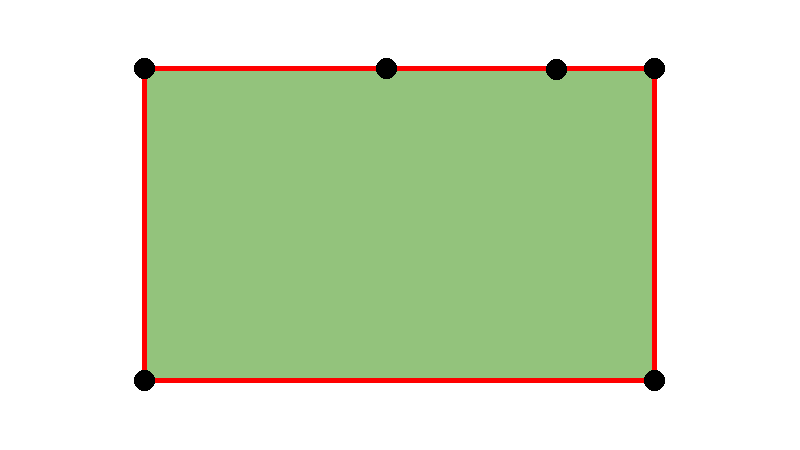} \\

    \rotatebox{90}{\parbox[c][\labelheight][c]{2.2cm}{\centering Circle}}
    \includegraphics[width=0.23\linewidth]{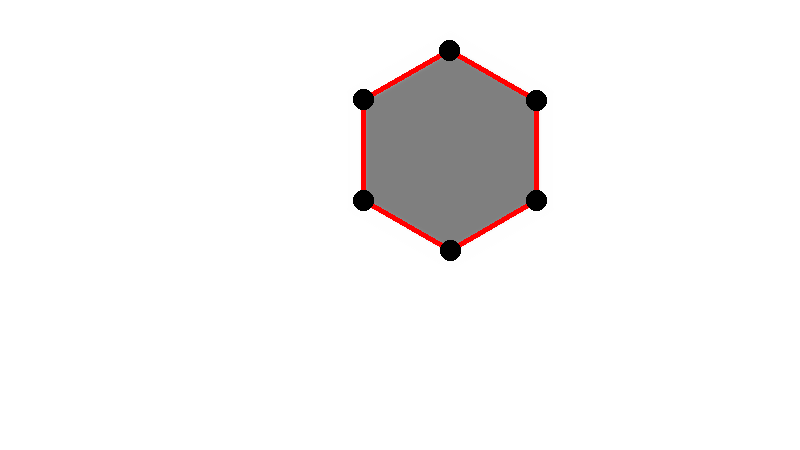} &
    \includegraphics[width=0.23\linewidth]{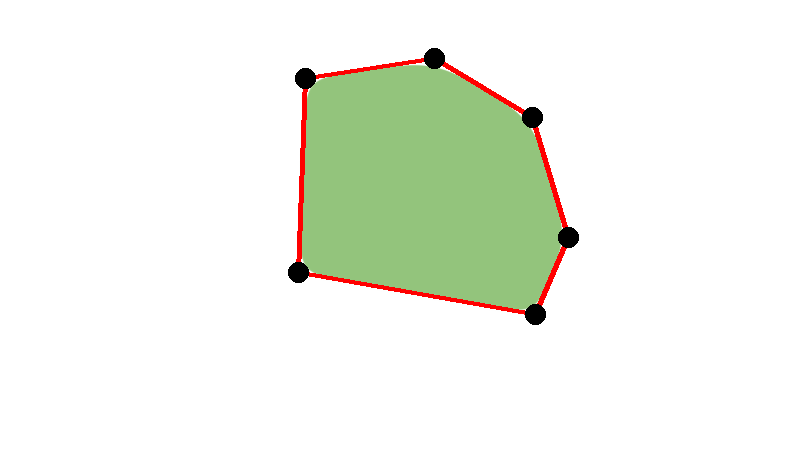} &
    \includegraphics[width=0.23\linewidth]{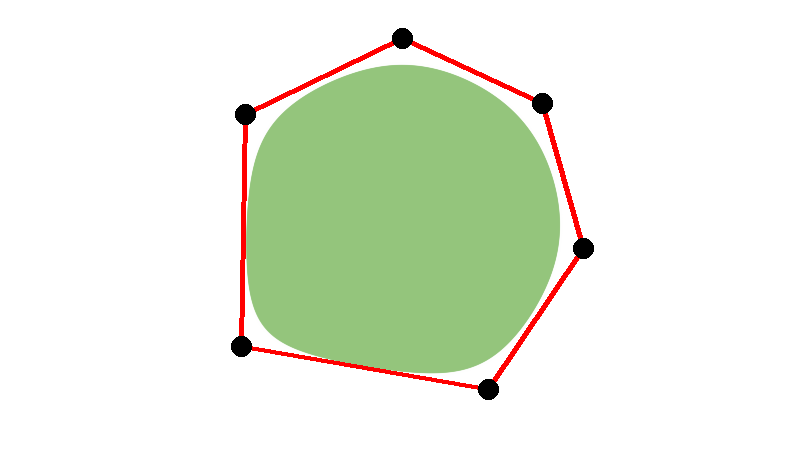} &
    \includegraphics[width=0.23\linewidth]{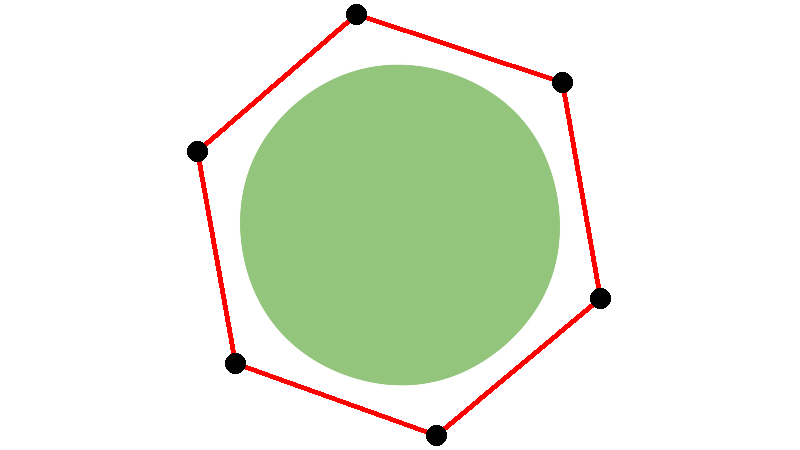} \\

    \rotatebox{90}{\parbox[c][\labelheight][c]{2.2cm}{\centering Isotropic Gaussian}}
    \includegraphics[width=0.23\linewidth]{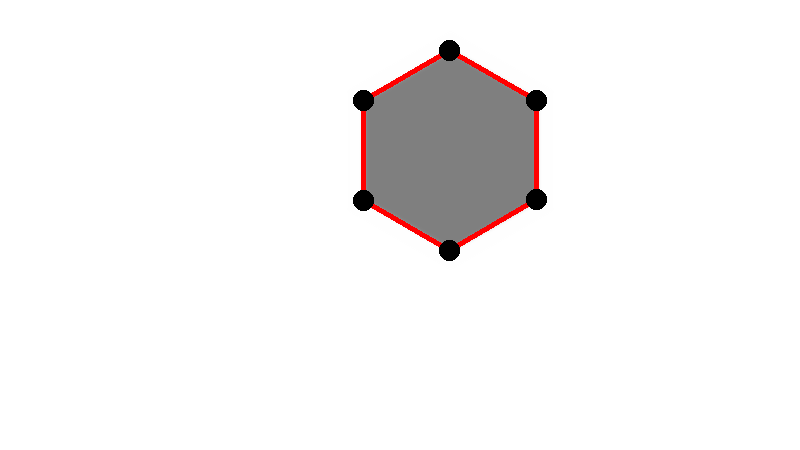} &
    \includegraphics[width=0.23\linewidth]{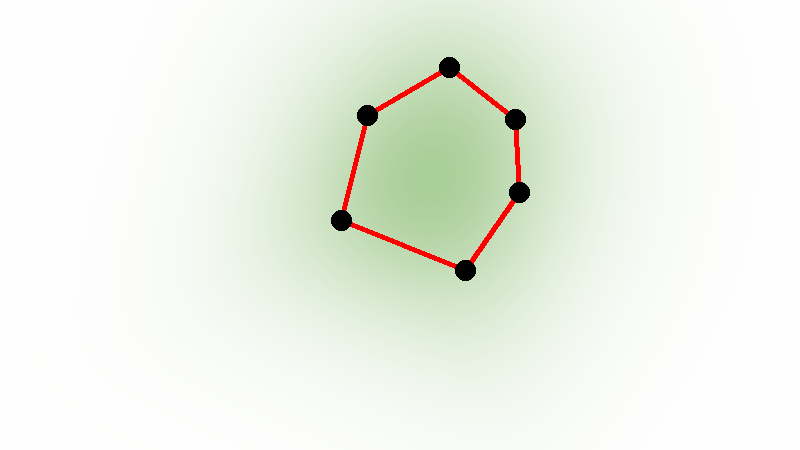} &
    \includegraphics[width=0.23\linewidth]{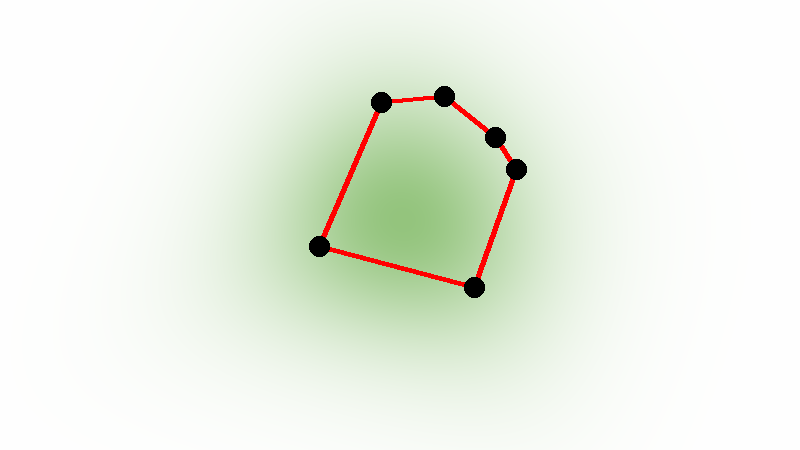} &
    \includegraphics[width=0.23\linewidth]{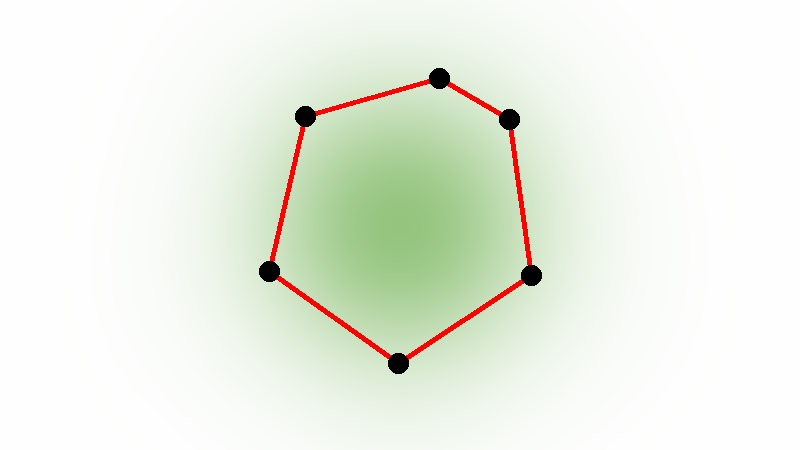} \\

    \rotatebox{90}{\parbox[c][\labelheight][c]{2.2cm}{\centering Anisotropic Gaussian}}
    \includegraphics[width=0.23\linewidth]{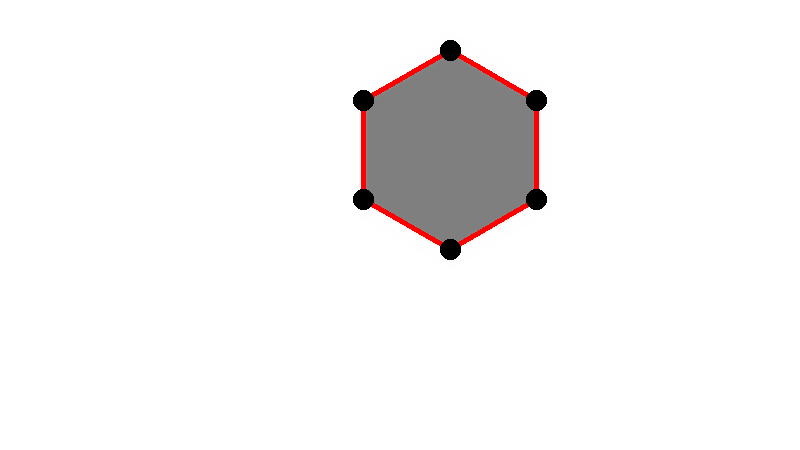} &
    \includegraphics[width=0.23\linewidth]{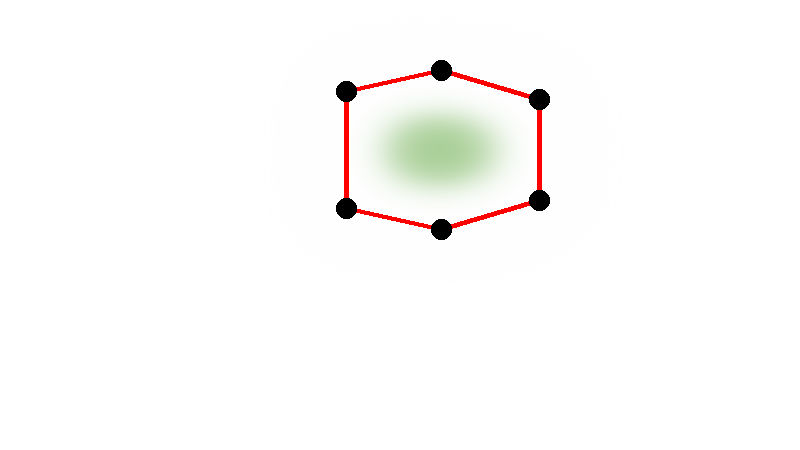} &
    \includegraphics[width=0.23\linewidth]{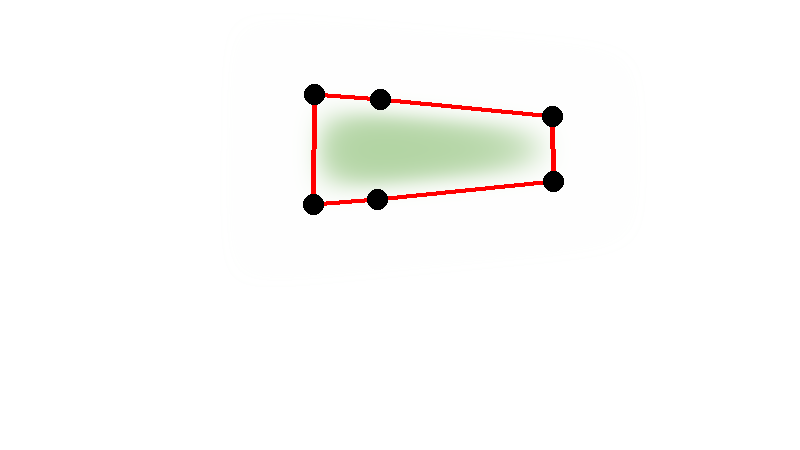} &
    \includegraphics[width=0.23\linewidth]{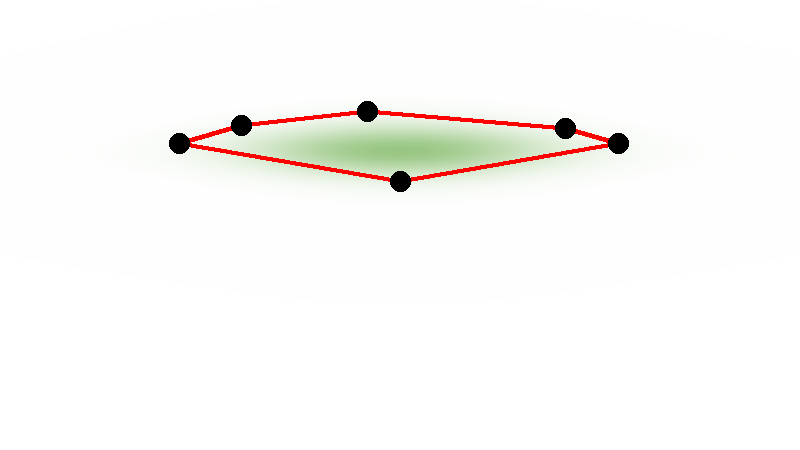} \\

    \multicolumn{4}{c}{\begin{tikzpicture}
        \draw[thick, ->] (0,0) -- (10,0) node[midway, below] {Training time};
    \end{tikzpicture}} \\

\end{tabular}
}
\caption{
Smooth convexes can represent a wide variety of shapes, whether hard or soft, dense or diffuse. They effectively approximate diverse geometries, including both polyhedra and Gaussians, while requiring fewer primitives for accurate representation. The red lines describe the convex hull, whereas the black dots represent the point set.
}
\label{fig:synth_data}
\end{figure*}
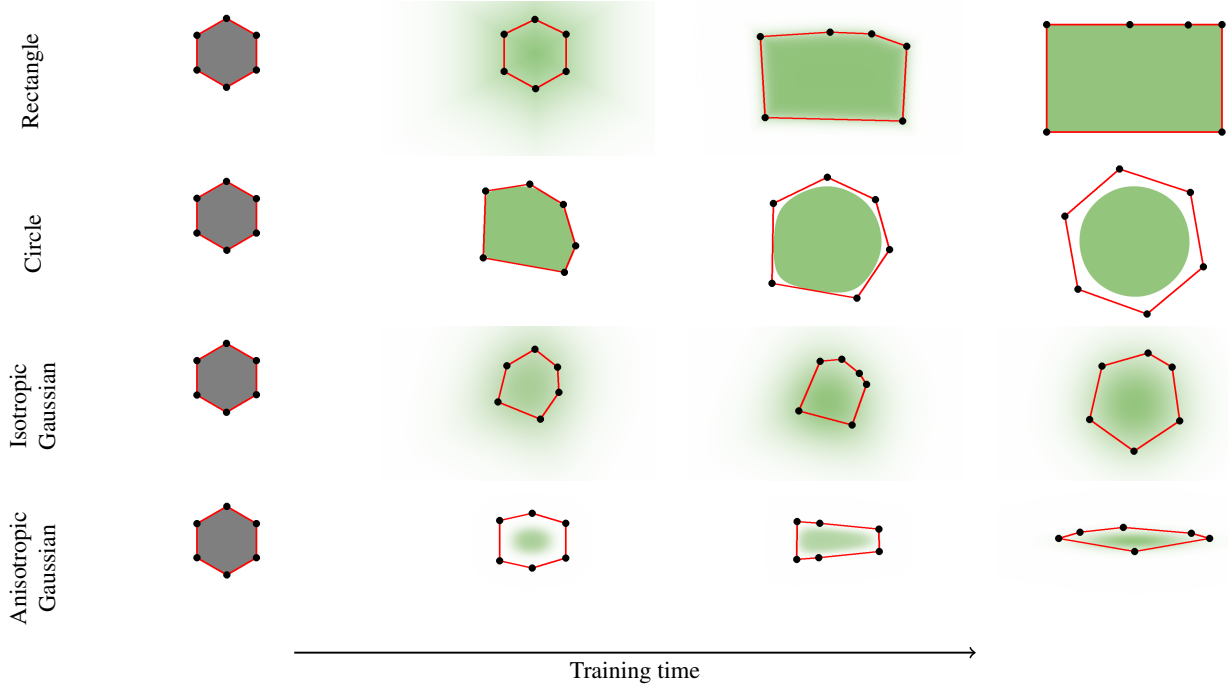

\begin{table}[h]
    \centering
          \tabcolsep=0.1cm
    \begin{tabular}{l|cc|cc}
     \multirowcell{1}{}  & \multirowcell{1}{Truck} & \multirowcell{1}{Train} & \multirowcell{1}{DrJohnson} & \multirowcell{1}{Playroom} \\
    \midrule
    3DGS & 0.148 & 0.218 & 0.244 & 0.241 \\
    2DGS & 0.173 & 0.251 & 0.257 & 0.257 \\
    GES  & 0.162 & 0.232  & 0.249 & 0.252  \\
    \hline
    3DCS  & \best 0.125  & \best 0.187  & \best 0.238  & \best 0.237 \\
    
    \end{tabular}
    \footnotesize
    \caption{LPIPS score for T\&T and DB datasets. }
    \label{tab:1}
\end{table}

\begin{table}[h]
    \centering
          \tabcolsep=0.1cm
    \begin{tabular}{l|cc|cc}
     \multirowcell{1}{}  & \multirowcell{1}{Truck} & \multirowcell{1}{Train} & \multirowcell{1}{DrJohnson} & \multirowcell{1}{Playroom} \\
    \midrule
    3DGS & 25.18 & 21.09 & 28.76 & 30.04  \\
    2DGS  & 25.12 & 21.14 & 28.95 & 30.05 \\
    GES  & 25.07 & 21.75 & 29.24 & 30.06  \\
    \hline
    3DCS  & \best 25.65  & \best 22.23 & \best 29.54 & \best 30.08  \\
    
    \end{tabular}
    \footnotesize
    \caption{PSNR score for T\&T and DB datasets. }
    \label{tab:2}
\end{table}

\begin{table}[h]
    \centering
          \tabcolsep=0.1cm
    \begin{tabular}{l|cc|cc}
     \multirowcell{1}{}  & \multirowcell{1}{Truck} & \multirowcell{1}{Train} & \multirowcell{1}{DrJohnson} & \multirowcell{1}{Playroom} \\
    \midrule
    3DGS & 0.879 & 0.802 & 0.899  & \best 0.906  \\
    2DGS  & 0.874 & 0.789 & 0.900 & \best 0.906 \\
    GES  & 0.872 &  0.800 & 0.899 & 0.902  \\
    \hline
    3DCS  & \best 0.882  & \best 0.820  & \best 0.902  & 0.902 \\
    
    \end{tabular}
    \footnotesize
    \caption{SSIM score for T\&T and DB datasets. }
    \label{tab:3}
\end{table}

\begin{table}[h]
    \centering
          \tabcolsep=0.1cm
    \resizebox{0.98\columnwidth}{!}{      
    \begin{tabular}{l|ccccc|cccc}
     \multirowcell{1}{}  & \multirowcell{1}{Bicycle} & \multirowcell{1}{Flowers} & \multirowcell{1}{Garden} & \multirowcell{1}{Stump} & \multirowcell{1}{Treehill} & \multirowcell{1}{Room} & \multirowcell{1}{Counter} & \multirowcell{1}{Kitchen} & \multirowcell{1}{Bonsai}\\
    \midrule
    3DGS & \best 0.205 & 0.336 & \best 0.103 & \best 0.210 & \best 0.317 & 0.220 & 0.204 & 0.129 & 0.205 \\
    2DGS   & 0.218 & 0.346 & 0.115 & 0.222 & 0.329 & 0.223 & 0.208 & 0.133 & 0.214 \\
    GES  & 0.272 & 0.342  & 0.110 & 0.218 & 0.331  & 0.220 & 0.202 & 0.127 & 0.206 \\
    \hline
    3DCS  &  0.216 & \best 0.322  & 0.113  & 0.227 & \best 0.317  & \best 0.193  &  \best 0.182 & \best 0.117 & \best 0.182 \\
    
    \end{tabular}
    }    
    \footnotesize
    \caption{LPIPS score for the MipNerf360 dataset. }
    \label{tab:4}
\end{table}

\begin{table}[h]
    \centering
          \tabcolsep=0.1cm
    \resizebox{0.98\columnwidth}{!}{      
    \begin{tabular}{l|ccccc|cccc}
     \multirowcell{1}{}  & \multirowcell{1}{Bicycle} & \multirowcell{1}{Flowers} & \multirowcell{1}{Garden} & \multirowcell{1}{Stump} & \multirowcell{1}{Treehill} & \multirowcell{1}{Room} & \multirowcell{1}{Counter} & \multirowcell{1}{Kitchen} & \multirowcell{1}{Bonsai}\\
    \midrule
    3DGS & \best 25.24 & \best 21.52 & \best 27.41 & \best 26.55 & 22.49 & 30.63 & 28.70 & 30.31 & 31.98 \\
    2DGS  & 24.87 & 21.15 & 26.95 & 26.47 & 22.27 & 31.06 & 28.55 & 30.50 & 31.52 \\
    GES  & 24.76 & 21.33  & 26.89 & 26.06 & 22.31  & 31.03 & 28.88  & 31.21 & 31.94 \\
    \hline
    3DCS  & 24.72  & 20.52  & 27.09  & 26.12 & \best 21.77  & \best 31.70  & \best 29.02  & \best 31.96 & \best 32.64 \\
    
    \end{tabular}
    }    
    \footnotesize
    \caption{PSNR score for the MipNerf360 dataset. }
    \label{tab:5}
\end{table}

\begin{table}[h]
    \centering
          \tabcolsep=0.1cm
    \resizebox{0.98\columnwidth}{!}{      
    \begin{tabular}{l|ccccc|cccc}
     \multirowcell{1}{}  & \multirowcell{1}{Bicycle} & \multirowcell{1}{Flowers} & \multirowcell{1}{Garden} & \multirowcell{1}{Stump} & \multirowcell{1}{Treehill} & \multirowcell{1}{Room} & \multirowcell{1}{Counter} & \multirowcell{1}{Kitchen} & \multirowcell{1}{Bonsai}\\
    \midrule
    3DGS &  \best 0.771 & \best 0.605 & \best 0.868 & \best 0.775 & \best 0.638 & 0.914 & 0.905 & 0.922 & 0.938 \\
    2DGS  & 0.752 & 0.588 & 0.852 & 0.765 & 0.627 & 0.912 & 0.900 & 0.919 & 0.933 \\
    GES  & 0.727  & 0.600 & 0.846  & 0.768 & 0.631 & 0.910 & 0.899 & 0.920 & 0.939 \\
    \hline
    3DCS  & 0.737  & 0.575  & 0.850 & 0.746 & 0.595  & \best 0.925  & \best 0.909 & \best \best 0.930 & \best 0.945 \\
    
    \end{tabular}
    }    
    \footnotesize
    \caption{SSIM score for the MipNerf360 dataset. }
    \label{tab:6}
\end{table}

\begin{figure*}[t]
\centering
\setlength\mytmplen{0.23\linewidth}
\resizebox{\linewidth}{!}{ 

\begin{tabular}{c@{\hskip 0.2in}c@{\hskip 0.2in}c@{\hskip 0.2in}c}
    
    \makebox[\mytmplen]{Ground Truth} &
    \makebox[\mytmplen]{\textbf{\methodname (ours)}} &
    \makebox[\mytmplen]{3DGS} &
    \makebox[\mytmplen]{2DGS} \\

    \rotatebox{90}{\parbox{2.2cm}{\centering Bicycle}}
    \zoomin{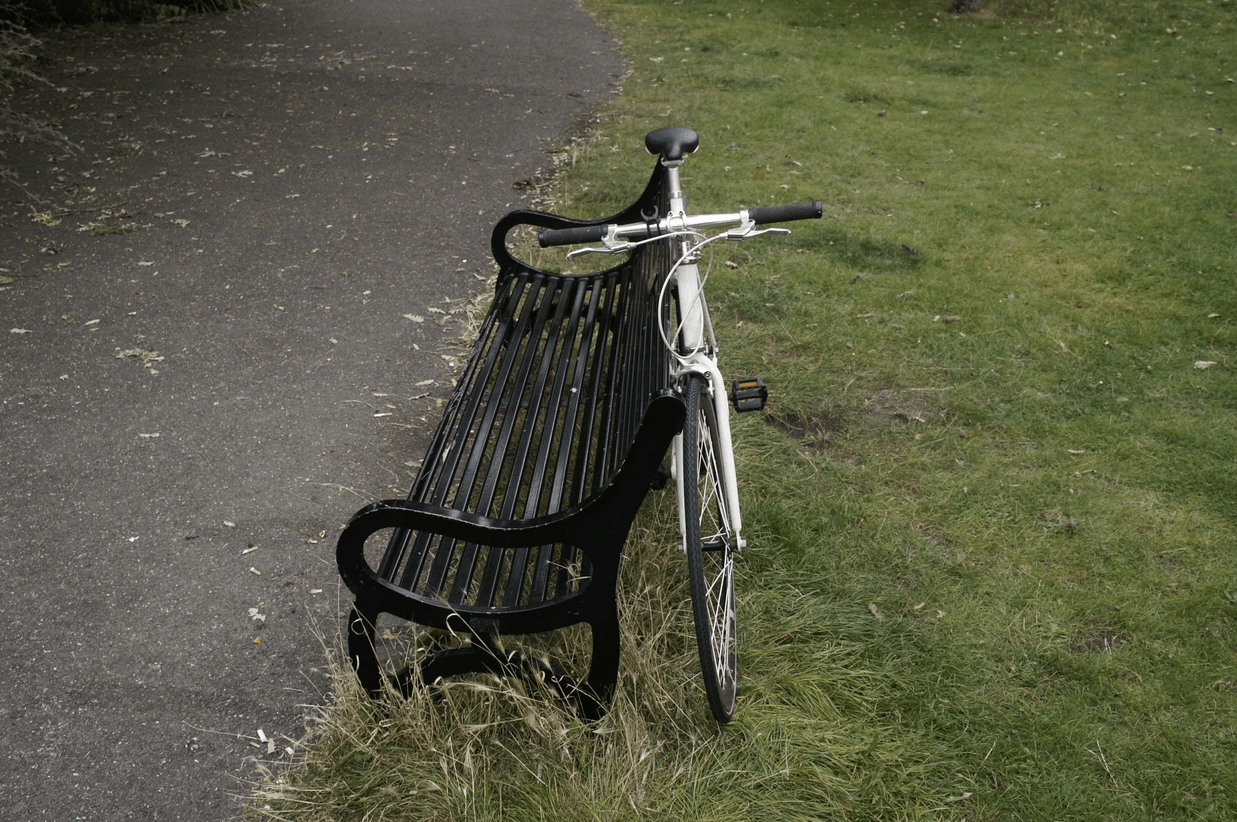}{0.45\mytmplen}{0.6\mytmplen}{0.16\mytmplen}{0.16\mytmplen}{1.2cm}{\mytmplen}{2.5}{red} &
    \zoomin{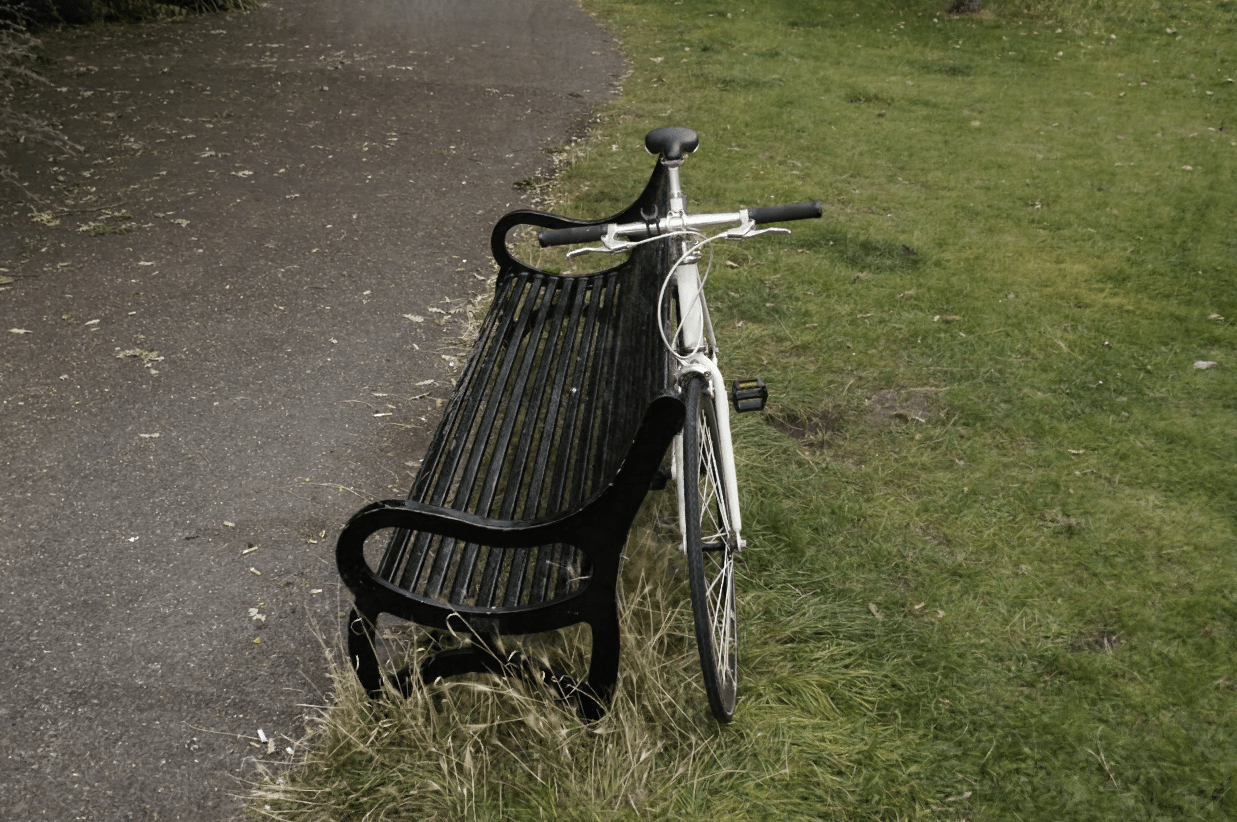}{0.45\mytmplen}{0.6\mytmplen}{0.16\mytmplen}{0.16\mytmplen}{1.2cm}{\mytmplen}{2.5}{red} &
    \zoomin{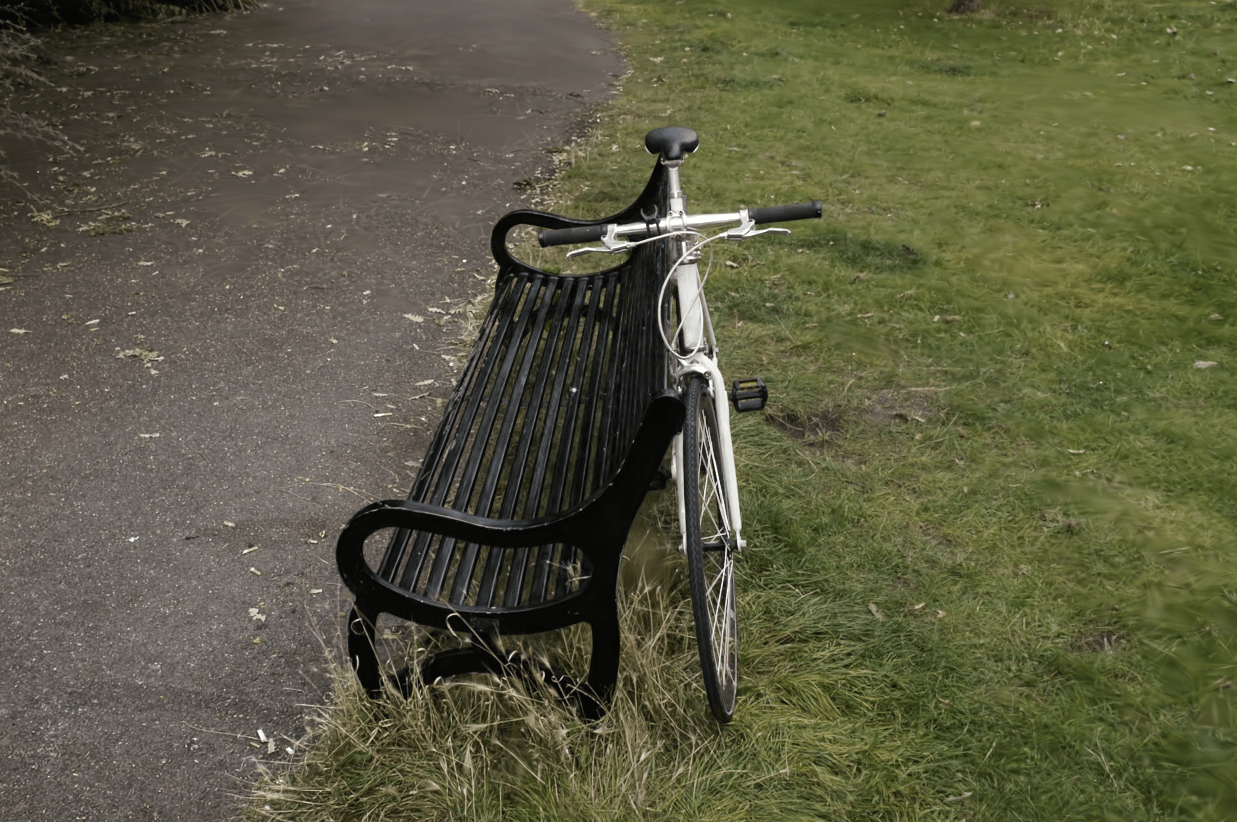}{0.45\mytmplen}{0.6\mytmplen}{0.16\mytmplen}{0.16\mytmplen}{1.2cm}{\mytmplen}{2.5}{red} &
    \zoomin{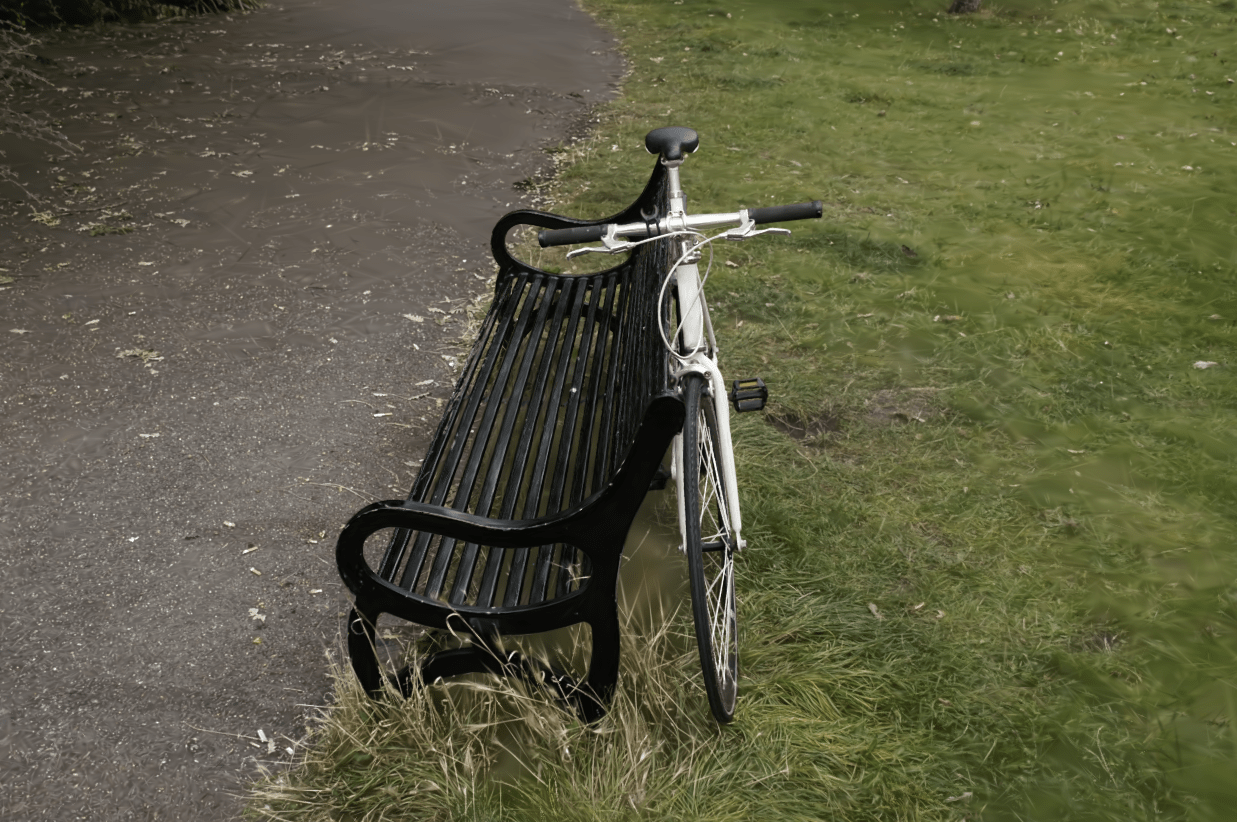}{0.45\mytmplen}{0.6\mytmplen}{0.16\mytmplen}{0.16\mytmplen}{1.2cm}{\mytmplen}{2.5}{red} \\
    
    \rotatebox{90}{\parbox{2.2cm}{\centering Stump}}
    \zoomin{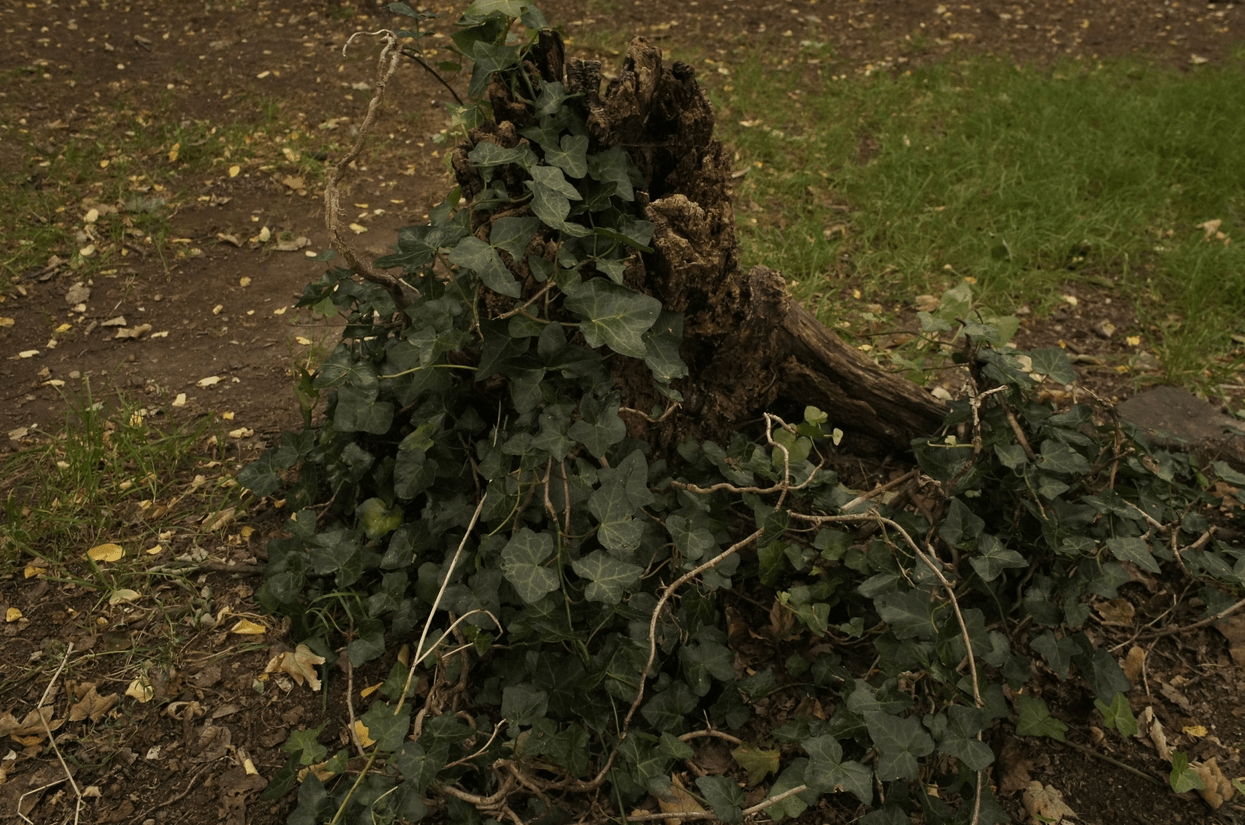}{0.7\mytmplen}{0.5\mytmplen}{0.16\mytmplen}{0.16\mytmplen}{1.2cm}{\mytmplen}{4}{red} &
    \zoomin{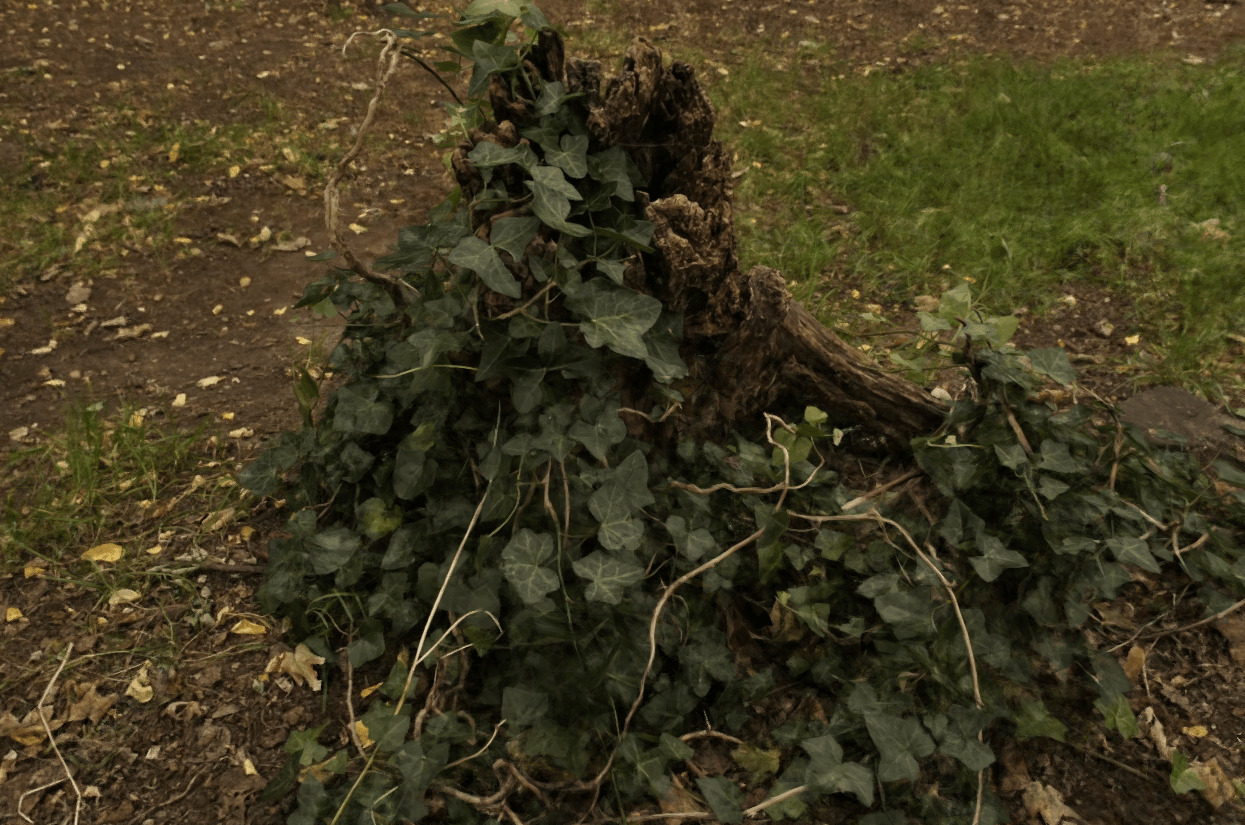}{0.7\mytmplen}{0.5\mytmplen}{0.16\mytmplen}{0.16\mytmplen}{1.2cm}{\mytmplen}{4}{red} &
    \zoomin{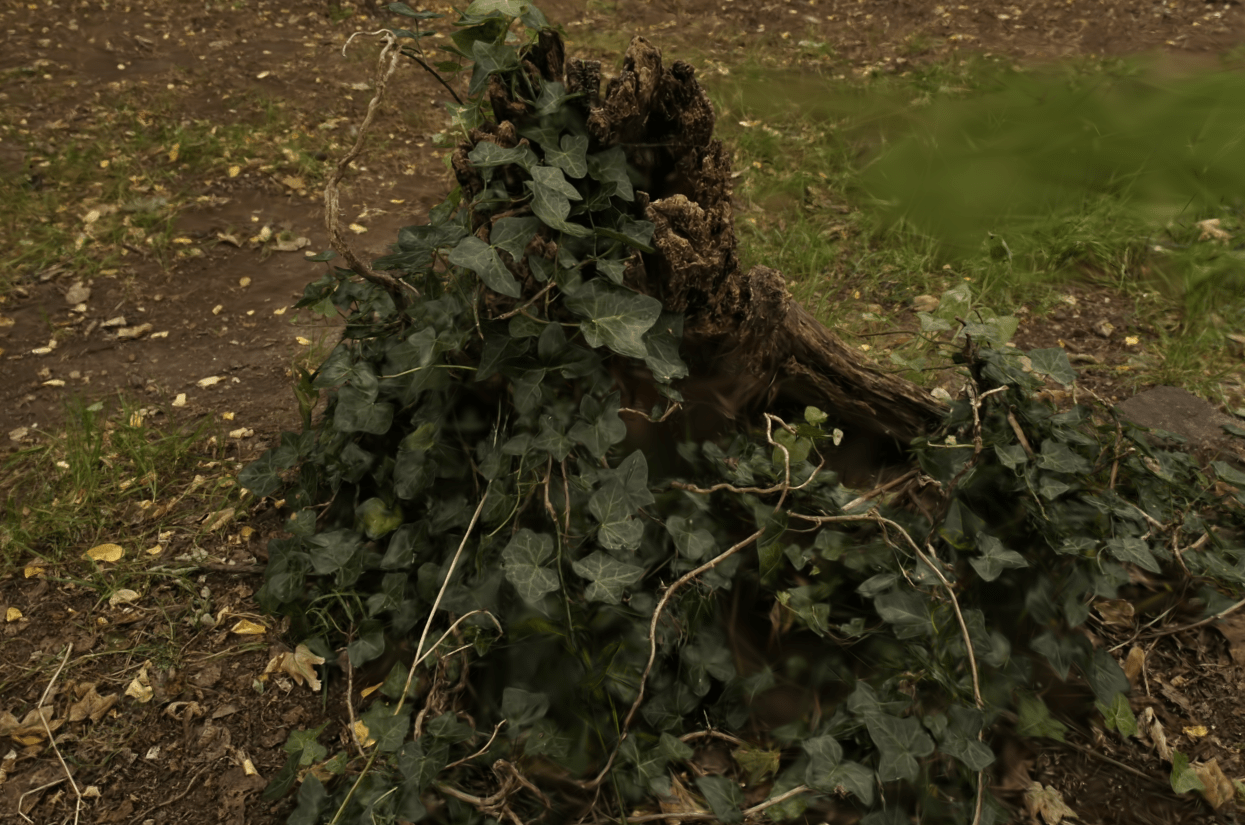}{0.7\mytmplen}{0.5\mytmplen}{0.16\mytmplen}{0.16\mytmplen}{1.2cm}{\mytmplen}{4}{red} &
    \zoomin{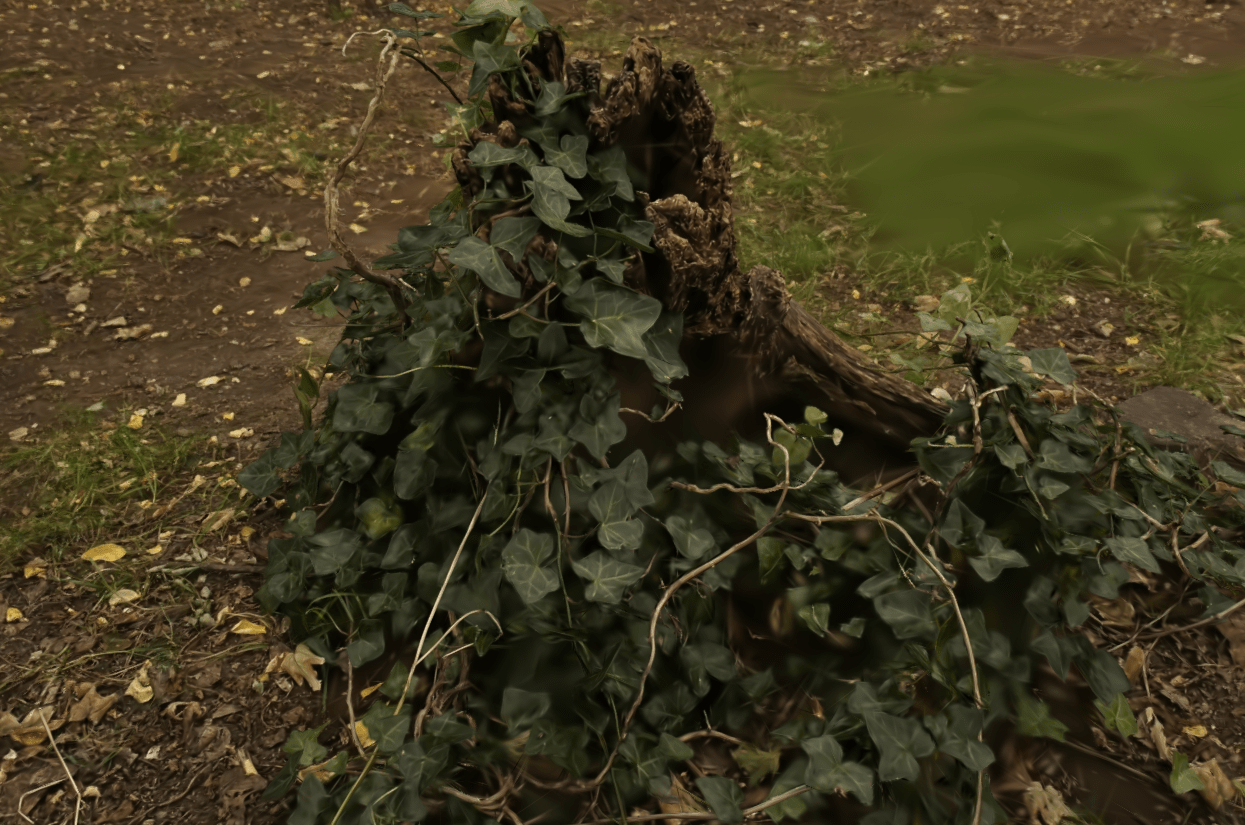}{0.7\mytmplen}{0.5\mytmplen}{0.16\mytmplen}{0.16\mytmplen}{1.2cm}{\mytmplen}{4}{red} \\

    \rotatebox{90}{\parbox{2.2cm}{\centering Flowers}}
    \zoomin{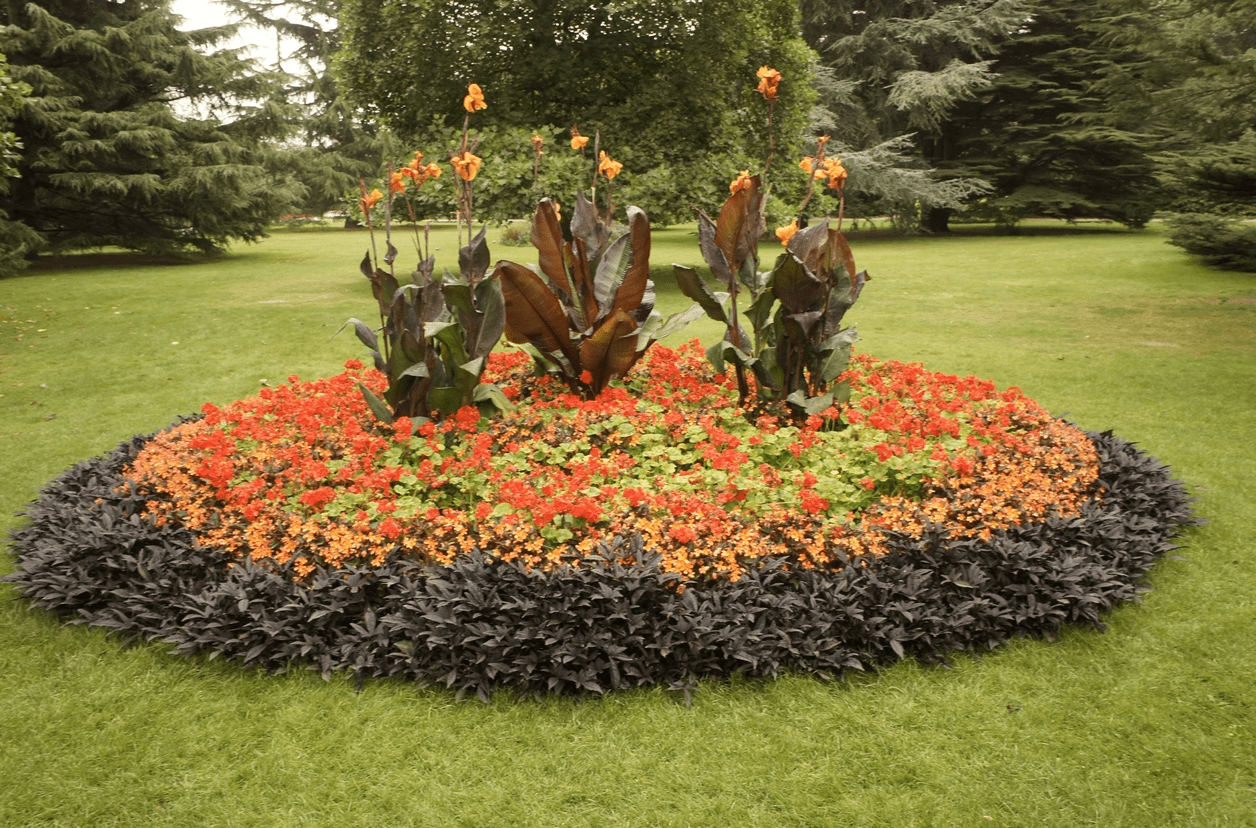}{0.92\mytmplen}{0.35\mytmplen}{0.16\mytmplen}{0.16\mytmplen}{1.2cm}{\mytmplen}{6.5}{red} &
    \zoomin{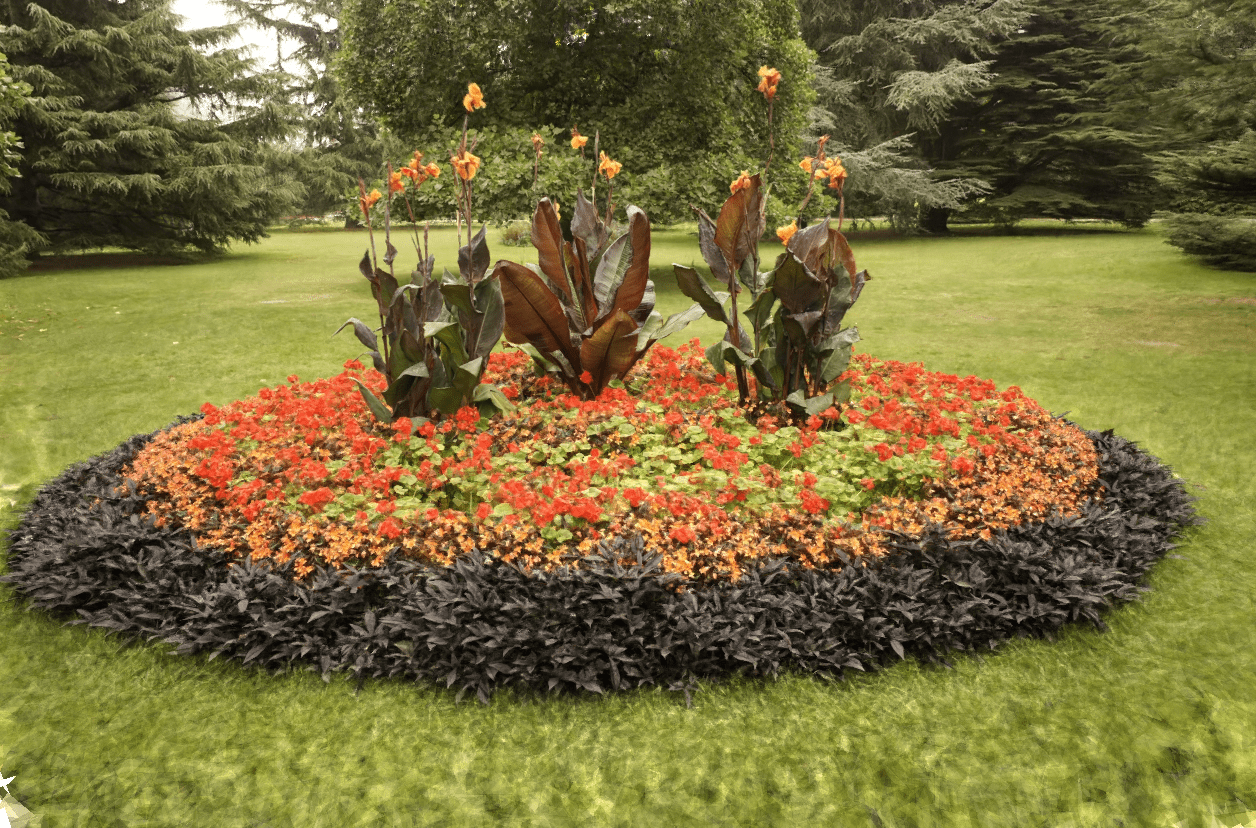}{0.92\mytmplen}{0.35\mytmplen}{0.16\mytmplen}{0.16\mytmplen}{1.2cm}{\mytmplen}{6.5}{red} &
    \zoomin{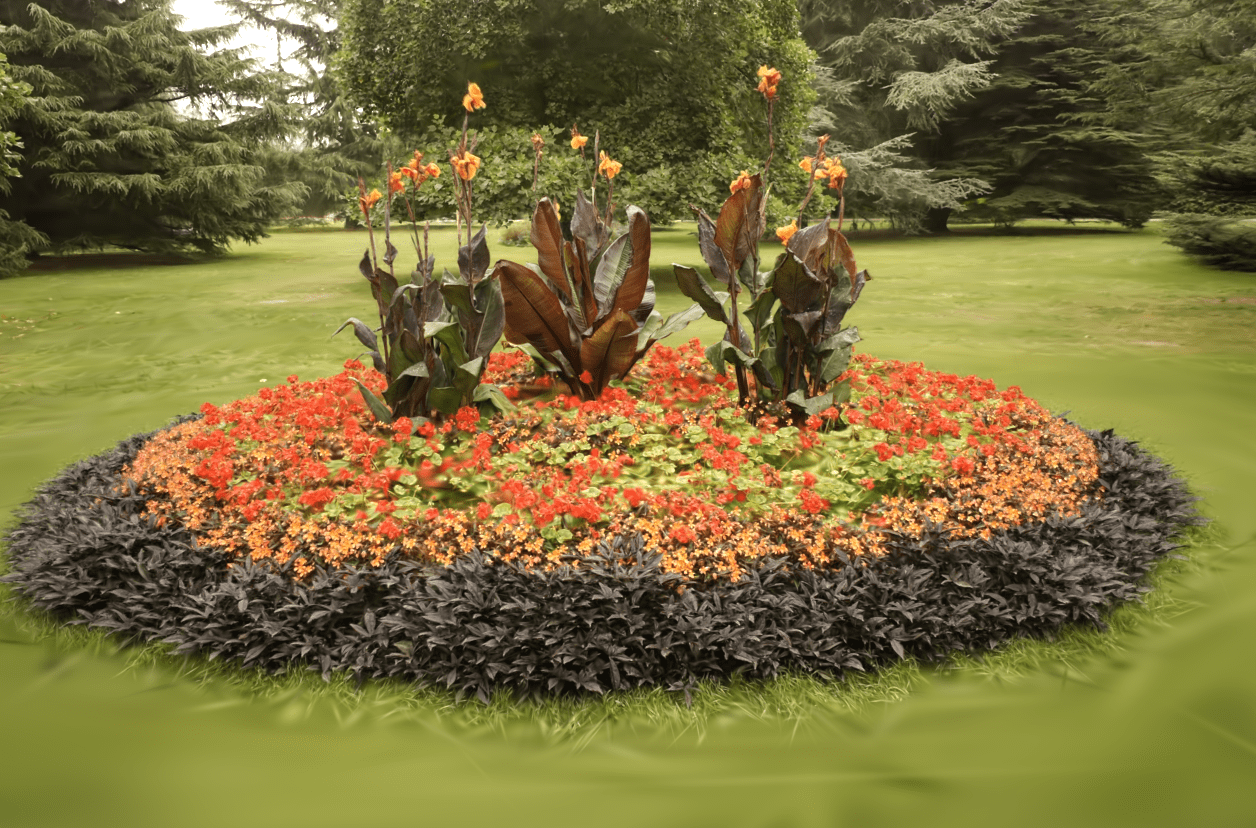}{0.92\mytmplen}{0.35\mytmplen}{0.16\mytmplen}{0.16\mytmplen}{1.2cm}{\mytmplen}{6.5}{red} &
    \zoomin{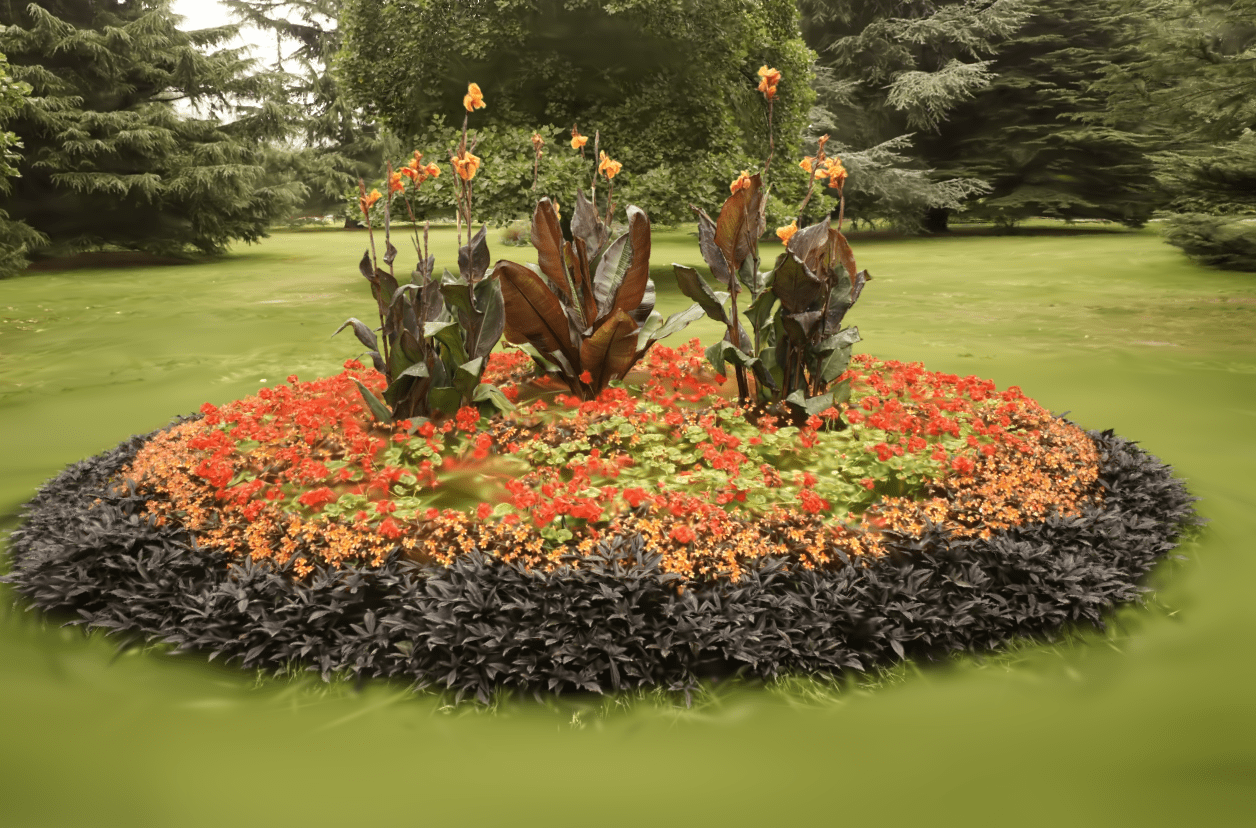}{0.92\mytmplen}{0.35\mytmplen}{0.16\mytmplen}{0.16\mytmplen}{1.2cm}{\mytmplen}{6.5}{red}\\

    \rotatebox{90}{\parbox{2.2cm}{\centering Train}}
    \zoomin{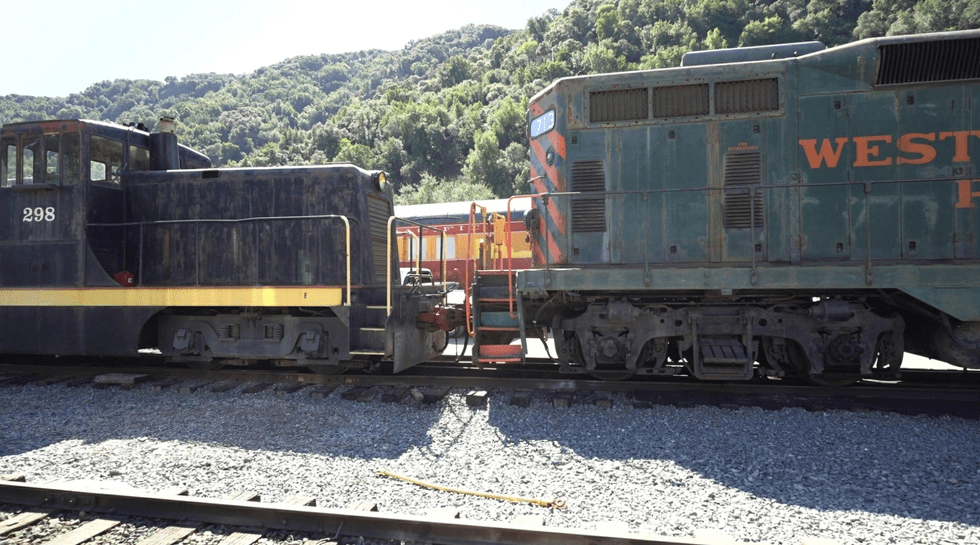}{0.35\mytmplen}{0.45\mytmplen}{0.16\mytmplen}{0.16\mytmplen}{1.2cm}{\mytmplen}{2.5}{red} &
    \zoomin{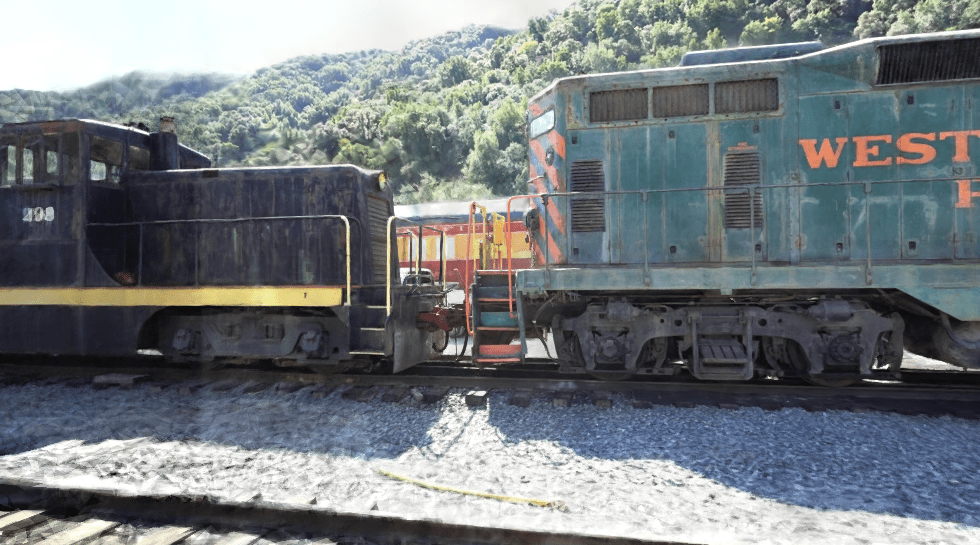}{0.35\mytmplen}{0.45\mytmplen}{0.16\mytmplen}{0.16\mytmplen}{1.2cm}{\mytmplen}{2.5}{red} &
    \zoomin{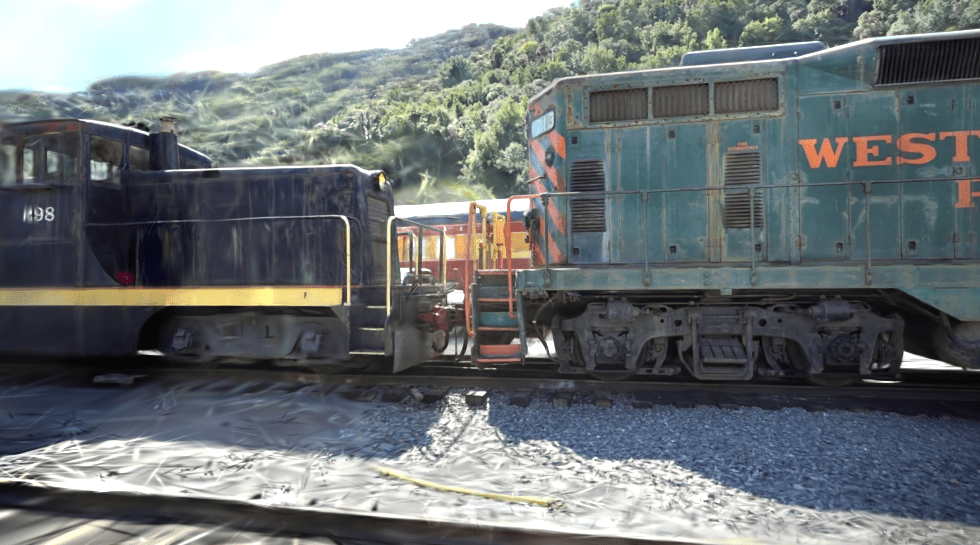}{0.35\mytmplen}{0.45\mytmplen}{0.16\mytmplen}{0.16\mytmplen}{1.2cm}{\mytmplen}{2.5}{red} &
    \zoomin{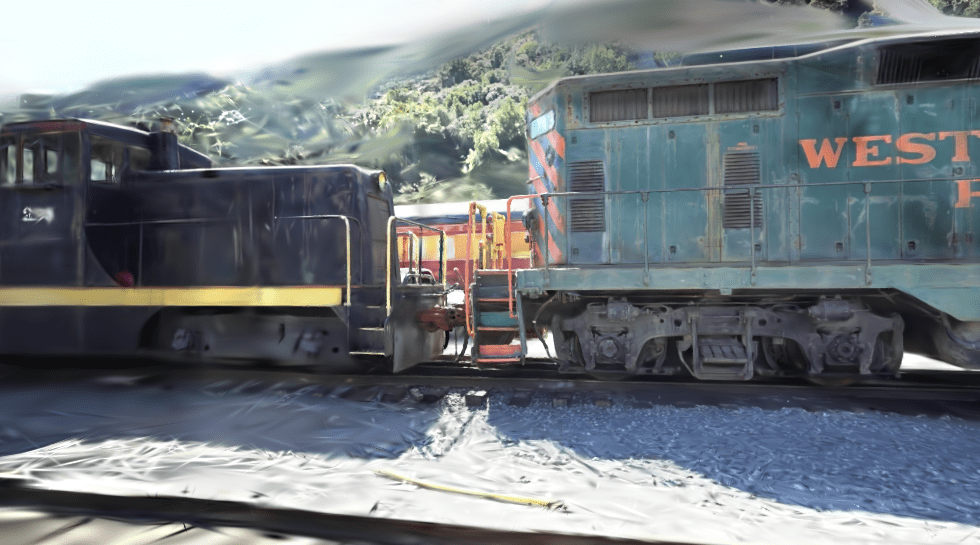}{0.35\mytmplen}{0.45\mytmplen}{0.16\mytmplen}{0.16\mytmplen}{1.2cm}{\mytmplen}{2.5}{red} \\

    \rotatebox{90}{\parbox{2.2cm}{\centering Counter}}
    \zoomin{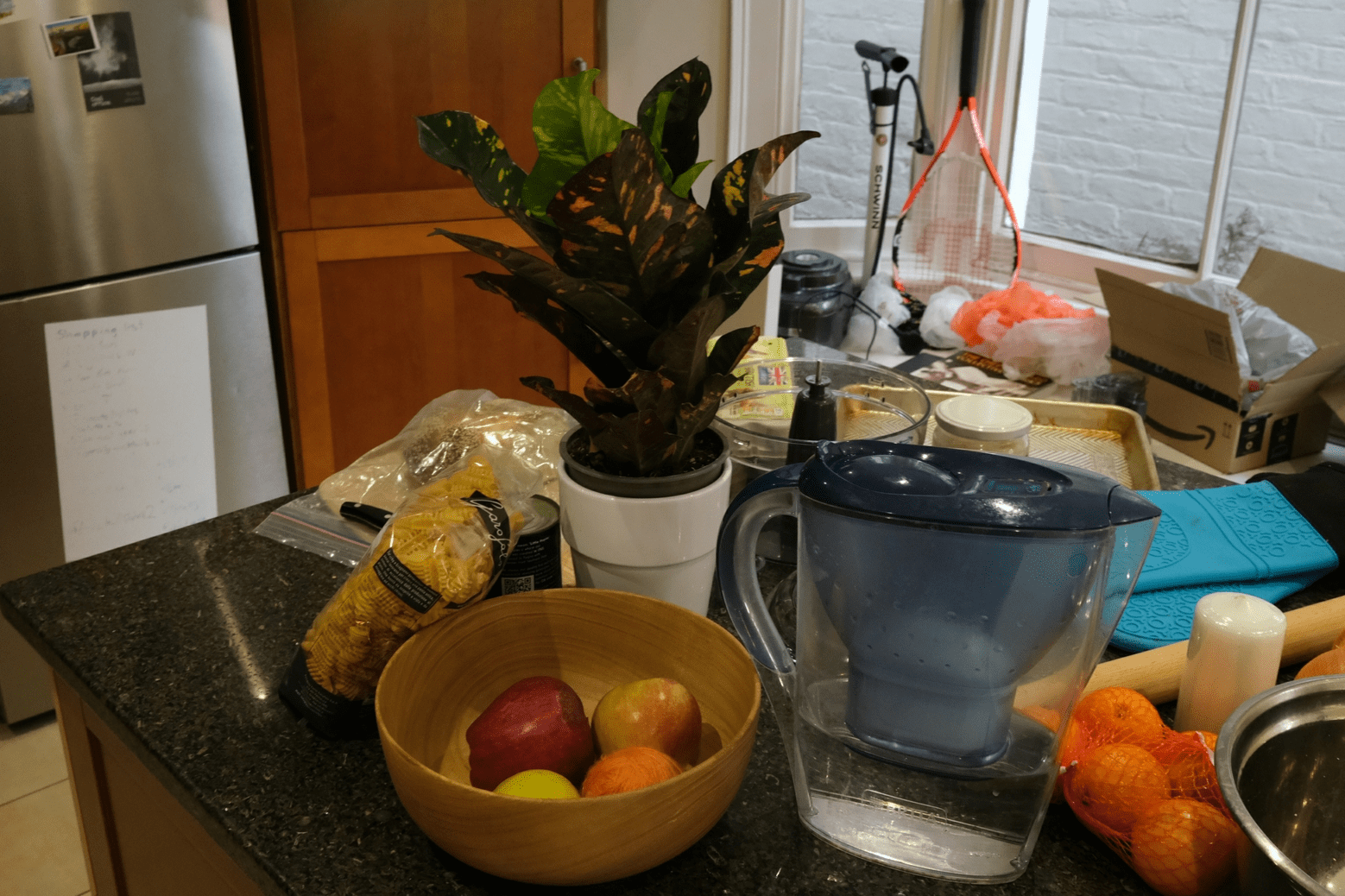}{0.08\mytmplen}{0.35\mytmplen}{0.16\mytmplen}{0.16\mytmplen}{1.2cm}{\mytmplen}{6.5}{red} &
    \zoomin{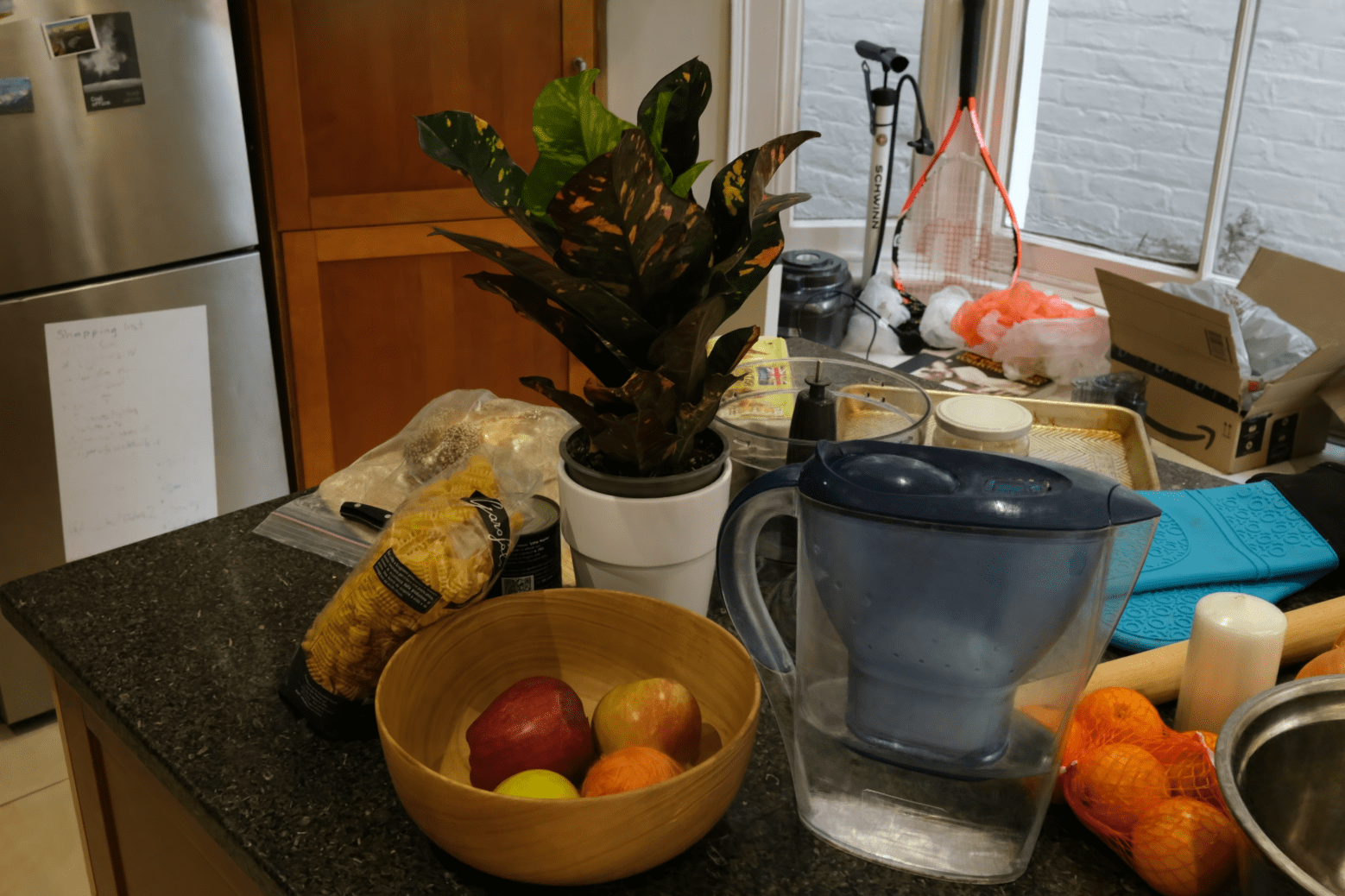}{0.08\mytmplen}{0.35\mytmplen}{0.16\mytmplen}{0.16\mytmplen}{1.2cm}{\mytmplen}{6.5}{red} &
    \zoomin{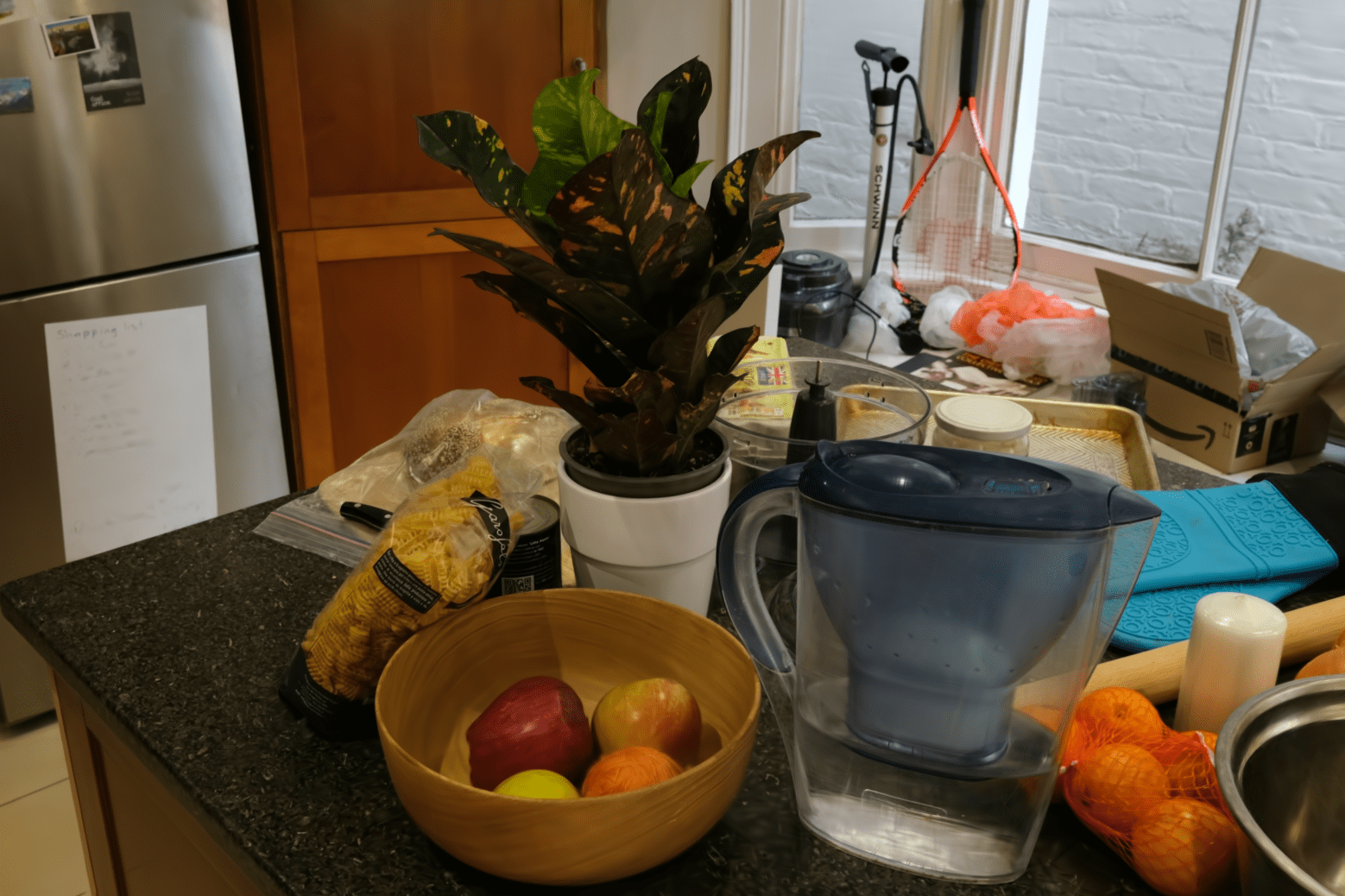}{0.08\mytmplen}{0.35\mytmplen}{0.16\mytmplen}{0.16\mytmplen}{1.2cm}{\mytmplen}{6.5}{red} &
    \zoomin{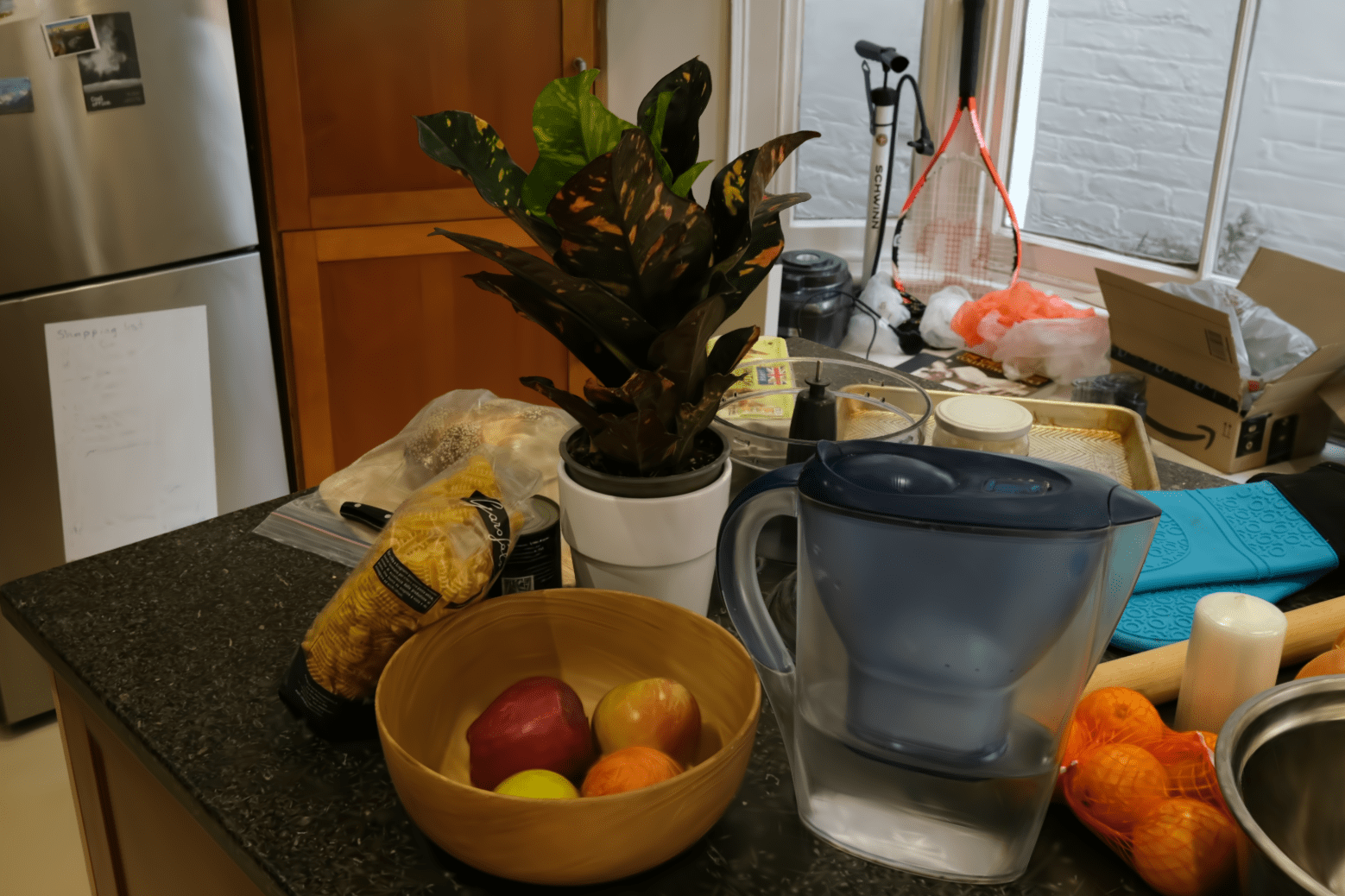}{0.08\mytmplen}{0.35\mytmplen}{0.16\mytmplen}{0.16\mytmplen}{1.2cm}{\mytmplen}{6.5}{red}\\

    \rotatebox{90}{\parbox{2.2cm}{\centering Truck}}
    \zoomin{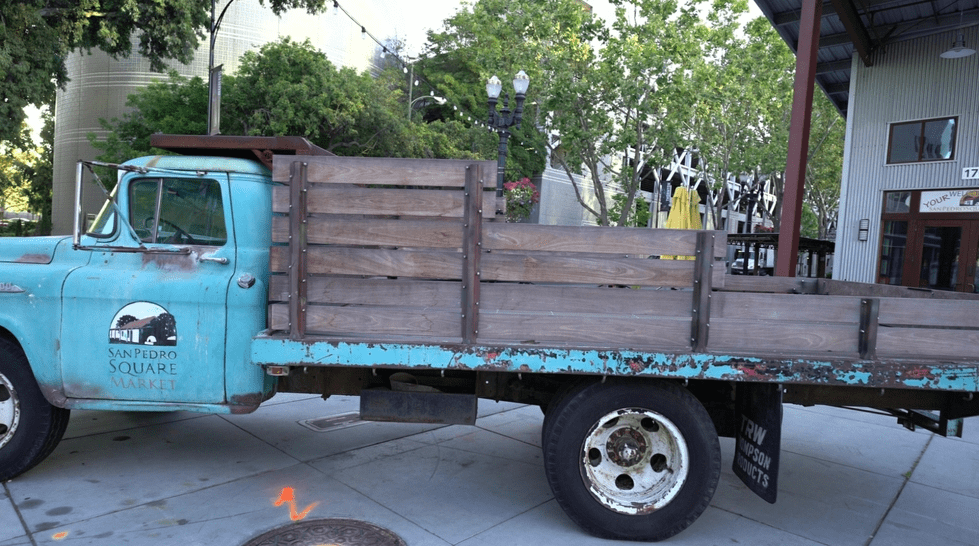}{0.94\mytmplen}{0.42\mytmplen}{0.16\mytmplen}{0.16\mytmplen}{1.2cm}{\mytmplen}{4}{red} &
    \zoomin{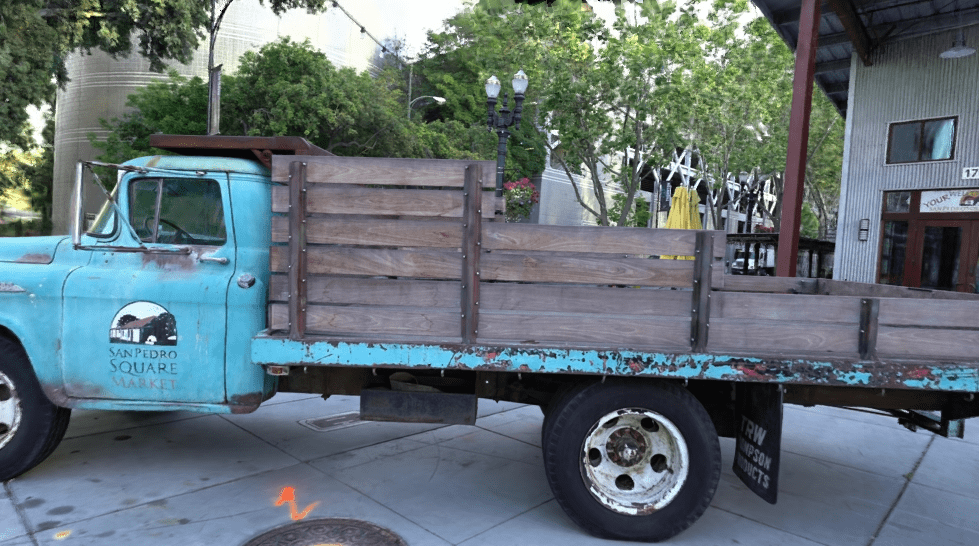}{0.94\mytmplen}{0.42\mytmplen}{0.16\mytmplen}{0.16\mytmplen}{1.2cm}{\mytmplen}{4}{red} &
    \zoomin{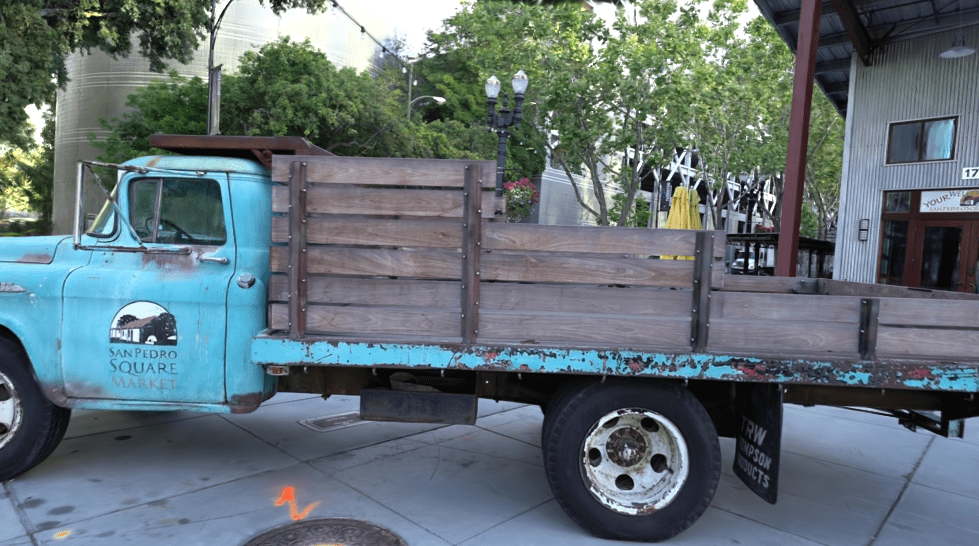}{0.94\mytmplen}{0.42\mytmplen}{0.16\mytmplen}{0.16\mytmplen}{1.2cm}{\mytmplen}{4}{red} &
    \zoomin{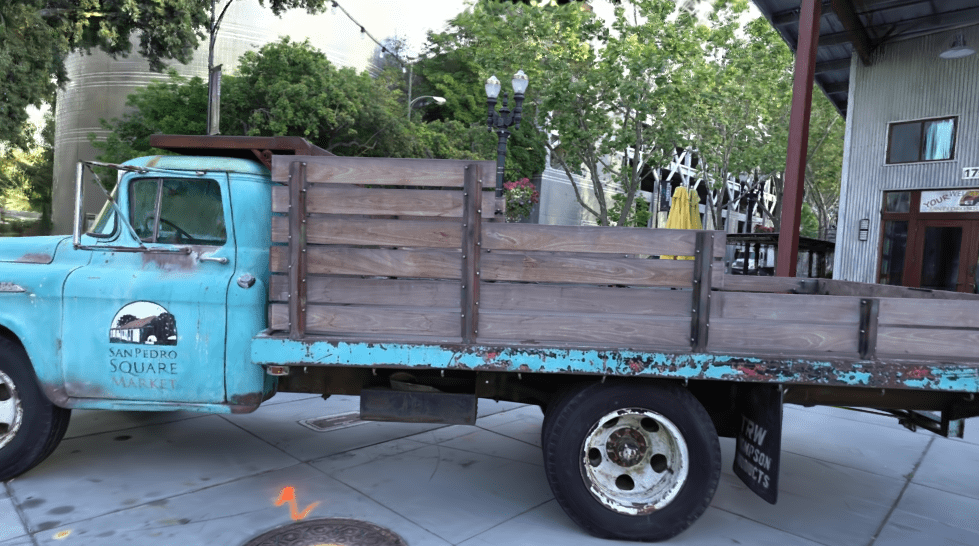}{0.94\mytmplen}{0.42\mytmplen}{0.16\mytmplen}{0.16\mytmplen}{1.2cm}{\mytmplen}{4}{red}\\

\end{tabular}
}

\caption{
\myTitle{Qualitative Comparison between \methodname, 3DGS and 2DGS.} 
3D Convex Splatting achieves high-quality novel view synthesis and fast rendering by representing scenes with 3D smooth convexes. In contrast, the softness of Gaussian primitives often results in blurring and loss of detail, while 3D Convex Splatting effectively captures sharp edges and fine details.
}
\label{fig:qualityresults_supp}
\end{figure*}

\section{More Results}

\subsection{Experiments on Synthetic Data}

Figure \ref{fig:synth_data} illustrates the optimization process of our smooth convexes on four distinct shapes during training.
Our convex shapes are highly versatile and capable of approximating a wide range of different shapes.

\subsection{Real-world Novel View Synthesis}

\mysection{Main results.}
Figure \ref{fig:qualityresults_supp} shows additional qualitative results, highlighting the capabilities of 3D Convex Splatting compared to 3D Gaussians and 2D Gaussians. 
The inherent softness of Gaussian primitives often leads to blurrier images and noticeable artifact-prone regions.
While PSNR favors such blurrier images due to imprecise image alignment, they lack high-quality detail.
Compared to Gaussian, 3DCS does not produce any blurry areas and often results in a rendering much closer to the ground truth. 
For instance, in the \textit{Bicycle} scene, Gaussian methods produce blurry artifacts on the street and in the grass, whereas 3D Convex Splatting achieves a result that closely matches the ground truth.
Tables \ref{tab:1} to \ref{tab:6} shows the complete quantitative results for each scene. 3D Convex Splatting outperforms 3DGS, 2DGS, and GES across all metrics on indoor scenes, the Deep Blending dataset, and the Tanks \& Temples dataset.

\end{document}